\documentclass{article}

\usepackage[dvipsnames,table]{xcolor}

\usepackage{amsfonts} %
\usepackage{graphicx} %
\usepackage{subcaption} %
\usepackage{comment}
\usepackage{booktabs} %
\usepackage{nicematrix} %
\usepackage{multirow}
\usepackage{makecell}
\usepackage{colortbl} %
\usepackage{pifont} %
\usepackage{bbding} %

\usepackage{microtype}
\usepackage{graphicx}
\usepackage{booktabs} %

\usepackage{hyperref}

\usepackage[accepted]{icml2025}

\usepackage{amsmath}
\usepackage{amssymb}
\usepackage{mathtools}
\usepackage{amsthm}

\usepackage[capitalize,noabbrev]{cleveref}

\theoremstyle{plain}

\theoremstyle{definition}

\theoremstyle{remark}

\usepackage[textsize=tiny]{todonotes}

\icmltitlerunning{
Mixture-of-Mamba: Enhancing Multi-Modal State-Space Models with Modality-Aware Sparsity
}

\begin{document}

\twocolumn[
\icmltitle{
Mixture-of-Mamba: Enhancing Multi-Modal 
State-Space Models \\
with Modality-Aware Sparsity
}

\icmlsetsymbol{equal}{*}

\begin{icmlauthorlist}

\icmlauthor{Weixin Liang}{equal,stanford}
\icmlauthor{Junhong Shen}{equal,cmu}
\icmlauthor{Genghan Zhang}{stanford}
\icmlauthor{Ning Dong}{meta}
\icmlauthor{Luke Zettlemoyer}{meta}
\icmlauthor{Lili Yu}{meta}
\end{icmlauthorlist}

\icmlaffiliation{stanford}{Department of Computer Science, Stanford University}
\icmlaffiliation{cmu}{Machine Learning Department, Carnegie Mellon University}
\icmlaffiliation{meta}{FAIR at Meta}

\icmlcorrespondingauthor{Weixin Liang}{wxliang@cs.stanford.edu}

\icmlkeywords{Machine Learning, ICML}

\vskip 0.3in
]

\printAffiliationsAndNotice{\icmlEqualContribution} %

\begin{abstract}

State Space Models (SSMs) have emerged as efficient alternatives to Transformers for sequential modeling, but their inability to leverage modality-specific features limits their performance in multi-modal pretraining. Here, we propose \textbf{Mixture-of-Mamba}, a novel SSM architecture that introduces \textit{modality-aware sparsity} through modality-specific parameterization of the Mamba block. 
Building on \textit{Mixture-of-Transformers} (\href{https://arxiv.org/abs/2411.04996}{W. Liang et al. \textit{arXiv}:2411.04996; 2024}), we extend the benefits of modality-aware sparsity to SSMs while preserving their computational efficiency. 
We evaluate Mixture-of-Mamba across three multi-modal pretraining settings: \textbf{Transfusion} (interleaved text and continuous image tokens with diffusion loss), \textbf{Chameleon} (interleaved text and discrete image tokens), and an extended three-modality framework incorporating \textbf{speech}. Mixture-of-Mamba consistently reaches the same loss values at earlier training steps with significantly reduced computational costs. In the Transfusion setting, Mixture-of-Mamba achieves equivalent image loss using only \textbf{34.76\%} of the training FLOPs at the \textbf{1.4B} scale. In the Chameleon setting, Mixture-of-Mamba reaches similar image loss with just \textbf{42.50\%} of the FLOPs at the \textbf{1.4B} scale, and similar text loss with just \textbf{65.40\%} of the FLOPs. 
In the three-modality setting, MoM matches speech loss at \textbf{24.80\%} of the FLOPs at the \textbf{1.4B} scale. Our ablation study highlights the synergistic effects of decoupling projection components, where joint decoupling yields greater gains than individual modifications. These results establish \textit{modality-aware sparsity} as a versatile and effective design principle, extending its impact from Transformers to SSMs and setting new benchmarks in multi-modal pretraining.
Our code can be accessed at \href{https://github.com/Weixin-Liang/Mixture-of-Mamba}{\texttt{https://github.com/}}\\\href{https://github.com/Weixin-Liang/Mixture-of-Mamba}{\texttt{Weixin-Liang/Mixture-of-Mamba}}.

\end{abstract}

\begin{figure}
    \centering
    \includegraphics[width=\linewidth]{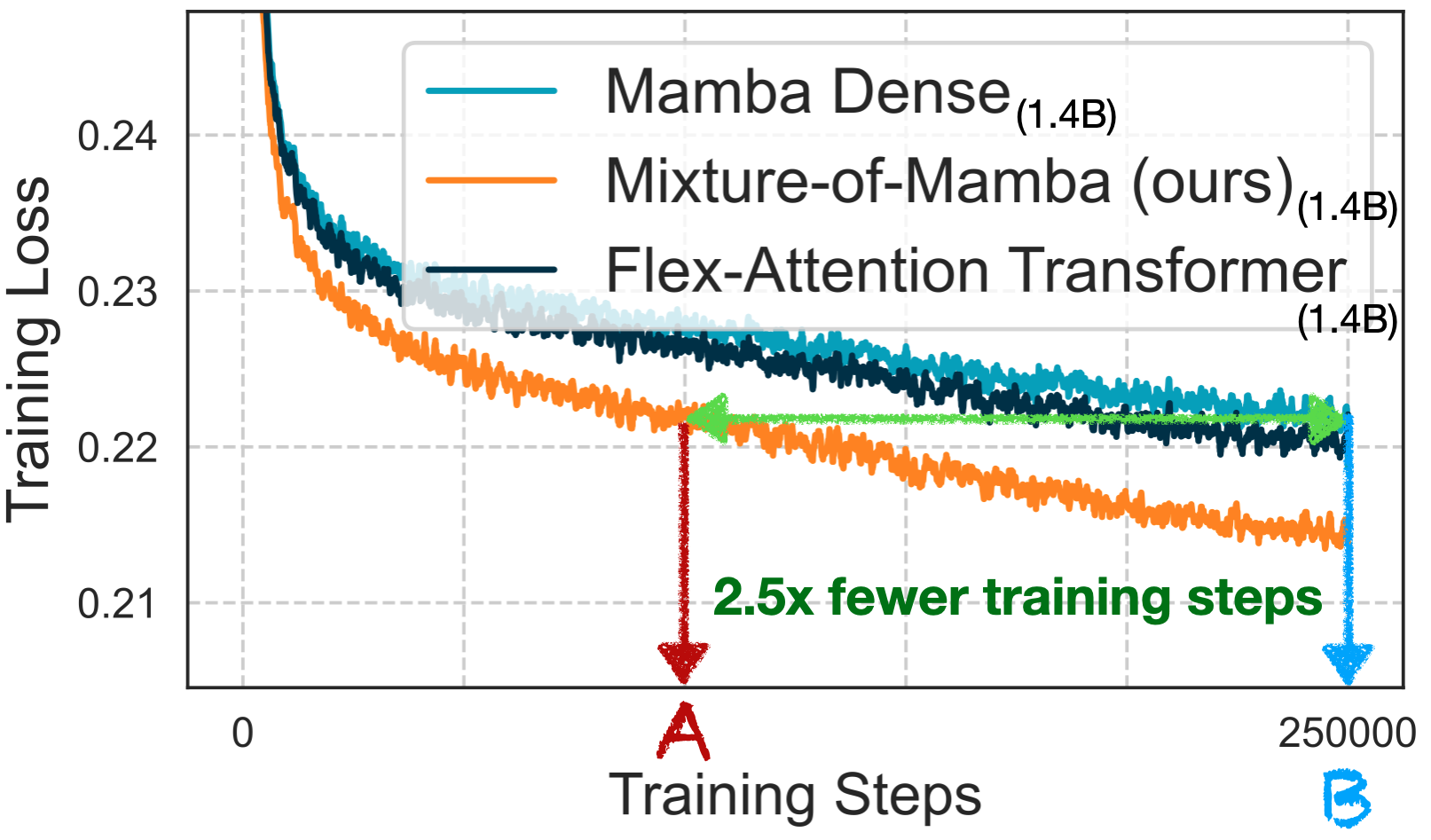}
    \vspace{-7mm}
    \caption{\textbf{Multi-modal pretraining on interleaved text and image data.} Training loss on the {image modality} is shown for models with 1.4B parameters: Mamba Dense (\textcolor{cyan}{cyan}), Flex-Attention Transformer (\textcolor{gray}{dark gray}), and Mixture-of-Mamba (\textcolor{orange}{orange}). The Mixture-of-Mamba achieves significantly lower training loss and requires \textbf{2.5x fewer training steps} (indicated by the green arrow) to reach the same loss level as the other baselines.}
    \label{fig:teaser}
    \vspace{-5mm}
\end{figure}

\section{Introduction}

State Space Models (SSMs)~\citep{gu2021efficiently, gu2023mamba} have emerged as efficient alternatives to Transformers for sequential modeling, offering linear scaling in sequence length and strong performance in single-modality tasks. {Mamba}, a recent SSM variant, has demonstrated exceptional efficiency and scalability across diverse tasks by leveraging advanced gating mechanisms and selective state-space scanning~\citep{gu2023mamba}. Despite these advantages, SSMs, including Mamba, remain inherently dense, applying the same set of parameters across all input tokens, regardless of modality. This uniform parameterization limits their ability to capture modality-specific features, leading to suboptimal performance in multi-modal pretraining.

Recent efforts have extended SSMs to multi-modal tasks. Works like VLMamba~\citep{qiao2024vl} and Cobra~\citep{zhao2024cobra} augment Mamba for vision-language modeling by adding LLaVA-style projection modules that map image features into the token space of Mamba. In the vision domain, Vision Mamba~\citep{zhu2024vision} and VMamba~\citep{liu2024vmamba} incorporate bidirectional scanning schemes and selective 2D scanning paths for image patch modeling. Similarly, Mamba has been explored for diffusion-based image and video generation, as seen in DiffuSSM~\citep{yan2024diffusion} and Zigma~\citep{hu2024zigma}, which employ unique state-space scanning patterns. While these approaches demonstrate the adaptability of Mamba, they are orthogonal to our focus, which introduces \textbf{modality-aware sparsity} directly into the Mamba block itself.

A promising approach to address such limitations is \textbf{model sparsity}, exemplified by Mixture-of-Experts (MoE)~\citep{jacobs1991adaptive, eigen2013learning, ShazeerMoE, gshard, switchtransformer, mixtral, branch-train-mix}. MoE reduces computational load by activating only a subset of model components for each input token, allowing experts to specialize in specific aspects of the data. Despite its potential, MoE-based architectures face challenges such as imbalanced expert utilization, bi-level optimization instability, and inefficient load balancing~\citep{ShazeerMoE, gshard, switchtransformer,Shen_Yang_2021, xu2024specialized}. These issues motivate the need for alternative sparse architectures that are computationally efficient and easier to optimize.

In multi-modal contexts, prior work~\citep{vlmo, imageasaforeignlanguage, shen2023vlmoe, lin2024moma} has introduced \textit{modality-aware sparsity} in Transformer-based MoE architectures. These approaches activate specific experts or parameters based on modality, enabling models to specialize in handling diverse data types. Other methods fine-tune modality-specific modules atop dense LLM backbones~\citep{wang2023cogvlm, he2024mars, shen2023cross, shen2024ups}. Such methods show that simple rule-based modality routing often outperforms learned routing, likely due to improved training stability and reduced optimization challenges.

The closest work to our approach is MoE-Mamba~\citep{pióro2024moemamba} and the related Blackmamba architecture~\citep{anthony2024blackmamba}, which interleave Mamba blocks with MoE-augmented MLP layers. While effective, these hybrid designs apply sparsity only to the MLP layers, leaving the dense Mamba blocks unmodified. In contrast, we present \textbf{Mixture-of-Mamba}, a novel architecture that directly introduces \textit{modality-aware sparsity} into the Mamba block itself. Inspired by Mixture-of-Transformers~\citep{liang2024mixture}, our approach dynamically selects modality-specific weights in every input processing component of Mamba, enabling stable and efficient multi-modal pretraining. Furthermore, prior work~\citep{liang2024mixture} shows that MoE techniques can complement sparse architectures like Mixture-of-Transformers, suggesting that Mixture-of-Mamba and MoE-based MLP sparsification can be combined to achieve further gains.

To rigorously evaluate Mixture-of-Mamba, we conduct experiments across three multi-modal pretraining settings:
\begin{itemize}
    \item \textbf{Transfusion:} Interleaved text and continuous image tokens with distinct autoregressive and diffusion-based objectives. Mixture-of-Mamba achieves equivalent image loss using only \textbf{34.76\%} of the training FLOPs at the \textbf{1.4B} scale.
    \item \textbf{Chameleon:} Interleaved text and discrete image tokens. Mixture-of-Mamba reaches similar image loss with just \textbf{42.50\%} of the FLOPs and similar text loss with only \textbf{65.40\%} of the FLOPs at the \textbf{1.4B} scale.
    \item \textbf{Three-Modality:} Extension of the Chameleon setting to include speech. Mixture-of-Mamba matches speech loss using only \textbf{24.80\%} of the FLOPs at the \textbf{1.4B} scale, while maintaining strong performance across image and text modalities.
\end{itemize}

Additionally, we perform an ablation study to analyze the contribution of modality-specific parameterization. Our findings reveal a synergistic effect: jointly decoupling all components yields greater gains than individual modifications, underscoring the importance of modality-aware sparsity as a holistic design principle.

In summary, Mixture-of-Mamba establishes a versatile and efficient architecture for SSMs by extending \textit{modality-aware sparsity} into the Mamba block. This approach delivers robust performance gains and substantial computational savings across diverse multi-modal settings, setting new benchmarks in scalable multi-modal pretraining.

\section{Method}

\subsection{The Mixture-of-Mamba Block}

\begin{algorithm}[htb]
    \caption{Mixture-of-Mamba block}
    \label{alg:mom}
    \begin{algorithmic}[1]
    \INPUT $F_{in}, A, W_{in\_proj}, W_{x\_proj}, W_{dt\_proj}, W_{out\_proj}, b, M$
    \OUTPUT $F_{out}$
    \STATE $x, z \leftarrow \mathcal{M}(F_{in}, W_{in\_proj}; M)$ \hfill $\triangleright$ Block starts
    \STATE $u \leftarrow \text{SiLU}(\text{Conv1D}(x))$ \hfill $\triangleright$ [b,$\ell$,d]
    \STATE $\delta, B, C \leftarrow \mathcal{M}(u, W_{x\_proj}; M)$ \hfill $\triangleright$ [b,$\ell$,(r,n,n)]
    \STATE $\Delta \leftarrow  \text{log}(1+\text{exp}((\mathcal{M}(\delta, W_{dt\_proj}, b; M))))$ %
    \STATE $\overline{A} \leftarrow  \Delta * A$ \hfill $\triangleright$ [b,$\ell$,d,n]
    \STATE $\overline{B} \leftarrow  \Delta * (u \times B)$ \hfill $\triangleright$ [b,$\ell$,d,n]
    \STATE $h = 0$ \hfill $\triangleright$ [b,d,n]
    \FOR {$i = 0 ... N-1$}
    \STATE $h = h * \overline{A_i} + \overline{B_i}$ \hfill $\triangleright$ [b,d,n]
    \STATE $y_i = h \cdot C_i$ \hfill $\triangleright$ [b,d]
    \ENDFOR
    \STATE $o \leftarrow  (y + u) * \text{SiLU}(z)$
    \STATE $F_{out} \leftarrow  \mathcal{M}(o, W_{out\_proj}; M)$ \hfill $\triangleright$ Block ends
    \STATE $ $
    \FUNCTION{$\mathcal{M}(X, W, b=\text{None}; M)$}
    \FOR{$\text{each modality } m\in M$}
    \STATE $I_m \leftarrow \{i:m_i=m\}$
    \STATE $X_m \leftarrow \{x_i:i\in I_m\}$
    \STATE $Y_m \leftarrow X_mW_m + b_m$
    \ENDFOR
    \STATE return $Y \leftarrow \cup_{m\in M}Y_m $
    \ENDFUNCTION
    \end{algorithmic}
\end{algorithm}
Our hypothesis is that explicitly parametrizing the selection in SSMs with the modality can improve the data efficiency of multi-modality training \cite{liang2024mixture}. 

Following the setting of other SSMs~\cite{gu2021efficiently}, Mixture-of-Mamba is composed of homogeneous Mixture-of-Mamba blocks (line 1-13 of Algorithm~\ref{alg:mom}). 

In Mixture-of-Mamba, \textit{modality-specific parameterization} is applied to all projections that explicitly process input features belonging to a single modality, including input projection (\ding{202} \(W_{{in\_proj}}\)), intermediate projections (\ding{203} \(W_{{x\_proj}}\) and \ding{204} \(W_{{dt\_proj}}\)), and output projection (\ding{205} \(W_{{out\_proj}}\)). Conv1D and state transitions $A$ remain shared because they operate across multiple features or on aggregated RNN-like states, where the notion of modality is not well-defined. 
After parametrized by modality $M$, the linear transformation $XW+b$ becomes $\mathcal{M}(X, W, b; M)$. $\mathcal{M}$ applies the weight of modality $m$ ($W_m$) to tokens of modality $m$ ($X_m$) in parallel based on the modality mask. The output shape of $\mathcal{M}$ is the same as the corresponding linear transformation.

The shape of $W_{in\_proj}$ is [f,(d,d)] where f is the feature dimension of input $F_{in}$ and d is the expanded feature dimension. These two projections are fused together for efficiency and $W_{x\_proj}$ uses the same technique. Line 1, 12 and 13 can be viewed as a SwiGLU~\cite{swiglu} around the conv+SSM (Line 2-12). $x$ is passed to conv+SSM and $z$ will be transformed to the gate in SwiGLU.

The \texttt{Conv1D} in Line 2 can help collect local information across time as observed in~\cite{sun2024learning}. Similarly, \texttt{Conv1D} can also gather local information across modalities and we keep the weight-sharing property of convolution without separating the convolution kernel into different modalities.

Line 3-12 is multi-modality selective SSM. It is composed of parameter preparation (line 3-6), RNN update (line 7-11), and residual connection (line 12). 

$\Delta$ is the discretization time step. It is derived from $u$ through a low-rank approximation $u\rightarrow\delta\rightarrow\Delta$ followed by a \texttt{softplus} as shown in Line 3 and 4. $A$ is of shape [d,n] and $\Delta$ is of shape [b,$\ell$,d] where b is batch size, $\ell$ is sequence length, and n is the state dimension. Line 5 is a broadcast element-wise multiplication where $\Delta$ is unsqueezed to [b,$\ell$,d,1] and repeated to [b,$\ell$,d,n]. Line 6 first applies a batched outer product between $u$ [b,$\ell$,d] and $B$ [b,$\ell$,n] whose result is element-wise multiplied with $\Delta$. Line 5 and 6 apply the selection to $A$,$B$ and get $\Bar{A},\Bar{B}$, respectively. $\Bar{B}$ can be viewed as a gated input $u$ and $\Bar{A}$ can be viewed as a selection gate on the state $h$. 

Line 7-10 is a typical RNN operator with state $h$ and output $y_i$. The $y_i$'s are concatenated together as output $y$. The gate application on input $u$ is fused with gate parameter preparation at line 6 for efficiency.

Line 12 first adds the input $u$ to the output $y$ as residual, which is the final output of SSM. Then, Line 12 applies the gate of ``SwiGLU'' to the output of SSM. Finally, line 13 projects $o$ back to the feature dimension.

\begin{figure}[tb]
    \centering
    \includegraphics[width=\linewidth]{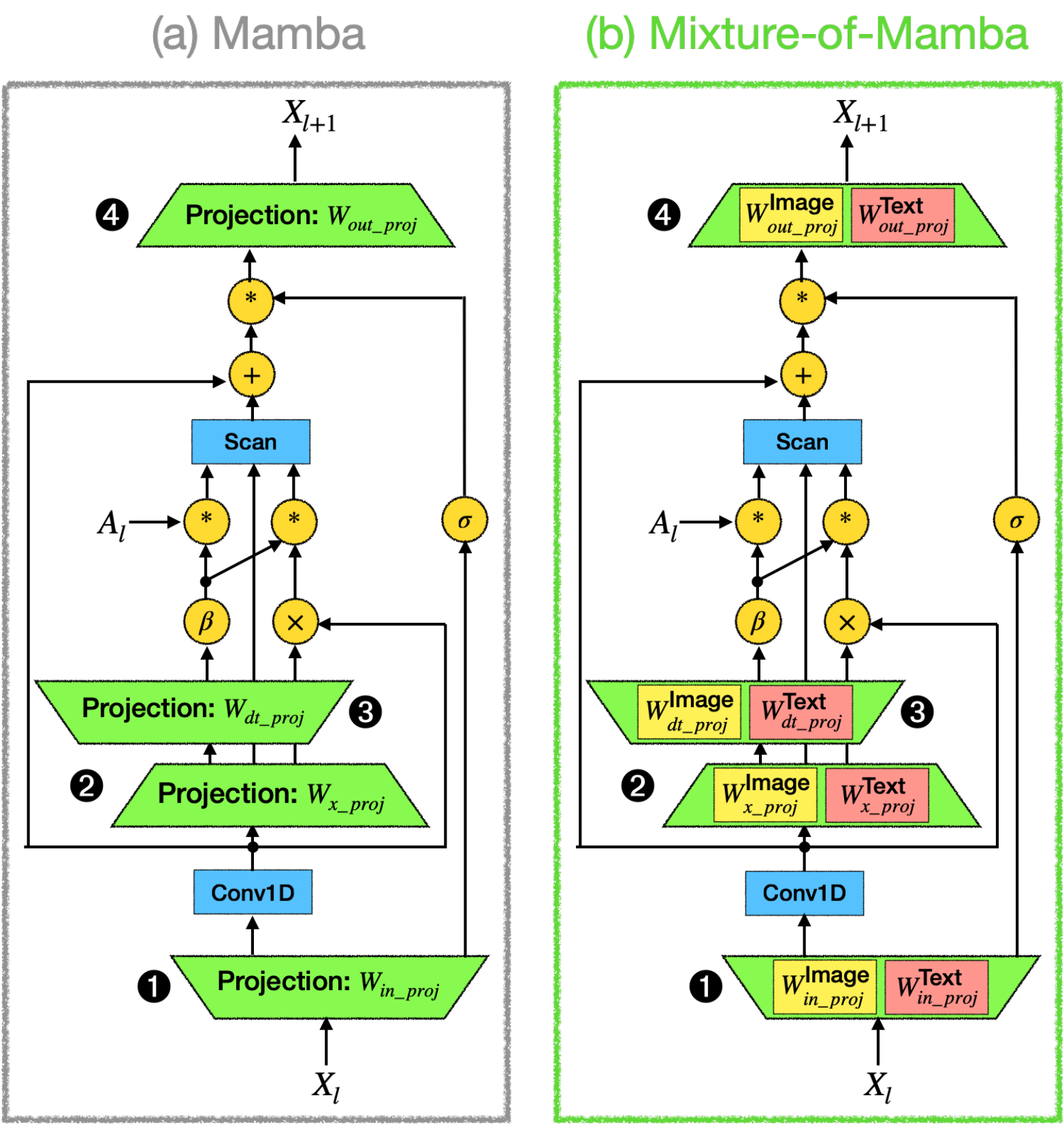}
    \vspace{-3mm}
\caption{
\small 
\textbf{Comparison of (a) the original Mamba block and (b) the proposed Mixture-of-Mamba block. }
In Mixture-of-Mamba, \textit{modality-specific parameterization} is applied to all projections that explicitly process input features belonging to a single modality, including input projection (\ding{202} \(W_{{in\_proj}}\)), intermediate projections (\ding{203} \(W_{{x\_proj}}\) and \ding{204} \(W_{{dt\_proj}}\)), and output projection (\ding{205} \(W_{{out\_proj}}\)). Conv1D and state transitions $A$ remain shared because they operate across multiple features or on aggregated RNN-like states, where the notion of modality is not well-defined. By selectively decoupling these projections, Mixture-of-Mamba enables \textit{modality-aware sparsity} without compromising computational efficiency.}
    \label{fig:method}
    \vspace{-3mm}
\end{figure}

\subsection{Multi-objective Training with Diffusion}

Following Transfusion~\cite{transfusion}, Mixture-of-Mamba is trained on interleaved multi-modal sequences of discrete text tokens and continuous image tokens using a combined objective that incorporates both language modeling and diffusion-based image generation. Each image is encoded as a sequence of latent patches using a Variational Autoencoder (VAE), where each patch is represented as a continuous vector. The patches are sequenced left-to-right, top-to-bottom, and inserted into the discrete text sequence.  

The diffusion process follows the Denoising Diffusion Probabilistic Models (DDPM)~\cite{ho2020denoising}, where Gaussian noise is progressively added to the latent image patches during the forward process. Given a clean latent patch $\mathbf{x}_0$, a noised version $\mathbf{x}_t$ at timestep $t$ is created as:
\begin{equation}
\mathbf{x}_t = \sqrt{\bar{\alpha}_t}\mathbf{x}_0 + \sqrt{1 - \bar{\alpha}_t} \boldsymbol{\epsilon}, \quad \boldsymbol{\epsilon} \sim \mathcal{N}(\mathbf{0}, \mathbf{I}),
\label{eq:forward}
\end{equation}
where $\bar{\alpha}_t$ is determined by a cosine noise schedule~\cite{nichol2021improved}, approximated as $\sqrt{\bar{\alpha}_t} \approx \cos(\frac{t}{T} \cdot \frac{\pi}{2})$ with adjustments. During training, noise is added to the latent patches at a randomly selected timestep $t$, and the model is optimized to predict the noise $\boldsymbol{\epsilon}$.

The overall training objective combines the autoregressive language modeling loss $\mathcal{L}_{\text{LM}}$, applied to the discrete text tokens, with the diffusion loss $\mathcal{L}_{\text{DDPM}}$, applied to the latent image patches:
\begin{equation}
\mathcal{L} = \mathcal{L}_{\text{LM}} + \lambda \cdot \mathcal{L}_{\text{DDPM}},
\end{equation}
where $\lambda$ balances the contributions of the two losses.

Importantly, the conditioning for image generation is naturally embedded within the interleaved sequence. When denoising image patches, the preceding tokens—including both text describing the image and prior images—serve as context for conditional generation. This unified approach enables Mixture-of-Mamba to leverage the modality-aware sparsity to efficiently model both local intra-image dependencies and long-range inter-modal relationships across the sequence.

\subsection{Training with Uniform Representations}

As an alternative to the multi-objective training paradigm, we explore a unified representation strategy in which both text and image modalities are represented as discrete tokens. Following the Chameleon framework~\cite{chameleonteam}, we treat the image data as sequences of discrete tokens obtained through a pre-trained VQ-VAE model~\cite{make_a_scene}. Specifically, each image is encoded into a fixed number of tokens (e.g., 1,024) by quantizing its latent features into a learned codebook. These tokens are then arranged sequentially, similar to the processing of text tokens, resulting in a uniform discrete representation across both modalities.

During training, both text and image tokens are processed using the same autoregressive objective, where the model learns to predict the next token in the sequence given all previous tokens. Formally, the training objective is:
\begin{equation}
\mathcal{L}_{\text{uniform}} = \mathbb{E}_{\mathbf{x}_{1:T}} \left[ -\log P(\mathbf{x}_t \mid \mathbf{x}_{1:t-1}) \right],
\end{equation}
where $\mathbf{x}_{1:T}$ represents the interleaved sequence of text and image tokens. This objective allows the model to treat text and image data equivalently, unifying the training process across modalities while relying solely on an autoregressive loss.
The use of discrete tokens for images simplifies the training procedure by removing the need for separate loss formulations, as in the diffusion-based approach. It also aligns with the inherent sequence-to-sequence nature of Mixture-of-Mamba, where the same modality-aware sparsity design can be applied seamlessly across the discrete text and image tokens.

\paragraph{Motivation and Robustness Testing.}  
We include this alternative strategy to evaluate the robustness of our Mixture-of-Mamba architecture under different choices of training objectives and data representations. By experimenting with uniform discrete representations, we demonstrate that Mixture-of-Mamba consistently outperforms Mamba Dense models across various settings, including both continuous (multi-objective) and discrete (uniform) representations. This highlights the versatility of Mixture-of-Mamba and its ability to deliver performance gains regardless of the underlying choice of modality representations or training objectives.

\begin{figure*}[t]
    \centering

    \begin{subfigure}[b]{0.24\textwidth}
        \centering
        \includegraphics[width=\textwidth]{Figures.HandDraw/loss_curve_transfusion_1.4b_moe4t1i_train_avg_image_loss_AVG.png}
        \caption{\textbf{1.4B} Image Training Loss}
    \end{subfigure}
    \hfill
    \begin{subfigure}[b]{0.24\textwidth}
       \centering
       \includegraphics[width=\textwidth]{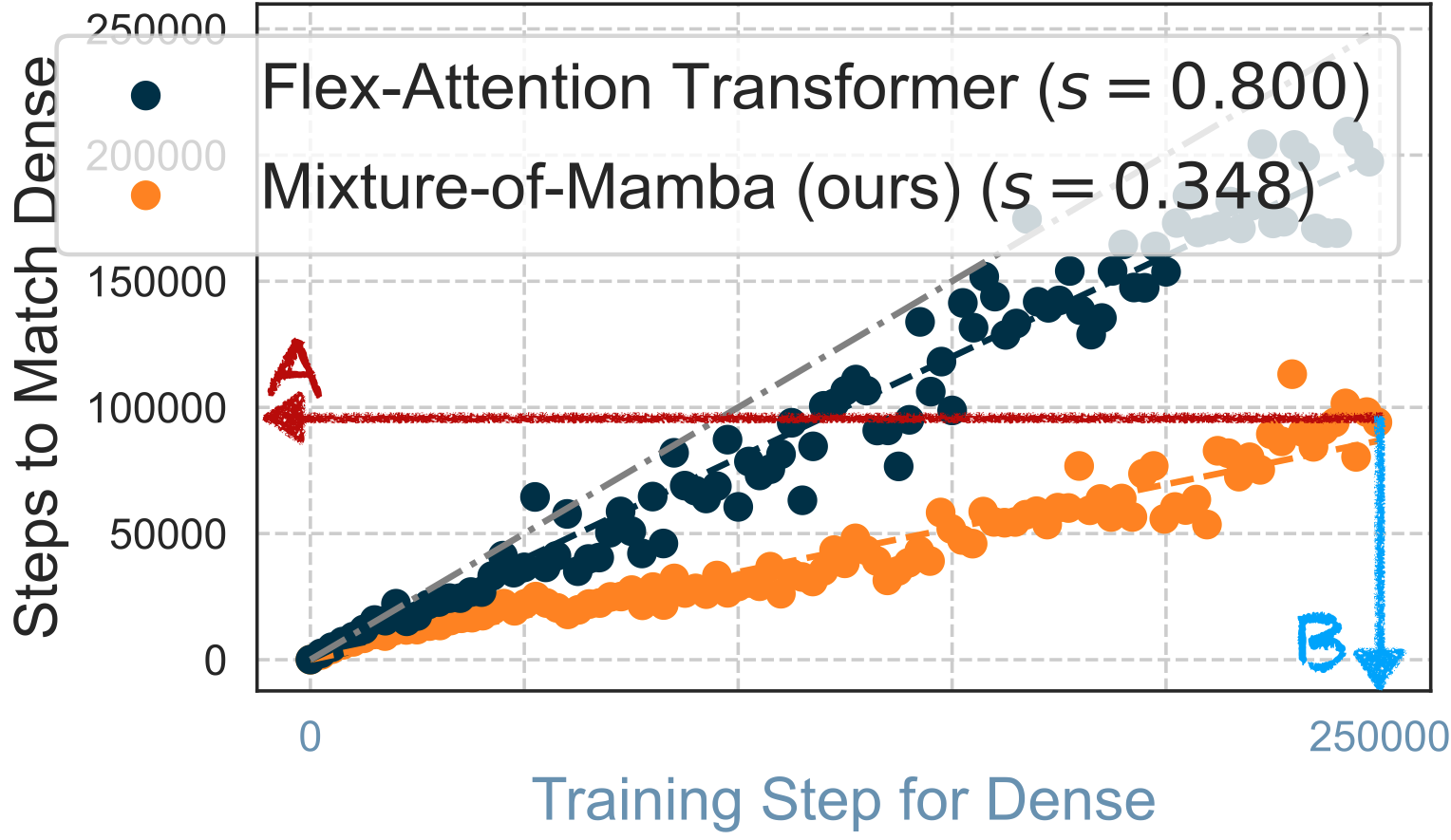}
       \caption{\textbf{1.4B} Image Loss Matching}
   \end{subfigure}
    \hfill   
   \begin{subfigure}[b]{0.24\textwidth}
        \centering
        \includegraphics[width=\textwidth]{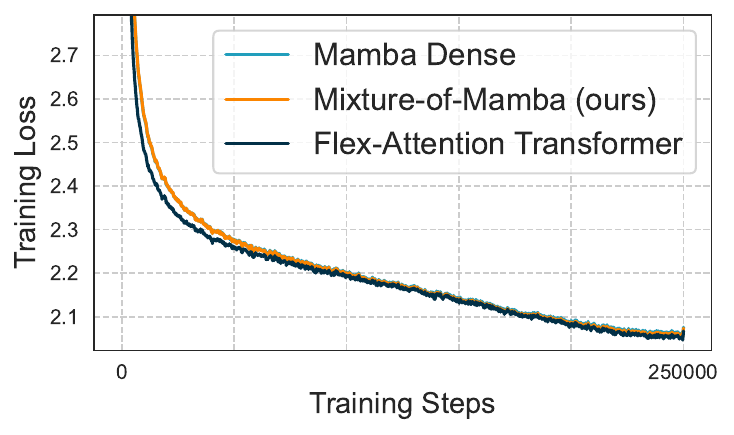}
        \caption{\textbf{1.4B} Text Training Loss}
    \end{subfigure}
    \hfill
    \begin{subfigure}[b]{0.24\textwidth}
       \centering
       \includegraphics[width=\textwidth]{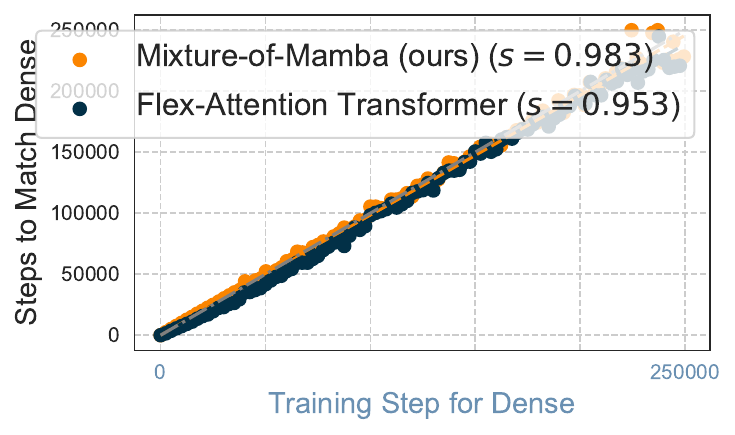}
       \caption{\textbf{1.4B} Text Loss Matching}
   \end{subfigure}

    \begin{subfigure}[b]{0.24\textwidth}
        \centering
        \includegraphics[width=\textwidth]{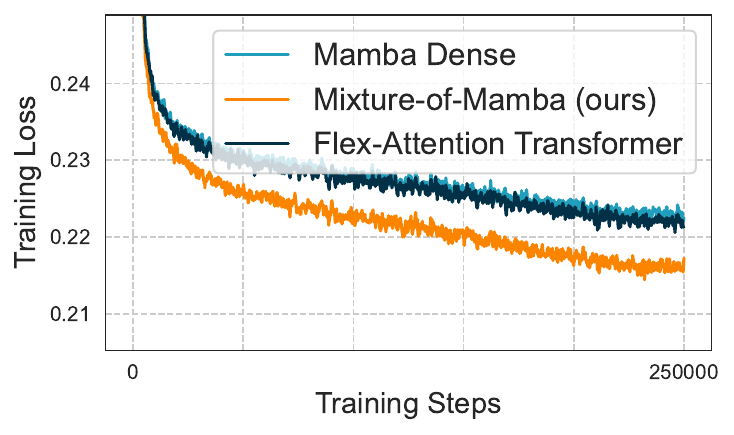}
        \caption{\textbf{760M} Image Training Loss}
    \end{subfigure}
    \hfill
    \begin{subfigure}[b]{0.24\textwidth}
       \centering
       \includegraphics[width=\textwidth]{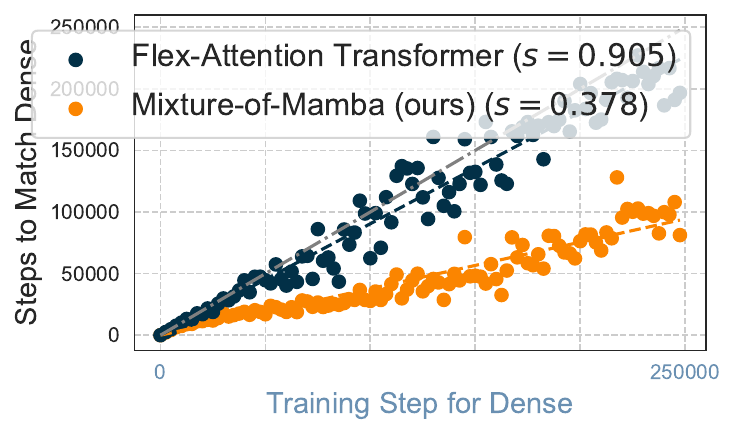}
       \caption{\textbf{760M} Image Loss Matching}
   \end{subfigure}
    \hfill   
   \begin{subfigure}[b]{0.24\textwidth}
        \centering
        \includegraphics[width=\textwidth]{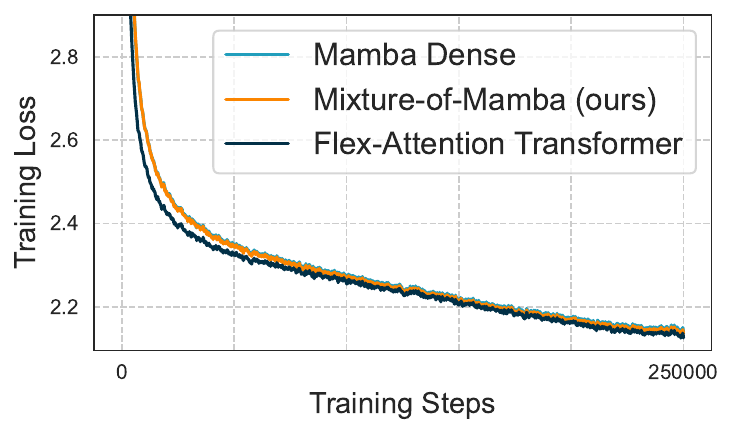}
        \caption{\textbf{760M} Text Training Loss}
    \end{subfigure}
    \hfill
    \begin{subfigure}[b]{0.24\textwidth}
       \centering
       \includegraphics[width=\textwidth]{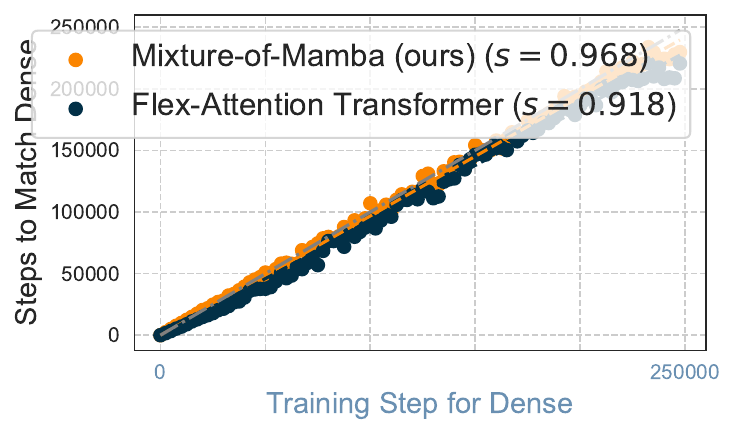}
       \caption{\textbf{760M} Text Loss Matching}
   \end{subfigure}

    \begin{subfigure}[b]{0.24\textwidth}
        \centering
        \includegraphics[width=\textwidth]{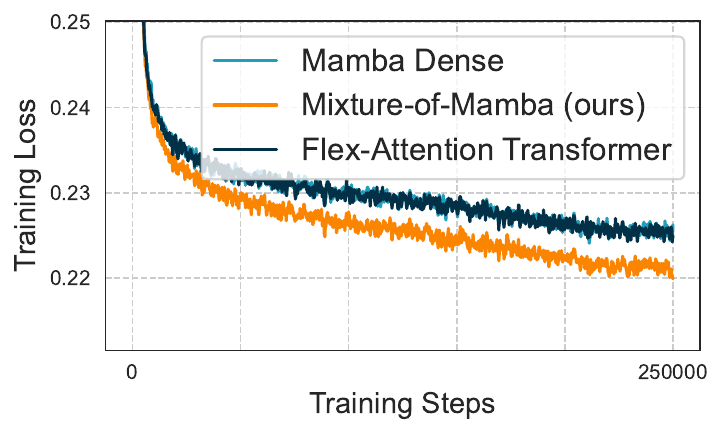}
        \caption{\textbf{163M} Image Training Loss}
    \end{subfigure}
    \hfill
    \begin{subfigure}[b]{0.24\textwidth}
       \centering
       \includegraphics[width=\textwidth]{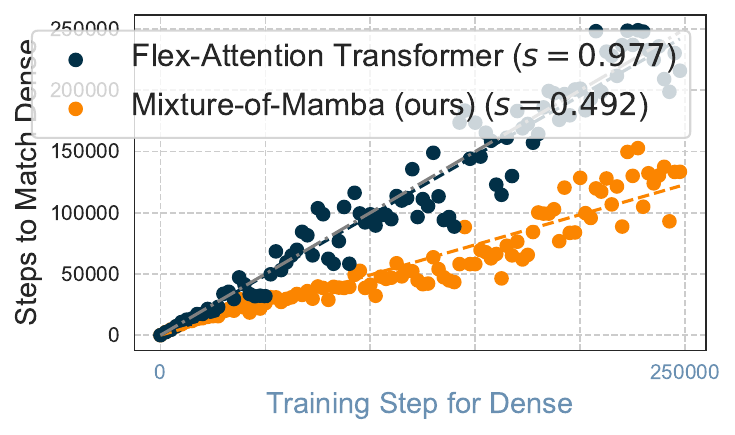}
       \caption{\textbf{163M} Image Loss Matching}
   \end{subfigure}
    \hfill   
   \begin{subfigure}[b]{0.24\textwidth}
        \centering
        \includegraphics[width=\textwidth]{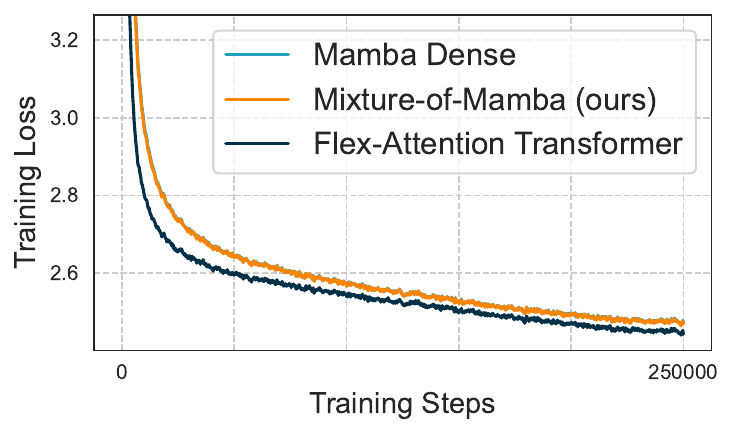}
        \caption{\textbf{163M} Text Training Loss}
    \end{subfigure}
    \hfill
    \begin{subfigure}[b]{0.24\textwidth}
       \centering
       \includegraphics[width=\textwidth]{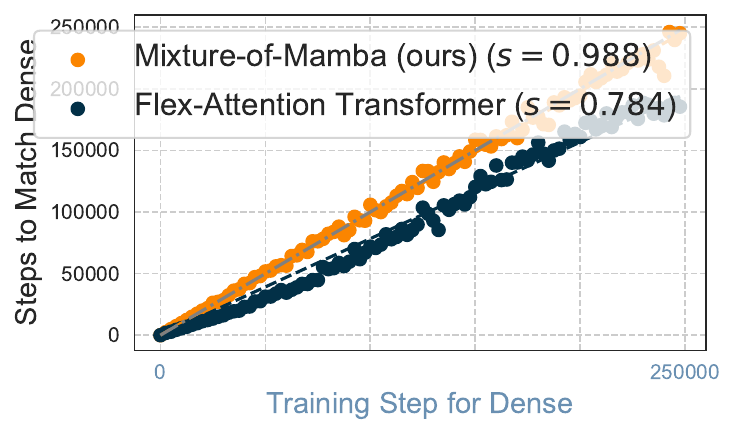}
       \caption{\textbf{163M} Text Loss Matching}
   \end{subfigure}

    \caption{\textbf{Multi-modal pretraining in the Transfusion setting on interleaved text and image data across model scales.} 
    Training loss and loss matching are reported for image and text modalities at three model sizes: \textbf{1.4B}, \textbf{760M}, and \textbf{163M}. 
    \textbf{(a, e, i)} Image training loss shows significant improvements for Mixture-of-Mamba (\textcolor{orange}{orange}), which consistently achieves lower loss compared to Mamba Dense (\textcolor{cyan}{cyan}) and Flex-Attention Transformer (\textcolor{darkgray}{dark gray}) across all scales. 
    \textbf{(b, f, j)} Image loss matching compares the training dynamics and shows that Mixture-of-Mamba and Flex-Attention Transformer reach the same loss values at earlier training steps compared to Mamba Dense. 
    \textbf{(c, g, k)} Text training loss shows competitive results, with Mixture-of-Mamba performing better than Mamba Dense and on par with the Flex-Attention Transformer. 
    \textbf{(d, h, l)} Text loss matching illustrates that Mixture-of-Mamba and Flex-Attention Transformer exhibit more efficient training dynamics than Mamba Dense, requiring fewer steps to achieve comparable loss values, though the primary improvements are observed in the image modality. 
    Overall, in the \textbf{Transfusion setting}, Mixture-of-Mamba demonstrates substantial gains in image loss and training efficiency while maintaining strong performance on text.}
    \label{fig:transfusion_training}
    
\end{figure*}

\begin{table*}[ht]
    \centering
    \resizebox{\textwidth}{!}{
    \begin{NiceTabular}{llccccc}[colortbl-like]
    \toprule
    \makecell{\textbf{Model} \\ \textbf{Scale}} & \textbf{Metric Category} & \textbf{Metric Name} & 
    \makecell{\colorbox{cyan!20}{\textbf{Mamba}} \\ \textbf{Loss ($\downarrow$)}}
    & \makecell{\colorbox{Dandelion!50}{\textbf{Mixture-of-}} \\ \textbf{\colorbox{Dandelion!50}{Mamba} Loss ($\downarrow$)}} & \makecell{\textbf{Performance} \\ \textbf{Gain (\%)  ($\uparrow$)}} & \makecell{\textbf{Relative Training FLOPs} \\ \textbf{to Match Mamba (\%) ($\downarrow$)}} \\
    \midrule
    \multirow{7}{*}{443M} & \multirow{3}{*}{\textbf{\colorbox{blue!20}{Image} Metrics}} & Training Loss          & 5.3558 & 5.1703 & \cellcolor{green!35}3.46\%  & \cellcolor{green!35}33.40\% \\
                          &                                      & Obelisc Val. Loss      & 4.5258 & 4.3546 & \cellcolor{green!40}3.78\%  & \cellcolor{green!35}35.10\% \\
                          &                                      & SSTK Val. Loss         & 5.9179 & 5.7471 & \cellcolor{green!30}2.89\%  & \cellcolor{green!35}35.30\% \\
    \cmidrule{2-7}
                          & \multirow{3}{*}{\textbf{\colorbox{OrangeRed!20}{Text} Metrics}} & Training Loss          & 2.4637 & 2.3864 & \cellcolor{green!30}3.14\%  & \cellcolor{green!30}62.00\% \\
                          &                                      & Obelisc Val. Loss      & 3.0544 & 2.9820 & \cellcolor{green!20}2.37\%  & \cellcolor{green!20}66.70\% \\
                          &                                      & SSTK Val. Loss         & 2.7569 & 2.6250 & \cellcolor{green!40}4.78\%  & \cellcolor{green!30}54.70\% \\
    \cmidrule{2-7}
                          & \multirow{1}{*}{\textbf{Overall}}     & Avg Training Loss      & 3.6584 & 3.5364 & \cellcolor{green!35}3.33\%  & \cellcolor{green!30}47.90\% \\
    \midrule
    \multirow{7}{*}{880M} & \multirow{3}{*}{\textbf{\colorbox{blue!20}{Image} Metrics}} & Training Loss          & 5.2260 & 5.1201 & \cellcolor{green!10}2.03\%  & \cellcolor{green!30}48.40\% \\
                          &                                      & Obelisc Val. Loss      & 4.4127 & 4.3105 & \cellcolor{green!20}2.32\%  & \cellcolor{green!30}49.30\% \\
                          &                                      & SSTK Val. Loss         & 5.7987 & 5.6986 & \cellcolor{green!10}1.73\%  & \cellcolor{green!30}50.50\% \\
    \cmidrule{2-7}
                          & \multirow{3}{*}{\textbf{\colorbox{OrangeRed!20}{Text} Metrics}} & Training Loss          & 2.3073 & 2.2438 & \cellcolor{green!20}2.75\%  & \cellcolor{green!30}65.60\% \\
                          &                                      & Obelisc Val. Loss      & 2.8886 & 2.8313 & \cellcolor{green!10}1.99\%  & \cellcolor{green!10}72.80\% \\
                          &                                      & SSTK Val. Loss         & 2.5483 & 2.4548 & \cellcolor{green!30}3.67\%  & \cellcolor{green!20}67.90\% \\
    \cmidrule{2-7}
                          & \multirow{1}{*}{\textbf{Overall}}     & Avg Training Loss      & 3.5130 & 3.4320 & \cellcolor{green!20}2.31\%  & \cellcolor{green!30}58.30\% \\
    \midrule
    \multirow{7}{*}{1.5B} & \multirow{3}{*}{\textbf{\colorbox{blue!20}{Image} Metrics}} & Training Loss          & 5.1892 & 5.0591 & \cellcolor{green!20}2.51\%  & \cellcolor{green!30}42.50\% \\
                          &                                      & Obelisc Val. Loss      & 4.3692 & 4.2510 & \cellcolor{green!20}2.71\%  & \cellcolor{green!30}44.50\% \\
                          &                                      & SSTK Val. Loss         & 5.7546 & 5.6335 & \cellcolor{green!10}2.10\%  & \cellcolor{green!30}44.60\% \\
    \cmidrule{2-7}
                          & \multirow{3}{*}{\textbf{\colorbox{OrangeRed!20}{Text} Metrics}} & Training Loss          & 2.2284 & 2.1614 & \cellcolor{green!30}3.01\%  & \cellcolor{green!30}65.40\% \\
                          &                                      & Obelisc Val. Loss      & 2.8020 & 2.7393 & \cellcolor{green!20}2.24\%  & \cellcolor{green!10}71.60\% \\
                          &                                      & SSTK Val. Loss         & 2.4614 & 2.3455 & \cellcolor{green!35}4.71\%  & \cellcolor{green!20}62.10\% \\
    \cmidrule{2-7}
                          & \multirow{1}{*}{\textbf{Overall}}     & Avg Training Loss      & 3.4602 & 3.3670 & \cellcolor{green!20}2.69\%  & \cellcolor{green!30}54.70\% \\
    \bottomrule
    \end{NiceTabular}}
\caption{\textbf{Training and validation metrics across model scales in the Chameleon setting.} In this setting, both image and text modalities are represented as discrete tokens. Mixture-of-Mamba achieves substantial performance improvements over Mamba Dense, with the \textbf{image modality} showing the largest gains. The \textbf{text modality} also exhibits significant improvements, in contrast to the Transfusion setting where text gains were more modest. The current table shows results for three model scales: \textbf{443M}, \textbf{880M}, and \textbf{1.5B}, due to space constraints. See Appendix Table~\ref{tab:full_chameleon_metrics} for the full results across all five model scales: \textbf{37M}, \textbf{94M}, \textbf{443M}, \textbf{880M}, and \textbf{1.5B}. These results further highlight the effectiveness and efficiency of Mixture-of-Mamba, which consistently achieves strong performance with reduced relative training FLOPs.}
\label{tab:chameleon_metrics}

\end{table*}

\begin{table*}[ht]
    \centering
    \resizebox{\textwidth}{!}{
    \begin{NiceTabular}{llccccc}[colortbl-like]
    \toprule
    \makecell{\textbf{Model} \\ \textbf{Scale}} & \textbf{Metric Category} & \textbf{Metric Name} & \textbf{\colorbox{cyan!20}{Mamba} Loss  ($\downarrow$)} & \makecell{\colorbox{Dandelion!50}{\textbf{Mixture-of-}} \\ \textbf{\colorbox{Dandelion!50}{Mamba} Loss ($\downarrow$)}} & \makecell{\textbf{Performance} \\ \textbf{Gain (\%)  ($\uparrow$)}} & \makecell{\textbf{Relative Training FLOPs} \\ \textbf{to Match Mamba (\%) ($\downarrow$)}} \\
    \midrule
    \multirow{4}{*}{37M} & \multirow{3}{*}{\textbf{\colorbox{SeaGreen!20}{Speech} Metrics}} & Training Loss          & 1.8159 & 1.6909 & \cellcolor{green!40}6.88\%  & \cellcolor{green!40}10.30\% \\
                         &                                      & LL60K Val. Loss        & 1.6756 & 1.5217 & \cellcolor{green!40}9.18\%  & \cellcolor{green!40}13.60\% \\
                         &                                      & PPL30K Val. Loss       & 1.8147 & 1.6845 & \cellcolor{green!40}7.17\%  & \cellcolor{green!40}13.60\% \\
    \cmidrule{2-7}
                         & \textbf{Overall Metrics}             & Avg Training Loss      & 4.2299 & 4.0759 & \cellcolor{green!30}3.64\%  & \cellcolor{green!30}45.00\% \\
    \midrule
    \multirow{4}{*}{94M} & \multirow{3}{*}{\textbf{\colorbox{SeaGreen!20}{Speech} Metrics}} & Training Loss          & 1.6911 & 1.5662 & \cellcolor{green!40}7.38\%  & \cellcolor{green!40}11.90\% \\
                         &                                      & LL60K Val. Loss        & 1.5235 & 1.3747 & \cellcolor{green!40}9.76\%  & \cellcolor{green!40}14.80\% \\
                         &                                      & PPL30K Val. Loss       & 1.6951 & 1.6152 & \cellcolor{green!30}4.71\%  & \cellcolor{green!40}12.60\% \\
    \cmidrule{2-7}
                         & \textbf{Overall Metrics}             & Avg Training Loss      & 3.7756 & 3.6371 & \cellcolor{green!30}3.67\%  & \cellcolor{green!30}43.10\% \\
    \midrule
    \multirow{4}{*}{443M} & \multirow{3}{*}{\textbf{\colorbox{SeaGreen!20}{Speech} Metrics}} & Training Loss          & 1.5414 & 1.4313 & \cellcolor{green!40}7.14\%  & \cellcolor{green!35}19.20\% \\
                          &                                      & LL60K Val. Loss        & 1.3466 & 1.2113 & \cellcolor{green!40}10.05\% & \cellcolor{green!35}24.70\% \\
                          &                                      & PPL30K Val. Loss       & 1.5634 & 1.4790 & \cellcolor{green!30}5.40\%  & \cellcolor{green!35}22.00\% \\
    \cmidrule{2-7}
                          & \textbf{Overall Metrics}             & Avg Training Loss      & 3.3317 & 3.2096 & \cellcolor{green!30}3.66\%  & \cellcolor{green!30}44.00\% \\
    \midrule
    \multirow{4}{*}{880M} & \multirow{3}{*}{\textbf{\colorbox{SeaGreen!20}{Speech} Metrics}} & Training Loss          & 1.4902 & 1.4054 & \cellcolor{green!40}5.69\%  & \cellcolor{green!30}22.40\% \\
                          &                                      & LL60K Val. Loss        & 1.2939 & 1.1757 & \cellcolor{green!40}9.13\%  & \cellcolor{green!30}30.10\% \\
                          &                                      & PPL30K Val. Loss       & 1.5400 & 1.4619 & \cellcolor{green!30}5.07\%  & \cellcolor{green!30}24.30\% \\
    \cmidrule{2-7}
                          & \textbf{Overall Metrics}             & Avg Training Loss      & 3.2289 & 3.1571 & \cellcolor{green!20}2.22\%  & \cellcolor{green!30}54.30\% \\
    \midrule
    \multirow{4}{*}{1.5B} & \multirow{3}{*}{\textbf{\colorbox{SeaGreen!20}{Speech} Metrics}} & Training Loss          & 1.4790 & 1.3940 & \cellcolor{green!30}5.75\%  & \cellcolor{green!30}24.80\% \\
                          &                                      & LL60K Val. Loss        & 1.2592 & 1.1552 & \cellcolor{green!40}8.26\%  & \cellcolor{green!30}32.10\% \\
                          &                                      & PPL30K Val. Loss       & 1.5200 & 1.4387 & \cellcolor{green!30}5.35\%  & \cellcolor{green!30}27.60\% \\
    \cmidrule{2-7}
                          & \textbf{Overall Metrics}             & Avg Training Loss      & 3.1507 & 3.0545 & \cellcolor{green!20}3.05\%  & \cellcolor{green!30}56.20\% \\
    \bottomrule
    \end{NiceTabular}}
    \caption{\textbf{Training and validation metrics across model scales with three modalities: image, text, and speech.} This setting extends the Chameleon framework by incorporating \textbf{speech} alongside image and text, with all modalities represented as discrete tokens. Mixture-of-Mamba achieves consistent improvements over Mamba Dense across all scales (\textbf{37M}, \textbf{94M}, \textbf{443M}, \textbf{880M}, and \textbf{1.5B}), particularly in the \textbf{speech modality}, where performance gains reach up to \textbf{9.18\%}. These gains are achieved with substantial reductions in training FLOPs, ranging from \textbf{10.30\%} to \textbf{56.20\%} relative to Mamba Dense. The results demonstrate that Mixture-of-Mamba generalizes effectively to a multi-modal setting with three modalities while delivering significant computational efficiency.}
    \label{tab:speech_chameleon_metrics}
\end{table*}

\begin{table}[ht]
    \centering
    \small 
    \resizebox{\linewidth}{!}{
    \begin{NiceTabular}{lcc}[colortbl-like]
    \toprule
    \textbf{Ablation Study} &  \makecell{\textbf{Avg Training} \\ \textbf{Loss ($\downarrow$)}} & \makecell{\textbf{Performance} \\ \textbf{Gain (\%) ($\uparrow$)}} \\
    \midrule
    {\footnotesize {\textbf{443M} \colorbox{cyan!20}{Mamba} (without \ding{202}\ding{203}\ding{204}\ding{205})}}                    & 3.3317 & 0\% (baseline) \\
    \ding{202} 
    {\footnotesize (decouple  $ \displaystyle W_{in\_proj}$)} & 3.2916 & \cellcolor{green!20} 1.22\% \\
    \ding{203} 
    {\footnotesize (decouple  $ \displaystyle W_{x\_proj}$)} & 3.3580 &  -0.79\% \\
    \ding{204} 
    {\footnotesize (decouple  $ \displaystyle W_{dt\_proj}$)} & 3.3525 & -0.62\% \\
    \ding{205} 
    {\footnotesize (decouple  $ \displaystyle W_{out\_proj}$)} & 3.3109 & \cellcolor{green!10} 0.63\% \\
    \ding{202}+\ding{203} 
    {\footnotesize (decouple $ \displaystyle W_{in\_proj}, W_{x\_proj}$)} & 3.2780 & \cellcolor{green!30} 1.64\% \\
    \ding{202}+\ding{204} 
    {\footnotesize (decouple $ \displaystyle W_{in\_proj}, W_{dt\_proj}$)} & 3.2687 & \cellcolor{green!35} 1.93\% \\
    \ding{202}+\ding{205} 
    {\footnotesize (decouple $ \displaystyle W_{in\_proj}, W_{out\_proj}$)} & 3.2599 & \cellcolor{green!40} 2.20\% \\
    \ding{203}+\ding{204} 
    {\footnotesize (decouple $ \displaystyle W_{x\_proj}, W_{dt\_proj}$)} & 3.3214 & \cellcolor{green!5} 0.31\% \\
    \ding{203}+\ding{205} 
    {\footnotesize (decouple $ \displaystyle W_{x\_proj}, W_{out\_proj}$)} & 3.2829 & \cellcolor{green!25} 1.49\% \\
    \ding{204}+\ding{205} 
    {\footnotesize (decouple $ \displaystyle W_{dt\_proj}, W_{out\_proj}$)} & 3.2509 & \cellcolor{green!45} 2.48\% \\
    \ding{202}+\ding{203}+\ding{204} 
    {\footnotesize (not decoupling $ \displaystyle W_{out\_proj}$)} & 3.2593 & \cellcolor{green!40} 2.22\% \\
    \ding{202}+\ding{203}+\ding{205} 
    {\footnotesize (not decoupling $ \displaystyle W_{dt\_proj}$)} & 3.2312 & \cellcolor{green!50} 3.11\% \\
    \ding{202}+\ding{204}+\ding{205} 
    {\footnotesize (not decoupling $ \displaystyle W_{x\_proj}$)} & 3.2342 & \cellcolor{green!50} 3.01\% \\
    \ding{203}+\ding{204}+\ding{205} 
    {\footnotesize (not decoupling $ \displaystyle W_{in\_proj}$)} & 3.2773 & \cellcolor{green!30} 1.66\% \\
    \ding{202}+\ding{203}+\ding{204}+\ding{205} 
    {\footnotesize \colorbox{Dandelion!50}{(Mixture-of-Mamba)}} 
     & \textbf{3.2096} 
     & \cellcolor{green!60}\textbf{3.80\%} \\
    \bottomrule
    \end{NiceTabular}}
    \caption{\textbf{Ablation study on the Chameleon + Speech setting.} This study evaluates the impact of decoupling individual components (\textbf{1}, \textbf{2}, \textbf{3}, \textbf{4}) and their combinations on model performance. The results demonstrate that decoupling all components (\textbf{1+2+3+4}, Mixture-of-Mamba) achieves the best performance with a \textbf{3.80\%} gain over the Mamba baseline. Notably, the performance gain achieved by decoupling all components together exceeds the sum of gains from decoupling each component individually, highlighting the synergistic effect of combined decoupling. Green shading indicates positive performance gains, with the darkest green highlighting the best configuration.}
    \label{tab:ablation_study}
\end{table}

\section{Results}

\subsection{Results in Multi-objective Training (Transfusion)}

We evaluate Mixture-of-Mamba (MoM) against Mamba Dense and Flex-Attention Transformer in the \textbf{Transfusion setting}, where pretraining is performed on interleaved text and image data across three model scales: \textbf{163M}, \textbf{760M}, and \textbf{1.4B}. 
See our training configuration in Appendix Table~\ref{tab:transfusion_model_config}. 
For clarity, performance gain is quantified as:
\[
\text{Performance Gain (\%)} = \frac{\text{Loss}_{\text{Dense}} - \text{Loss}_{\text{Mixture}}}{\text{Loss}_{\text{Dense}}} \times 100,
\]
where $\text{Loss}_{\text{Dense}}$ and $\text{Loss}_{\text{Mixture}}$ are the final losses of Mamba Dense and Mixture-of-Mamba, respectively. Relative training FLOPs reflect the computational cost required for MoM to match the training dynamics (similar loss) of Mamba Dense. The detailed results are summarized in Table~\ref{tab:transfusion_metrics} and Figure~\ref{fig:transfusion_training}, with further visualizations provided in Appendix Figures~\ref{fig:appendix_text_val}, \ref{fig:appendix_image_val_cc12m}, and \ref{fig:appendix_training_loss}.\footnote{%
Flex-Attention Transformer (i.e., Transfusion~\cite{transfusion}) combines both attention patterns by applying causal attention to every element in the sequence and \textit{bidirectional attention within the elements of each individual image}. This makes Flex-Attention Transformer an overestimated baseline for transformers because both Mamba and Mixture-of-Mamba are strictly causal across all elements, while Flex-Attention Transformer benefits from bidirectional attention within images.
}

\paragraph{Image Modality.} Mixture-of-Mamba (MoM) consistently demonstrates superior performance in \textbf{image modality training loss} across all model scales. At the \textbf{1.4B} scale, MoM achieves a training loss of \textbf{0.2138}, outperforming Mamba Dense by \textbf{2.20\%} while requiring only \textbf{34.76\%} of the training FLOPs. Similar trends are observed at smaller scales: at the \textbf{760M} scale, MoM achieves a training loss of \textbf{0.2172}, a \textbf{2.37\%} improvement over Mamba Dense, while reducing training FLOPs to \textbf{37.76\%}.

The validation loss curves on the CC12M dataset ((Table~\ref{tab:transfusion_metrics}, Appendix Figure~\ref{fig:appendix_image_val_cc12m}) further illustrate these trends. Mixture-of-Mamba consistently achieves lower image validation loss compared to Mamba Dense and Flex-Attention Transformer, with the improvements becoming more pronounced as model size increases. Additionally, loss matching curves demonstrate that MoM reaches equivalent loss values at earlier training steps, highlighting its improved training efficiency.

\paragraph{Text Modality.} In the text modality, Mixture-of-Mamba consistently outperforms Mamba Dense across both training and validation metrics. At the \textbf{1.4B} scale, MoM achieves lower validation losses on both the C4 (\textbf{2.2695}) and Wikipedia (\textbf{1.7164}) datasets compared to Mamba Dense, despite their similar training losses. This indicates better generalization to unseen text data. Importantly, MoM also performs comparably to or better than Flex-Attention Transformer, particularly on validation losses, as shown in Appendix Figure~\ref{fig:appendix_text_val}. Similar trends are observed at smaller scales (\textbf{760M} and \textbf{163M}), where MoM reduces validation losses while maintaining high training efficiency.

Loss matching results in Appendix Figure~\ref{fig:appendix_text_val} (b, f, j) confirm that Mixture-of-Mamba aligns closely with or surpasses Mamba Dense, reaching comparable loss values earlier during training. These improvements highlight MoM's strong performance in text tasks while maintaining its computational efficiency.

\paragraph{Overall Performance and Efficiency.} Across both image and text modalities, Mixture-of-Mamba consistently outperforms Mamba Dense in terms of loss reduction while requiring significantly fewer training FLOPs to achieve similar learning dynamics. At the \textbf{1.4B} scale, MoM improves the overall training loss by \textbf{0.84\%} while requiring only \textbf{83.10\%} of the training FLOPs. At smaller scales, such as \textbf{760M} and \textbf{163M}, MoM reduces the overall training loss by up to \textbf{0.94\%}, while requiring just \textbf{82.94\%} and \textbf{86.11\%} of the FLOPs, respectively (Table~\ref{tab:transfusion_metrics}, Appendix Figure~\ref{fig:appendix_training_loss}).
These results, summarized in Table~\ref{tab:transfusion_metrics} and Figure~\ref{fig:transfusion_training}, and further supported by Appendix Figures~\ref{fig:appendix_text_val}, \ref{fig:appendix_image_val_cc12m}, and \ref{fig:appendix_training_loss}, underscoring MoM’s effectiveness, scalability, and efficiency in the \textbf{Transfusion setting}.

\subsection{Results in Training with Uniform Representations (Chameleon)}

We evaluate Mixture-of-Mamba (MoM) in the \textbf{Chameleon setting}, where both \textbf{image} and \textbf{text} modalities are represented as discrete tokens. See our training configuration in Appendix Table~\ref{tab:chameleon_model_config}. Results are summarized in Table~\ref{tab:chameleon_metrics}, with full results across all five scales (\textbf{37M}, \textbf{94M}, \textbf{443M}, \textbf{880M}, and \textbf{1.5B}) provided in Appendix Table~\ref{tab:full_chameleon_metrics}. Training dynamics and validation loss trends are visualized in Appendix Figures~\ref{fig:chameleon_training}, \ref{fig:appendix_chameleon_obelisc}, and \ref{fig:appendix_chameleon_shutterstock}.

\paragraph{Image Modality.} Mixture-of-Mamba (MoM) consistently demonstrates better performance in \textbf{image modality training loss} across all model scales, achieving substantial efficiency gains over Mamba Dense. At the \textbf{443M} scale, MoM achieves a training loss of \textbf{5.1703}, a \textbf{3.46\%} improvement over Mamba Dense, while requiring only \textbf{33.40\%} of the training FLOPs. Similar trends are observed at other scales: at the largest \textbf{1.5B} scale, MoM achieves a training loss of \textbf{5.0591}, a \textbf{2.51\%} improvement, with only \textbf{42.50\%} of the training FLOPs. At the smallest \textbf{37M} scale, MoM reduces training loss to \textbf{5.9561}, outperforming Mamba Dense by \textbf{2.85\%} while requiring just \textbf{25.90\%} of the FLOPs (Appendix Table~\ref{tab:full_chameleon_metrics}). These results highlight MoM's ability to achieve improved performance and convergence efficiency consistently in the image modality across all model scales.

\paragraph{Text Modality.} Mixture-of-Mamba (MoM) demonstrates consistent improvements in \textbf{text modality training loss} across all model scales. At the largest \textbf{1.5B} scale, MoM reduces training loss to \textbf{2.1614}, a \textbf{3.01\%} improvement over Mamba Dense, while requiring only \textbf{65.40\%} of the training FLOPs. Validation loss on Obelisc and a proprietary version of the Shutterstock datasets (SSTK) exhibits similar trends, with MoM achieving notable improvements in loss values while maintaining significant efficiency gains (Appendix Figures~\ref{fig:appendix_chameleon_obelisc} and \ref{fig:appendix_chameleon_shutterstock}). These results further highlight MoM's ability to deliver strong text performance with improved convergence efficiency.
These results highlight Mixture-of-Mamba’s robust and efficient improvements in the Chameleon setting across both image and text modalities, with substantial computational savings.

\subsection{Results in Training with Three Modalities
(Chameleon+Speech)
}

To evaluate the robustness and scalability of Mixture-of-Mamba (MoM), we extend the Chameleon framework to include a third modality: \textbf{speech}, alongside image and text, with all modalities represented as discrete tokens. Speech data is tokenized using an in-house tokenizer, a variant of DinoSR~\citep{liu2024dinosrselfdistillationonlineclustering}, which extracts semantic tokens with a vocabulary size of 500, where each token corresponds to 40ms of audio content. Results are summarized in Table~\ref{tab:speech_chameleon_metrics}, with additional training dynamics and evaluation loss trends visualized in Appendix Figures~\ref{fig:appendix_chameleon_speech_part1}, \ref{fig:appendix_chameleon_speech_part2}, \ref{fig:appendix_chameleon_speech_obelisc}, and \ref{fig:appendix_chameleon_speech_Shutterstock}.

\paragraph{Speech Modality.} Mixture-of-Mamba (MoM) achieves substantial improvements in \textbf{speech modality training loss} across all model scales. At the \textbf{443M} scale, MoM improves speech training loss by \textbf{7.14\%} compared to Mamba Dense. To match the training loss achieved by Mamba Dense, MoM requires only \textbf{19.20\%} of the training FLOPs, demonstrating significant efficiency gains. Similar trends hold at the largest \textbf{1.5B} scale, where MoM achieves a \textbf{5.75\%} improvement in speech training loss and matches Mamba Dense's loss with just \textbf{24.80\%} of the training FLOPs.

\textbf{Overall training loss} is consistently reduced across scales. At the \textbf{1.5B} scale, MoM lowers the overall training loss by \textbf{3.05\%}. When targeting the same loss as Mamba Dense, MoM achieves this with a \textbf{56.20\%} reduction in relative training FLOPs, highlighting its improved computational efficiency.

Performance in the \textbf{image} and \textbf{text} modalities similarly shows consistent improvements in training and validation losses relative to Mamba Dense. Full results and trends are presented in Appendix Figures~\ref{fig:appendix_chameleon_speech_obelisc} and \ref{fig:appendix_chameleon_speech_Shutterstock}, where MoM's robust performance across all three modalities is further validated.

\subsection{Ablation Study on Decoupling Components}

To better understand the design choices underpinning Mixture-of-Mamba, we conduct an ablation study on the \textbf{Chameleon + Speech setting} at the \textbf{443M} scale. We evaluate the impact of decoupling four key components—\( W_{\text{in-proj}} \) (\ding{202}), \( W_{\text{x-proj}} \) (\ding{203}), \( W_{\text{dt-proj}} \) (\ding{204}), and \( W_{\text{out-proj}} \) (\ding{205})—individually and in various combinations. This analysis enables us to test both individual and combined contributions to the model's overall performance.

The results show that decoupling components individually yields varying degrees of improvement, with performance gains ranging from \textbf{0.63\%} (\( W_{\text{out-proj}} \)) to \textbf{1.22\%} (\( W_{\text{in-proj}} \)). Interestingly, some components (\( W_{\text{x-proj}} \) and \( W_{\text{dt-proj}} \)) exhibit minimal or even slightly negative impact when decoupled alone. However, decoupling multiple components in combination leads to significantly larger gains. For example, decoupling \( W_{\text{in-proj}} \) and \( W_{\text{out-proj}} \) (\ding{202}+\ding{205}) achieves a \textbf{2.20\%} improvement, while decoupling three components (\ding{202}+\ding{203}+\ding{205}) further increases the gain to \textbf{3.11\%}.

Most importantly, decoupling all four components simultaneously (\ding{202}+\ding{203}+\ding{204}+\ding{205}, Mixture-of-Mamba) achieves the largest improvement, with a performance gain of \textbf{3.80\%} over the Mamba baseline. This result highlights a key observation: the gain from decoupling all components together exceeds the sum of individual gains, demonstrating a synergistic effect. The combination of all decoupled projections enables better parameter allocation across modalities, leading to more efficient and effective learning.
In summary, the ablation study confirms that the design of Mixture-of-Mamba is both effective and interdependent. Decoupling all key components simultaneously is important to achieving the observed substantial performance gains.

\section{Related Work}

\subsection{State-Space Models and Multi-Modal Extensions}

State-space models (SSMs)~\citep{gu2021efficiently, gu2023mamba} have recently gained traction as computationally efficient alternatives to Transformers for sequential modeling. Mamba~\citep{gu2023mamba}, in particular, demonstrates strong performance on single-modality tasks by leveraging linear time complexity and advanced gating mechanisms. Extending Mamba to multi-modal tasks remains an active research area.

In vision-language modeling, VLMamba~\citep{qiao2024vl} and Cobra~\citep{zhao2024cobra} augment Mamba by incorporating LLaVA-style projection modules, enabling image features to be mapped into the token space of the Mamba model for sequence modeling. In the vision domain, Vision Mamba~\citep{zhu2024vision} introduces bidirectional scanning by chaining forward and backward SSM blocks, while VMamba~\citep{liu2024vmamba} further enhances image patch processing with a 2D Selective Scan (SS2D) module that traverses patches across multiple scanning paths.

For diffusion-based models, works such as DiffuSSM~\citep{yan2024diffusion} and Zigma~\citep{hu2024zigma} replace attention mechanisms with SSMs for image and video generation. Zigma introduces a zigzag scanning scheme to improve efficiency for sequential diffusion tasks, while other approaches~\citep{mo2024scaling, fei2024scalable} explore bi-directional SSM architectures. While these works highlight the flexibility of Mamba in generative tasks, they focus primarily on architectural modifications for specific domains rather than general multi-modal pretraining.

The most related work to ours is MoE-Mamba~\citep{pióro2024moemamba} and Blackmamba~\citep{anthony2024blackmamba}, which interleave Mamba blocks with MoE-augmented MLPs to introduce sparsity. However, these hybrid designs apply sparsity only to the MLP layers, leaving the dense Mamba block unmodified. In contrast, our proposed Mixture-of-Mamba integrates modality-aware sparsity directly into the Mamba block by decoupling its projection components, enabling specialized computations for different modalities. This general design complements existing methods and offers new opportunities for computationally efficient multi-modal pretraining.

\subsection{Sparse Architectures for Multi-Modal Pretraining}

Model sparsity, particularly Mixture-of-Experts (MoE), has been extensively explored in Transformers to reduce computational cost~\citep{jacobs1991adaptive, eigen2013learning, ShazeerMoE, gshard, switchtransformer, mixtral}. MoE selectively activates subsets of parameters for each input token, allowing the model to specialize in different aspects of the data. However, challenges such as expert imbalance, bi-level optimization, and load balancing remain prevalent~\citep{ShazeerMoE, gshard, nasbench360}.

In multi-modal tasks, modality-aware sparsity has emerged as an effective strategy. Works such as VLMo~\citep{shen2023vlmoe}, MoMA~\citep{lin2024moma}, and related approaches~\citep{imageasaforeignlanguage, shen2022dash,bao2022vlmo, long2023multiway, shen2025cat} assign modality-specific experts to handle the unique statistical properties of text, images, and other data types. This improves specialization while avoiding the complexities of learned routing mechanisms~\citep{liang2022mind}.

Transformer-based architectures have further extended sparsity into attention mechanisms~\citep{wang2023cogvlm, llmtags,shen2024jetmoe, liu2024playgroundv3improvingtexttoimage,shen2024scribeagent}. CogVLM~\citep{wang2023cogvlm} applies sparse techniques on top of a pre-trained Vicuna-7B model but remains limited to generating text outputs. Concurrently, Playground v3 (PGv3)~\citep{liu2024playgroundv3improvingtexttoimage} integrates DiT-style image transformers with a frozen LLaMA-3 backbone to achieve state-of-the-art performance in text-to-image generation.

Our work differs fundamentally in two key aspects. First, Mixture-of-Mamba introduces \textit{modality-aware sparsity} into the Mamba block itself, generalizing sparse architectures beyond Transformers to SSMs. Unlike prior works that sparsify only the MLP or attention components, we decouple projection components of the Mamba block, enabling efficient and specialized computations across modalities. Second, Mixture-of-Mamba is trained from scratch for multi-modal generation tasks, unlike approaches like CogVLM and PGv3 that fine-tune pre-trained backbones.

Furthermore, our design is complementary to existing MoE techniques. Prior work~\citep{liang2024mixture} has demonstrated that MoE-based sparsification can be combined with sparse architectures like Mixture-of-Transformers to achieve additional gains. Similarly, Mixture-of-Mamba can serve as a versatile and computationally efficient solution, offering new pathways for scalable multi-modal pretraining.

\section{Conclusion}
In this work, we introduced \textbf{Mixture-of-Mamba}, a novel extension of state-space models (SSMs) that incorporates \textit{modality-aware sparsity} through modality-specific parameterization. By enabling modality-specific specialization while preserving the computational efficiency of SSMs, Mixture-of-Mamba consistently outperforms dense baselines across three multi-modal settings: Transfusion (interleaved text and continuous image tokens), Chameleon (interleaved text and discrete image tokens), and an extended Chameleon+Speech framework. Our results demonstrate substantial improvements in loss reduction, with training efficiency gains reaching more than \textbf{double} the computational efficiency compared to dense SSMs. Ablation studies further reveal a synergistic effect from jointly decoupling key projection components, highlighting the effectiveness of modality-aware sparsity. These findings establish Mixture-of-Mamba as a scalable and efficient architecture for multi-modal pretraining, paving the way for future exploration in dynamic sparsity and broader multi-modal applications.

\section*{Impact Statement}

This work introduces efficiency improvements in multi-modal machine learning systems through modality-aware sparsity techniques. The primary impact is computational efficiency - Mixture-of-Mamba reduces computational costs by up to 65\% while maintaining or improving performance. This has positive environmental implications through reduced energy consumption and democratizes access to multi-modal AI systems by lowering computational resource requirements. While these advances could enable beneficial applications in education, accessibility, and human-computer interaction, we acknowledge they could also facilitate potentially concerning applications. We encourage the research community to consider appropriate guidelines for responsible deployment of such technologies.

\nocite{langley00}

\newpage
\clearpage

\bibliography{paper}
\bibliographystyle{icml2025}

\newpage
\appendix
\onecolumn

\begin{table*}[ht]
    \centering
    \resizebox{\textwidth}{!}{
    \begin{NiceTabular}{llcccccc}[colortbl-like]
    \toprule
    \makecell{\textbf{Model} \\ \textbf{Scale}} & 
    \makecell{\textbf{Metric} \\ \textbf{Category}}
    & \textbf{Metric Name} & 
    \makecell{\colorbox{cyan!20}{\textbf{Mamba}} \\ \textbf{Loss ($\downarrow$)}}
    & 
    \makecell{
    \colorbox{black!20}{\textbf{Flex-Attention}} \\ 
    \colorbox{black!20}{\textbf{Transformer}} \\ \textbf{Loss ($\downarrow$)}}     
    & 
    \makecell{\colorbox{Dandelion!50}{\textbf{Mixture-of-}} \\ \textbf{\colorbox{Dandelion!50}{Mamba}} \\
    \textbf{Loss ($\downarrow$)}
    } 
    & \makecell{
        \textbf{Performance} \\
        \textbf{Gain over} \\
        \textbf{Mamba (\%) ($\uparrow$)}} 
    & 
    \makecell{\textbf{Relative Training} \\
    \textbf{FLOPs to Match}  \\
    \textbf{Mamba (\%)} \textbf{($\downarrow$)}} \\
    
    \midrule
    \multirow{8}{*}{163M} & \multirow{2}{*}{\textbf{\colorbox{blue!20}{Image} Metrics}} & Training Loss    & 0.2262 & 0.2250 & \textit{\textcolor{gray}{0.2199}} &  \cellcolor{green!30}2.80\% 
    & 
    \cellcolor{green!40}49.21\% \\
                         &                                 & CC12M Val. Loss        & 0.2295 & 0.2293 & \textit{\textcolor{gray}{0.2255}} & \cellcolor{green!20}1.74\% & \cellcolor{green!40}50.61\% \\
    \cmidrule{2-8}
                         & \multirow{3}{*}{\textbf{\colorbox{OrangeRed!20}{Text} Metrics}}  & Avg Training Loss      & 2.4702 & \textit{\textcolor{gray}{2.4424}} & 2.4690 & \cellcolor{green!10}0.05\% & \cellcolor{green!20}98.80\% \\
                         &                                 & C4 Val. Loss           & 2.6917 & \textit{\textcolor{gray}{2.6862}} & 2.6912 & \cellcolor{green!10}0.02\% & \cellcolor{green!20}99.88\% \\
                         &                                 & Wikipedia Val. Loss    & 2.1884 & \textit{\textcolor{gray}{2.1715}} & 2.1870 & \cellcolor{green!10}0.06\% & \cellcolor{green!20}99.81\% \\
    \cmidrule{2-8}
                         & \textbf{Overall}               & Train Avg Loss         & 3.6014 & \textit{\textcolor{gray}{3.5674}} & 3.5685 & \cellcolor{green!20}0.91\% & \cellcolor{green!30}86.11\% \\
    \midrule
    \multirow{8}{*}{760M} & \multirow{2}{*}{\textbf{\colorbox{blue!20}{Image} Metrics}} & Training Loss    & 0.2225 & 0.2213 & \textit{\textcolor{gray}{0.2172}} & \cellcolor{green!30}2.37\% & \cellcolor{green!40}37.76\% \\
                         &                                 & CC12M Val. Loss        & 0.2272 & 0.2253 & \textit{\textcolor{gray}{0.2201}} & \cellcolor{green!30}3.13\% & \cellcolor{green!40}35.27\% \\
    \cmidrule{2-8}
                         & \multirow{3}{*}{\textbf{\colorbox{OrangeRed!20}{Text} Metrics}}  & Avg Training Loss      & 2.1394 & \textit{\textcolor{gray}{2.1253}} & 2.1353 & \cellcolor{green!10}0.19\% & \cellcolor{green!20}96.82\% \\
                         &                                 & C4 Val. Loss           & 2.3593 & \textit{\textcolor{gray}{2.3559}} & 2.3555 & \cellcolor{green!10}0.16\% & \cellcolor{green!20}99.01\% \\
                         &                                 & Wikipedia Val. Loss    & 1.8191 & \textit{\textcolor{gray}{1.8143}} & 1.8149 & \cellcolor{green!10}0.23\% & \cellcolor{green!20}99.11\% \\
    \cmidrule{2-8}
                         & \textbf{Overall}               & Train Avg Loss         & 3.2519 & 3.2318 & \textit{\textcolor{gray}{3.2214}} & \cellcolor{green!20}0.94\% & \cellcolor{green!30}82.94\% \\
    \midrule
    \multirow{8}{*}{1.4B} & \multirow{2}{*}{\textbf{\colorbox{blue!20}{Image} Metrics}} & Training Loss    & 0.2186 & 0.2221 & \textit{\textcolor{gray}{0.2138}} & \cellcolor{green!30}2.20\% & \cellcolor{green!40}34.76\% \\
                         &                                 & CC12M Val. Loss        & 0.2264 & 0.2247 & \textit{\textcolor{gray}{0.2190}} & \cellcolor{green!30}3.29\% & \cellcolor{green!40}36.15\% \\
    \cmidrule{2-8}
                         & \multirow{3}{*}{\textbf{\colorbox{OrangeRed!20}{Text} Metrics}}  & Avg Training Loss      & 2.0761 & \textit{\textcolor{gray}{2.0673}} & 2.0737 & \cellcolor{green!10}0.12\% & \cellcolor{green!20}98.27\% \\
                         &                                 & C4 Val. Loss           & 2.2726 & 2.2728 & \textit{\textcolor{gray}{2.2695}} & \cellcolor{green!10}0.13\% & \cellcolor{green!20}99.34\% \\
                         &                                 & Wikipedia Val. Loss    & 1.7205 & 1.7218 & \textit{\textcolor{gray}{1.7164}} & \cellcolor{green!10}0.24\% & \cellcolor{green!20}99.30\% \\
    \cmidrule{2-8}
                         & \textbf{Overall}               & Train Avg Loss         & 3.1693 & 3.1777 & \textit{\textcolor{gray}{3.1429}} & \cellcolor{green!20}0.84\% & \cellcolor{green!30}83.10\% \\
    \bottomrule
    \end{NiceTabular}}
    \caption{\textbf{Training and validation metrics across model scales in the Transfusion setting.} Loss values are reported for image and text modalities at three model sizes: \textbf{163M}, \textbf{760M}, and \textbf{1.4B}. Mixture-of-Mamba consistently achieves competitive or superior performance in image metrics and maintains strong text performance compared to Mamba Dense and Flex-Attention Transformer. The table also reports relative training FLOPs required for Mixture-of-Mamba and Flex-Attention Transformer to match Mamba's training dynamics, highlighting improved training efficiency. Best loss values in each row are highlighted.}
    \label{tab:transfusion_metrics}
\end{table*}

\begin{table*}[ht]
    \centering
    \resizebox{\textwidth}{!}{
    \begin{NiceTabular}{lcccccccc}[colortbl-like]
    \toprule
    \textbf{Model Size} & \textbf{Hidden Dim.} & \textbf{Layers} & \textbf{Heads} & \textbf{Seq. Length} & \textbf{Batch Size/GPU} & \textbf{GPUs} & \textbf{Tokens/Batch} & \textbf{Steps} \\
    \midrule
    163M  & 768 & 16  & 12  & 4,096 & 4   & 56  & 1,048,576 & 250,000  \\
    760M  & 1,536 & 24  & 24  & 4,096 & 4   & 56 & 1,048,576 & 250,000  \\
    1.4B  & 2,048 & 24  & 16  & 4,096 & 2   & 128 & 1,048,576 & 250,000  \\
    \bottomrule
    \end{NiceTabular}
    }
    \caption{\textbf{Architectural specifications and training configurations of models across different parameter scales (Transfusion setting).} 
    }
    \label{tab:transfusion_model_config}
\end{table*}

\begin{table*}[ht]
    \centering
    \resizebox{\textwidth}{!}{
    \begin{NiceTabular}{lcccccccc}[colortbl-like]
    \toprule
    \textbf{Model Size} & \textbf{Hidden Dim.} & \textbf{Layers} & \textbf{Heads} & \textbf{Seq. Length} & \textbf{Batch Size/GPU} & \textbf{GPUs} & \textbf{Tokens/Batch} & \textbf{Steps} \\
    \midrule
    37M   & 256  & 4   & 8   & 4,096 & 2  & 64  & 524,288 & 160,000  \\
    94M   & 512  & 8   & 8   & 4,096 & 2   & 64  & 524,288 & 160,000  \\
    443M  & 1,024 & 24  & 16  & 4,096 & 2   & 64  & 524,288 & 160,000  \\
    880M  & 1,536 & 24  & 24  & 4,096 & 2   & 64 & 524,288 & 120,000  \\
    1.5B  & 2,048 & 24  & 16  & 4,096 & 1   & 128 & 524,288 & 120,000  \\
    \bottomrule
    \end{NiceTabular}
    }
    \caption{\textbf{Architectural specifications and training configurations of models across different parameter scales (Chameleon setting and Chameleon+Speech setting).} 
    }
    \label{tab:chameleon_model_config}
\end{table*}

\begin{table*}[ht]
    \centering
    \resizebox{\textwidth}{!}{
    \begin{NiceTabular}{llccccc}[colortbl-like]
    \toprule
    \makecell{\textbf{Model} \\ \textbf{Scale}} & \textbf{Metric Category} & \textbf{Metric Name} & 
    \makecell{\colorbox{cyan!20}{\textbf{Mamba}} \\ \textbf{Loss ($\downarrow$)}}
    & \makecell{\colorbox{Dandelion!50}{\textbf{Mixture-of-}} \\ \textbf{\colorbox{Dandelion!50}{Mamba} Loss ($\downarrow$)}} & \makecell{\textbf{Performance} \\ \textbf{Gain (\%)  ($\uparrow$)}} & \makecell{\textbf{Relative Training FLOPs} \\ \textbf{to Match Mamba (\%) ($\downarrow$)}} \\
    \midrule
    \multirow{7}{*}{37M} & \multirow{3}{*}{\textbf{\colorbox{blue!20}{Image} Metrics}} & Training Loss          & 6.1308 & 5.9561 & \cellcolor{green!30}2.85\%  & \cellcolor{green!40}25.90\% \\
                         &                                      & Obelisc Val. Loss      & 5.2866 & 5.1124 & \cellcolor{green!35}3.29\%  & \cellcolor{green!40}26.60\% \\
                         &                                      & SSTK Val. Loss         & 6.6694 & 6.5023 & \cellcolor{green!30}2.51\%  & \cellcolor{green!40}27.50\% \\
    \cmidrule{2-7}
                         & \multirow{3}{*}{\textbf{\colorbox{OrangeRed!20}{Text} Metrics}} & Training Loss          & 3.6262 & 3.5175 & \cellcolor{green!30}3.00\%  & \cellcolor{green!30}60.90\% \\
                         &                                      & Obelisc Val. Loss      & 4.1244 & 4.0469 & \cellcolor{green!20}1.88\%  & \cellcolor{green!20}64.80\% \\
                         &                                      & SSTK Val. Loss         & 4.0417 & 3.9533 & \cellcolor{green!20}2.19\%  & \cellcolor{green!30}57.50\% \\
    \cmidrule{2-7}
                         & \multirow{1}{*}{\textbf{Overall}}     & Avg Training Loss      & 4.6607 & 4.5247 & \cellcolor{green!30}2.92\%  & \cellcolor{green!30}50.70\% \\
    \midrule
    \multirow{7}{*}{94M} & \multirow{3}{*}{\textbf{\colorbox{blue!20}{Image} Metrics}} & Training Loss          & 5.7609 & 5.6057 & \cellcolor{green!30}2.69\%  & \cellcolor{green!35}35.70\% \\
                         &                                      & Obelisc Val. Loss      & 4.9231 & 4.7683 & \cellcolor{green!30}3.14\%  & \cellcolor{green!35}35.30\% \\
                         &                                      & SSTK Val. Loss         & 6.3130 & 6.1652 & \cellcolor{green!20}2.34\%  & \cellcolor{green!35}37.00\% \\
    \cmidrule{2-7}
                         & \multirow{3}{*}{\textbf{\colorbox{OrangeRed!20}{Text} Metrics}} & Training Loss          & 3.0294 & 2.9414 & \cellcolor{green!30}2.90\%  & \cellcolor{green!30}58.40\% \\
                         &                                      & Obelisc Val. Loss      & 3.6016 & 3.5270 & \cellcolor{green!20}2.07\%  & \cellcolor{green!20}62.60\% \\
                         &                                      & SSTK Val. Loss         & 3.4109 & 3.2901 & \cellcolor{green!35}3.54\%  & \cellcolor{green!30}61.40\% \\
    \cmidrule{2-7}
                         & \multirow{1}{*}{\textbf{Overall}}     & Avg Training Loss      & 4.1577 & 4.0419 & \cellcolor{green!30}2.78\%  & \cellcolor{green!30}49.80\% \\
    \midrule
    \multirow{7}{*}{443M} & \multirow{3}{*}{\textbf{\colorbox{blue!20}{Image} Metrics}} & Training Loss          & 5.3558 & 5.1703 & \cellcolor{green!35}3.46\%  & \cellcolor{green!35}33.40\% \\
                          &                                      & Obelisc Val. Loss      & 4.5258 & 4.3546 & \cellcolor{green!40}3.78\%  & \cellcolor{green!35}35.10\% \\
                          &                                      & SSTK Val. Loss         & 5.9179 & 5.7471 & \cellcolor{green!30}2.89\%  & \cellcolor{green!35}35.30\% \\
    \cmidrule{2-7}
                          & \multirow{3}{*}{\textbf{\colorbox{OrangeRed!20}{Text} Metrics}} & Training Loss          & 2.4637 & 2.3864 & \cellcolor{green!30}3.14\%  & \cellcolor{green!30}62.00\% \\
                          &                                      & Obelisc Val. Loss      & 3.0544 & 2.9820 & \cellcolor{green!20}2.37\%  & \cellcolor{green!20}66.70\% \\
                          &                                      & SSTK Val. Loss         & 2.7569 & 2.6250 & \cellcolor{green!40}4.78\%  & \cellcolor{green!30}54.70\% \\
    \cmidrule{2-7}
                          & \multirow{1}{*}{\textbf{Overall}}     & Avg Training Loss      & 3.6584 & 3.5364 & \cellcolor{green!35}3.33\%  & \cellcolor{green!30}47.90\% \\
    \midrule
    \multirow{7}{*}{880M} & \multirow{3}{*}{\textbf{\colorbox{blue!20}{Image} Metrics}} & Training Loss          & 5.2260 & 5.1201 & \cellcolor{green!10}2.03\%  & \cellcolor{green!30}48.40\% \\
                          &                                      & Obelisc Val. Loss      & 4.4127 & 4.3105 & \cellcolor{green!20}2.32\%  & \cellcolor{green!30}49.30\% \\
                          &                                      & SSTK Val. Loss         & 5.7987 & 5.6986 & \cellcolor{green!10}1.73\%  & \cellcolor{green!30}50.50\% \\
    \cmidrule{2-7}
                          & \multirow{3}{*}{\textbf{\colorbox{OrangeRed!20}{Text} Metrics}} & Training Loss          & 2.3073 & 2.2438 & \cellcolor{green!20}2.75\%  & \cellcolor{green!30}65.60\% \\
                          &                                      & Obelisc Val. Loss      & 2.8886 & 2.8313 & \cellcolor{green!10}1.99\%  & \cellcolor{green!10}72.80\% \\
                          &                                      & SSTK Val. Loss         & 2.5483 & 2.4548 & \cellcolor{green!30}3.67\%  & \cellcolor{green!20}67.90\% \\
    \cmidrule{2-7}
                          & \multirow{1}{*}{\textbf{Overall}}     & Avg Training Loss      & 3.5130 & 3.4320 & \cellcolor{green!20}2.31\%  & \cellcolor{green!30}58.30\% \\
    \midrule
    \multirow{7}{*}{1.5B} & \multirow{3}{*}{\textbf{\colorbox{blue!20}{Image} Metrics}} & Training Loss          & 5.1892 & 5.0591 & \cellcolor{green!20}2.51\%  & \cellcolor{green!30}42.50\% \\
                          &                                      & Obelisc Val. Loss      & 4.3692 & 4.2510 & \cellcolor{green!20}2.71\%  & \cellcolor{green!30}44.50\% \\
                          &                                      & SSTK Val. Loss         & 5.7546 & 5.6335 & \cellcolor{green!10}2.10\%  & \cellcolor{green!30}44.60\% \\
    \cmidrule{2-7}
                          & \multirow{3}{*}{\textbf{\colorbox{OrangeRed!20}{Text} Metrics}} & Training Loss          & 2.2284 & 2.1614 & \cellcolor{green!30}3.01\%  & \cellcolor{green!30}65.40\% \\
                          &                                      & Obelisc Val. Loss      & 2.8020 & 2.7393 & \cellcolor{green!20}2.24\%  & \cellcolor{green!10}71.60\% \\
                          &                                      & SSTK Val. Loss         & 2.4614 & 2.3455 & \cellcolor{green!35}4.71\%  & \cellcolor{green!20}62.10\% \\
    \cmidrule{2-7}
                          & \multirow{1}{*}{\textbf{Overall}}     & Avg Training Loss      & 3.4602 & 3.3670 & \cellcolor{green!20}2.69\%  & \cellcolor{green!30}54.70\% \\
    \bottomrule
    \end{NiceTabular}}
    \caption{\textbf{Training and validation metrics across model scales in the Chameleon setting.} In this setting, both image and text modalities are represented as discrete tokens. Mixture-of-Mamba achieves substantial performance improvements over Mamba Dense, with the \textbf{image modality} showing the largest gains across all five model scales: \textbf{37M}, \textbf{94M}, \textbf{443M}, \textbf{880M}, and \textbf{1.5B}. Notably, the \textbf{text modality} also exhibits significant improvements, in contrast to the Transfusion setting where text gains were more modest. These results further highlight the effectiveness and efficiency of Mixture-of-Mamba, which consistently achieves strong performance with reduced relative training FLOPs.}
    \label{tab:full_chameleon_metrics}
\end{table*}

\clearpage
\newpage

\begin{figure*}[t]
    \centering

   \begin{subfigure}[b]{0.24\textwidth}
        \centering
        \includegraphics[width=\textwidth]{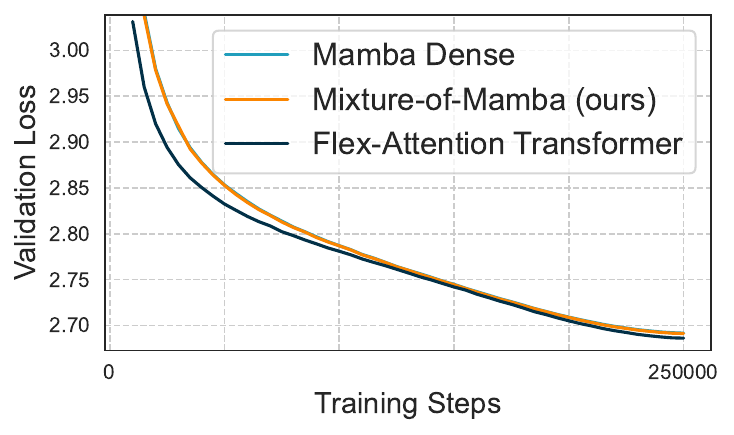}
        \caption{\textbf{163M} C4 Val. Loss}
    \end{subfigure}
    \hfill
    \begin{subfigure}[b]{0.24\textwidth}
       \centering
       \includegraphics[width=\textwidth]{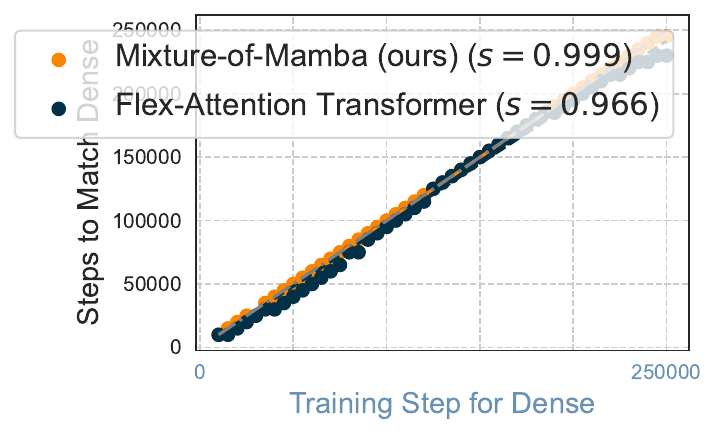}
       \caption{C4 Loss Matching}
   \end{subfigure}
   \hfill
   \begin{subfigure}[b]{0.24\textwidth}
        \centering
        \includegraphics[width=\textwidth]{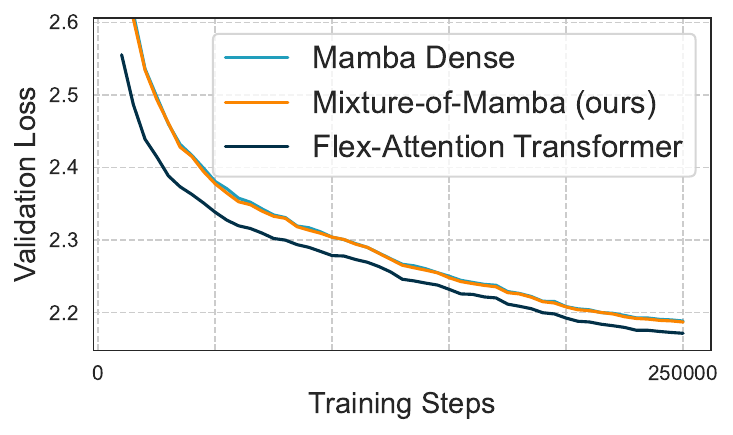}
        \caption{\textbf{163M} Wikipedia Val. Loss}
    \end{subfigure}
    \begin{subfigure}[b]{0.24\textwidth}
       \centering
       \includegraphics[width=\textwidth]{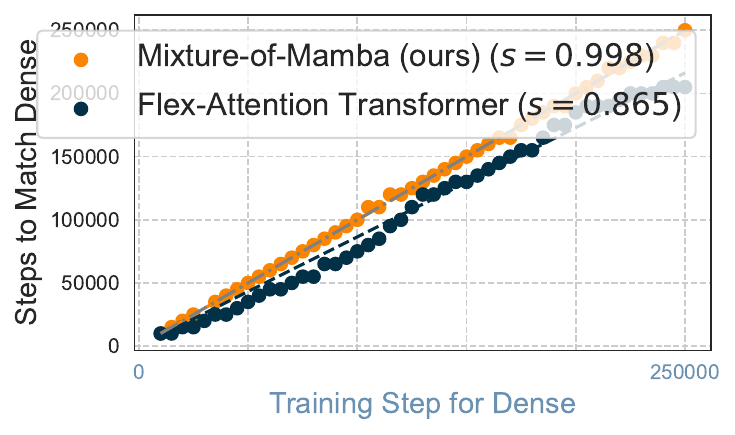}
       \caption{Wikipedia Loss Matching}
   \end{subfigure}

   \begin{subfigure}[b]{0.24\textwidth}
        \centering
        \includegraphics[width=\textwidth]{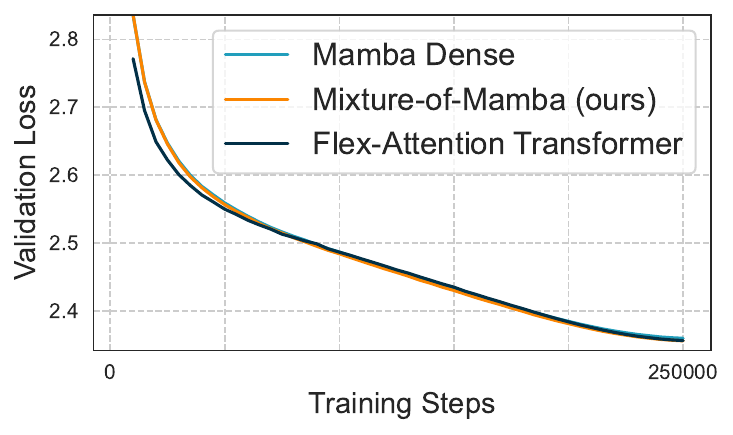}
        \caption{\textbf{760M} C4 Val. Loss}
    \end{subfigure}
    \hfill
    \begin{subfigure}[b]{0.24\textwidth}
    \centering
    \includegraphics[width=\textwidth]{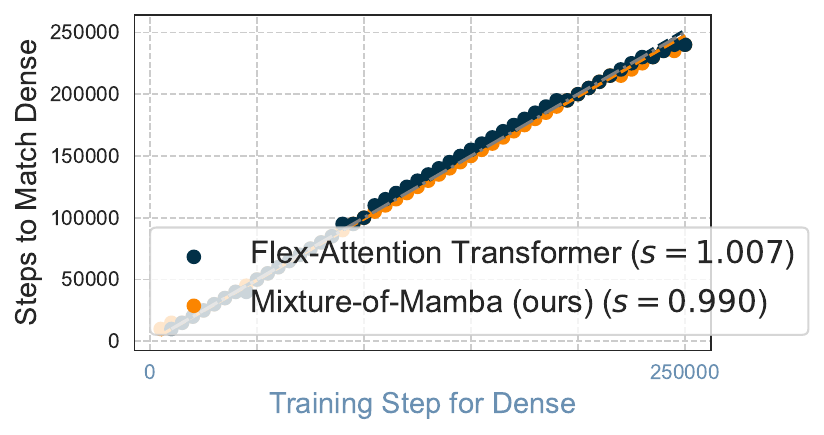}
    \caption{C4 Loss Matching}
    \end{subfigure}
    \hfill
    \begin{subfigure}[b]{0.24\textwidth}
        \centering
        \includegraphics[width=\textwidth]{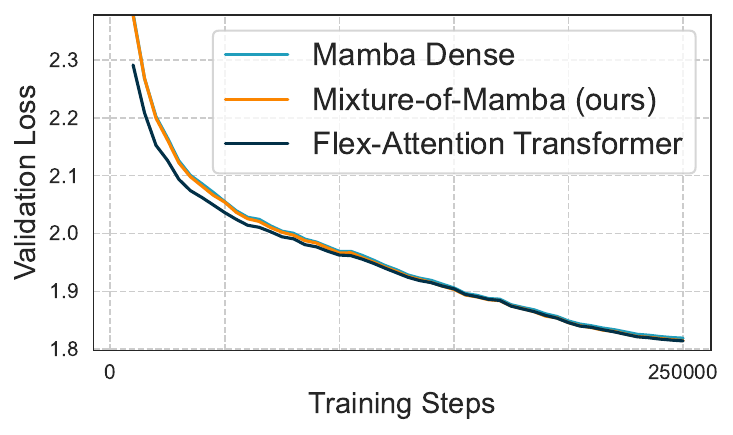}
        \caption{\textbf{760M} Wikipedia Val. Loss}
    \end{subfigure}
    \begin{subfigure}[b]{0.24\textwidth}
    \centering
    \includegraphics[width=\textwidth]{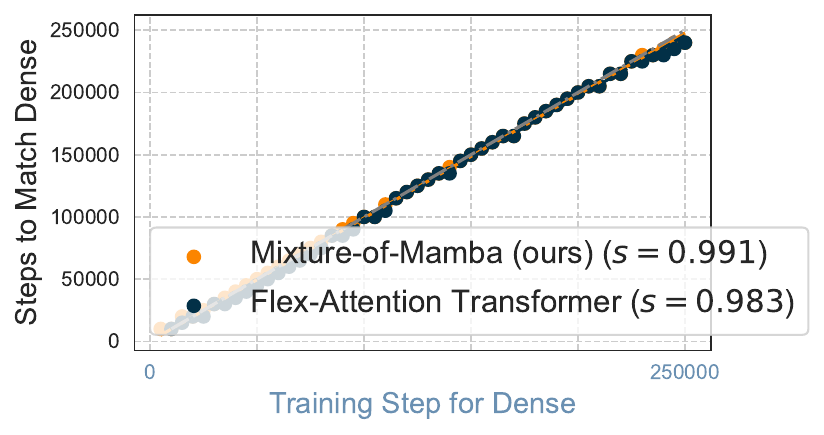}
    \caption{Wikipedia Loss Matching}
    \end{subfigure}

    \begin{subfigure}[b]{0.24\textwidth}
        \centering
        \includegraphics[width=\textwidth]{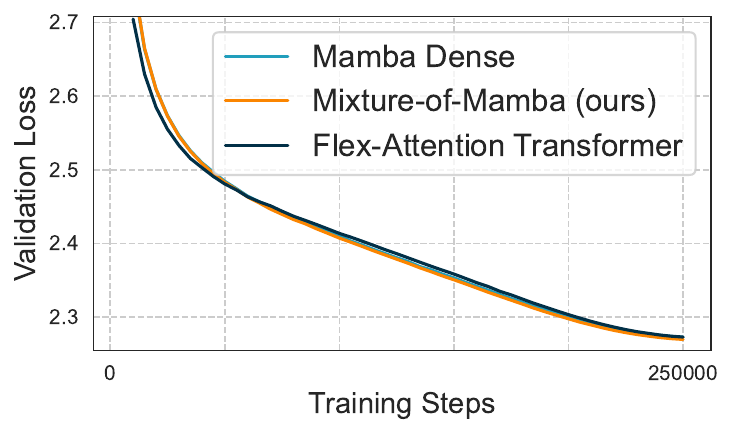}
        \caption{\textbf{1.4B} C4 Val. Loss}
    \end{subfigure}
    \hfill
    \begin{subfigure}[b]{0.24\textwidth}
    \centering
    \includegraphics[width=\textwidth]{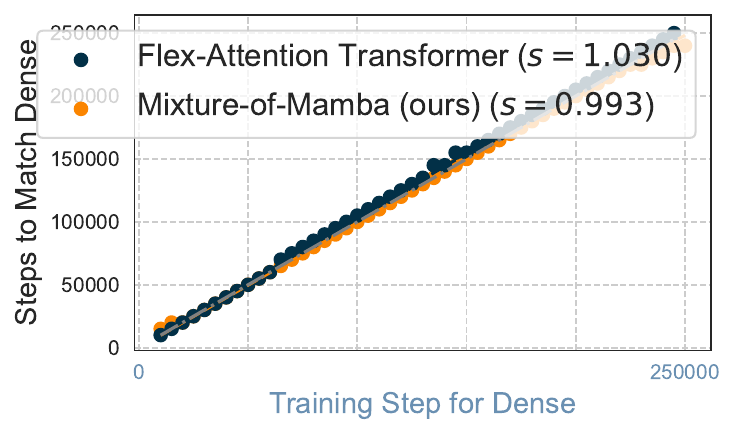}
    \caption{C4 Loss Matching}
    \end{subfigure}
    \hfill
    \begin{subfigure}[b]{0.24\textwidth}
        \centering
        \includegraphics[width=\textwidth]{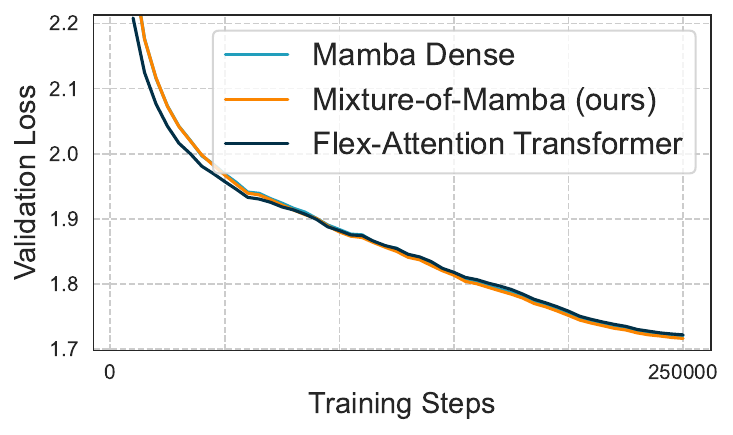}
        \caption{\textbf{1.4B} Wikipedia Val. Loss}
    \end{subfigure}
    \begin{subfigure}[b]{0.24\textwidth}
    \centering
    \includegraphics[width=\textwidth]{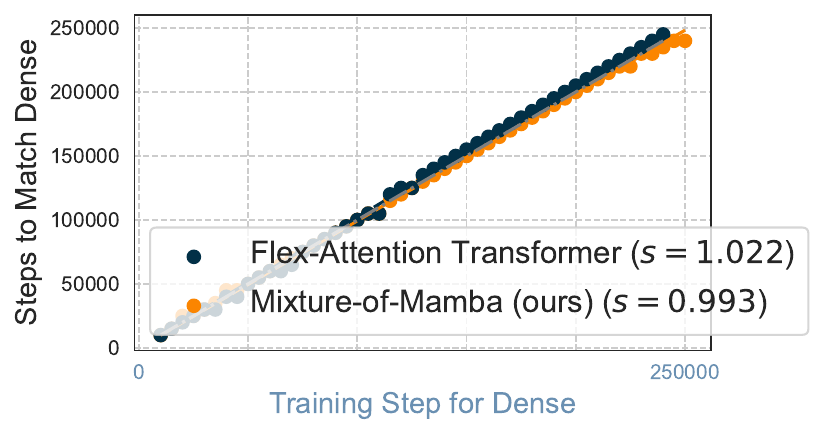}
    \caption{Wikipedia Loss Matching}
    \end{subfigure}  

    \caption{Validation loss and loss matching for text modality across model scales (\textbf{C4} and \textbf{Wikipedia} datasets) during multi-modal pretraining in the \textbf{Transfusion setting}. Results are shown for Mixture-of-Mamba, Mamba Dense, and Flex-Attention Transformer at three model scales: \textbf{163M}, \textbf{760M}, and \textbf{1.4B}. 
    (\textbf{a, e, i}) Validation loss on the \textbf{C4 dataset} shows that Mixture-of-Mamba achieves comparable performance at \textbf{163M} and performs marginally better than Mamba Dense and Flex-Attention Transformer at the \textbf{760M} and \textbf{1.4B} scales. 
    (\textbf{b, f, j}) Loss matching for C4 demonstrates that Mixture-of-Mamba reaches similar or slightly lower loss values at earlier training steps compared to Mamba Dense.
    (\textbf{c, g, k}) Validation loss on the \textbf{Wikipedia dataset} follows a similar trend, with Mixture-of-Mamba showing marginal improvements at the \textbf{760M} and \textbf{1.4B} scales.
    (\textbf{d, h, l}) Loss matching for Wikipedia illustrates efficient training dynamics, with Mixture-of-Mamba aligning closely with Flex-Attention Transformer while reaching comparable or slightly lower loss values than Mamba Dense.
    Overall, Mixture-of-Mamba demonstrates moderate improvements over both baselines at the larger scales (\textbf{760M} and \textbf{1.4B}).}
    \label{fig:appendix_text_val}
    
\end{figure*}

\clearpage
\newpage

\begin{figure}[t]
    \centering

   \begin{subfigure}[b]{0.24\textwidth}
        \centering
        \includegraphics[width=\textwidth]{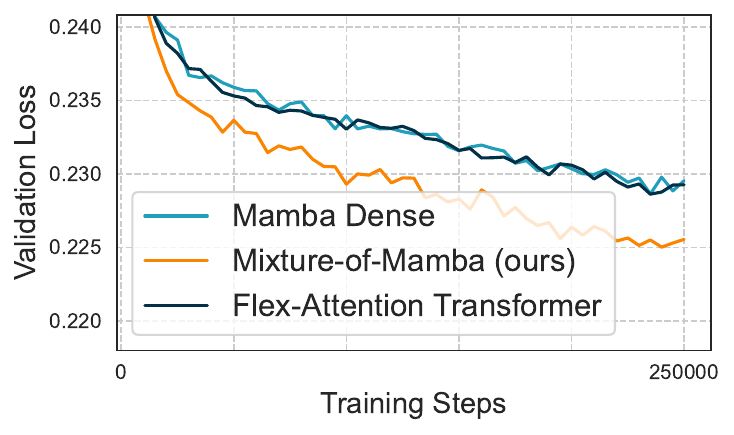}
        \caption{\textbf{163M} Image Val. Loss}
    \end{subfigure}
    \begin{subfigure}[b]{0.24\textwidth}
       \centering
       \includegraphics[width=\textwidth]{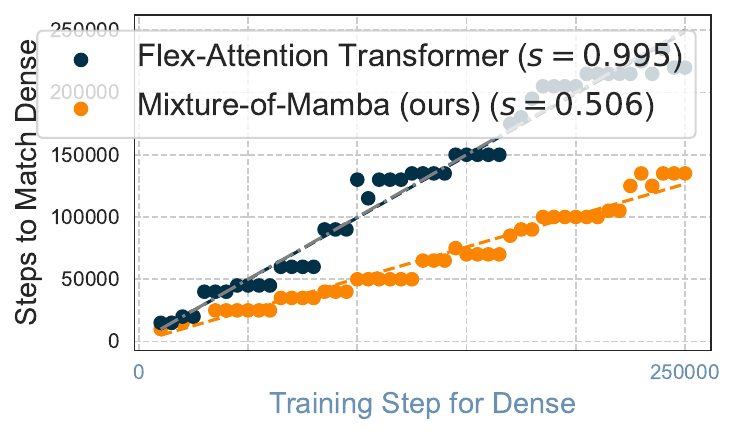}
       \caption{Image Val. Loss Matching}
   \end{subfigure}

    \begin{subfigure}[b]{0.24\textwidth}
        \centering
        \includegraphics[width=\textwidth]{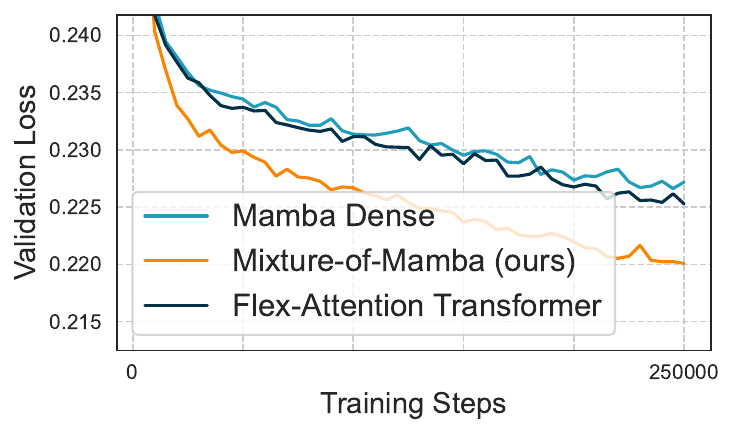}
        \caption{\textbf{760M} Image Val. Loss}
    \end{subfigure}
    \begin{subfigure}[b]{0.24\textwidth}
    \centering
    \includegraphics[width=\textwidth]{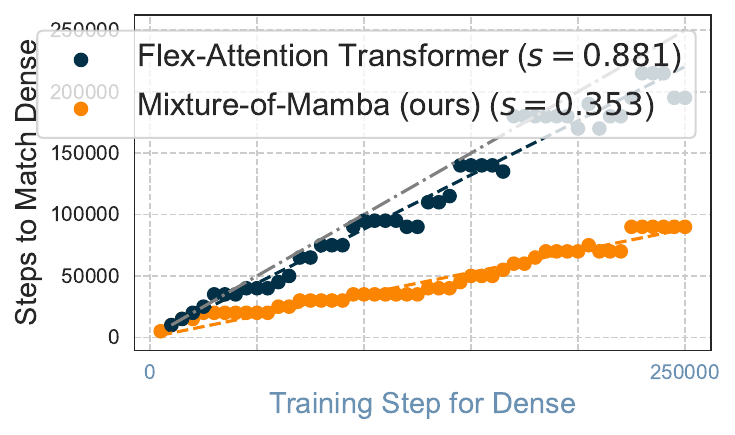}
    \caption{Image Val. Loss Matching}
    \end{subfigure}

    \begin{subfigure}[b]{0.24\textwidth}
        \centering
        \includegraphics[width=\textwidth]{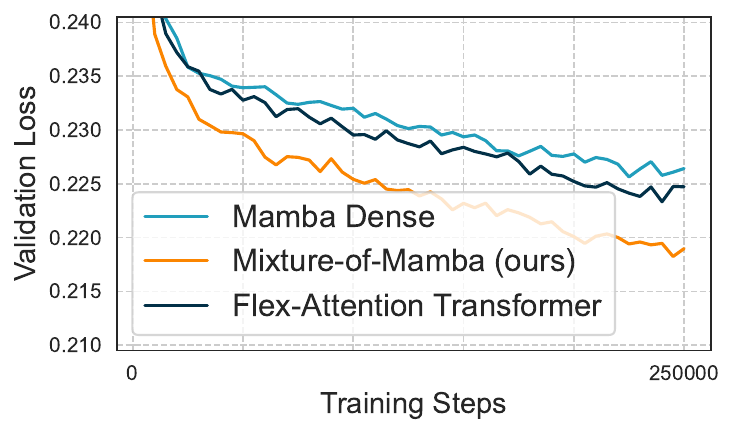}
        \caption{\textbf{1.4B} Image Val. Loss}
    \end{subfigure}
    \begin{subfigure}[b]{0.24\textwidth}
    \centering
    \includegraphics[width=\textwidth]{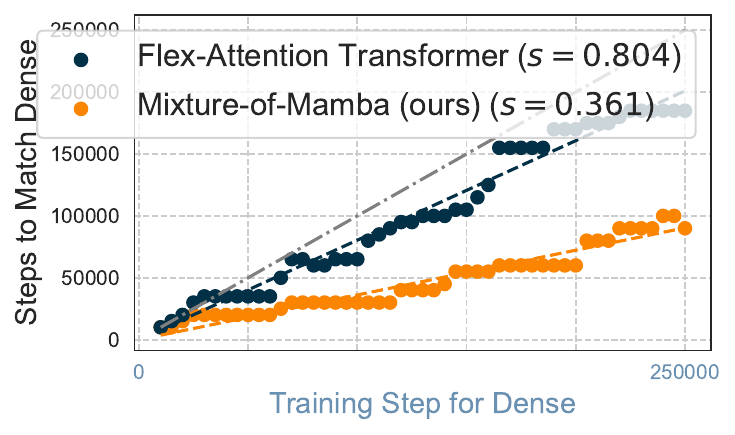}
    \caption{Image Val. Loss Matching}
    \end{subfigure}

    \caption{Image validation loss and loss matching on the \textbf{CC12M dataset} across three model scales: \textbf{163M}, \textbf{760M}, and \textbf{1.4B} during multi-modal pretraining in the \textbf{Transfusion setting}.  
    (\textbf{a, c, e}) Validation loss curves show that Mixture-of-Mamba achieves substantially lower image validation loss compared to Mamba Dense and Flex-Attention Transformer across all scales, with the improvement becoming more pronounced as model size increases.  
    (\textbf{b, d, f}) Loss matching curves demonstrate that Mixture-of-Mamba reaches the same loss values at earlier training steps compared to Mamba Dense, highlighting improved training efficiency.  
    Overall, Mixture-of-Mamba achieves large improvements in image validation loss on the \textbf{CC12M dataset}, showcasing its effectiveness in the image modality.}
    \label{fig:appendix_image_val_cc12m}
\end{figure}

\clearpage
\newpage

\begin{figure*}[t]
    \centering
    \begin{subfigure}[b]{0.24\textwidth}
        \centering
        \includegraphics[width=\textwidth]{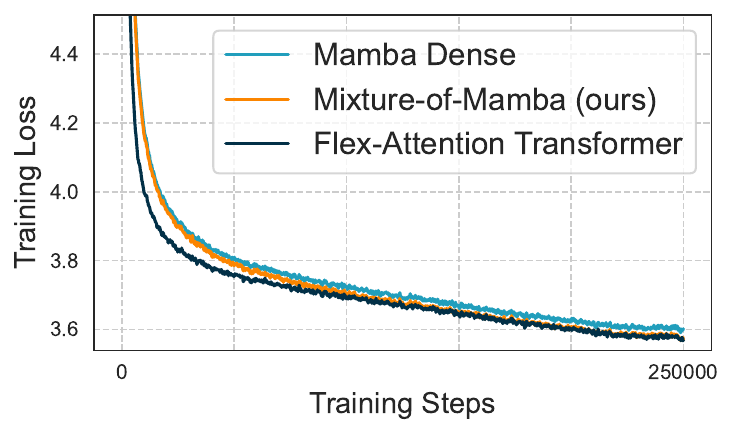}
        \caption{\textbf{163M} Avg Training Loss}
    \end{subfigure}
    \begin{subfigure}[b]{0.24\textwidth}
       \centering
       \includegraphics[width=\textwidth]{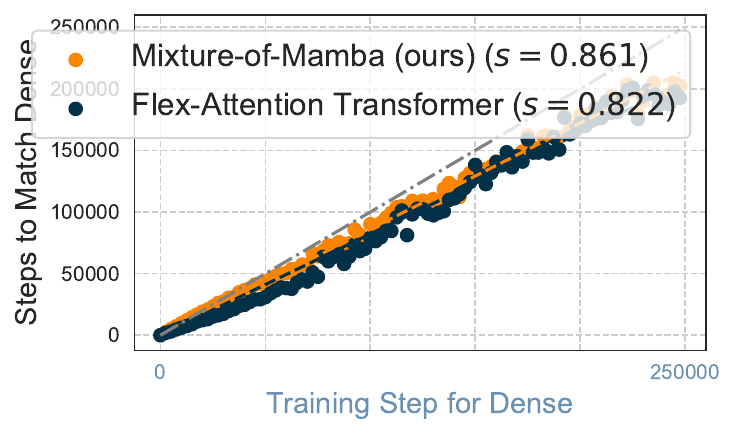}
       \caption{Avg Loss Matching}
   \end{subfigure}

   \begin{subfigure}[b]{0.24\textwidth}
        \centering
        \includegraphics[width=\textwidth]{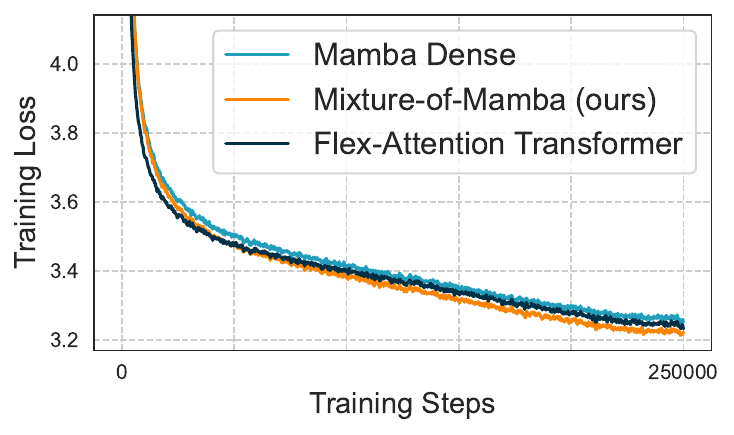}
        \caption{\textbf{760M} Avg Training Loss}
    \end{subfigure}
    \begin{subfigure}[b]{0.24\textwidth}
    \centering
    \includegraphics[width=\textwidth]{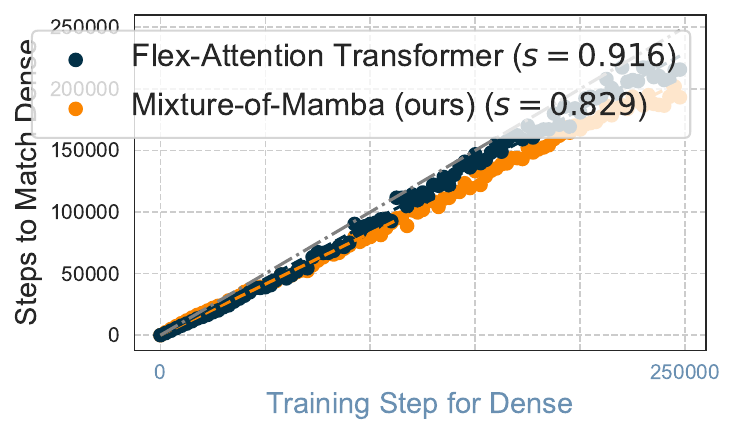}
    \caption{Avg Loss Matching}
    \end{subfigure}

    \begin{subfigure}[b]{0.24\textwidth}
        \centering
        \includegraphics[width=\textwidth]{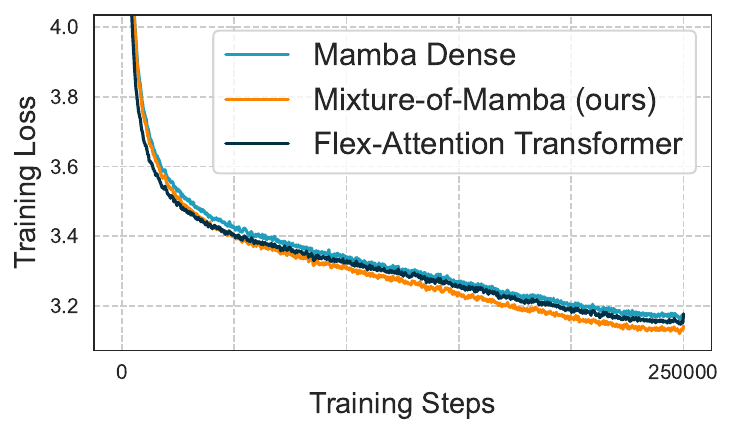}
        \caption{\textbf{1.4B} Avg Training Loss}
    \end{subfigure}
    \begin{subfigure}[b]{0.24\textwidth}
    \centering
    \includegraphics[width=\textwidth]{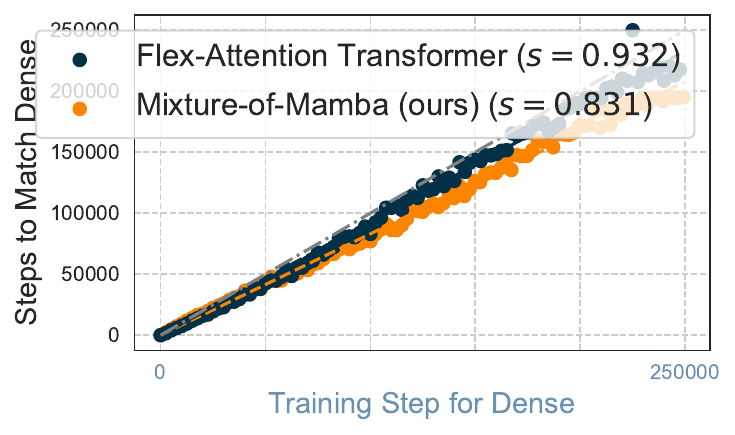}
    \caption{Avg Loss Matching}
    \end{subfigure}

    \caption{Overall training loss and loss matching during multi-modal pretraining in the \textbf{Transfusion setting}. Results are shown for Mixture-of-Mamba, Mamba Dense, and Flex-Attention Transformer at three model scales: \textbf{163M}, \textbf{760M}, and \textbf{1.4B}.  
    (\textbf{a, c, e}) Training loss averaged across the image and text modalities demonstrates that Mixture-of-Mamba achieves substantial improvements over Mamba Dense, with a notable reduction in training loss across all scales.  
    (\textbf{b, d, f}) Loss matching results show that Mixture-of-Mamba and Flex-Attention Transformer {reach the same loss values at earlier training steps} compared to Mamba Dense, highlighting improved training efficiency.  
    \textit{Note:} The image loss in the Transfusion setting corresponds to the diffusion loss, which is of smaller magnitude compared to the cross-entropy loss in the text modality.  
    Overall, Mixture-of-Mamba demonstrates significant gains in training loss and efficiency across multi-modal pretraining.}
    \label{fig:appendix_training_loss}
\end{figure*}

\begin{figure*}[t]
    \centering
    \begin{subfigure}[b]{0.24\textwidth}
        \centering
        \includegraphics[width=\textwidth]{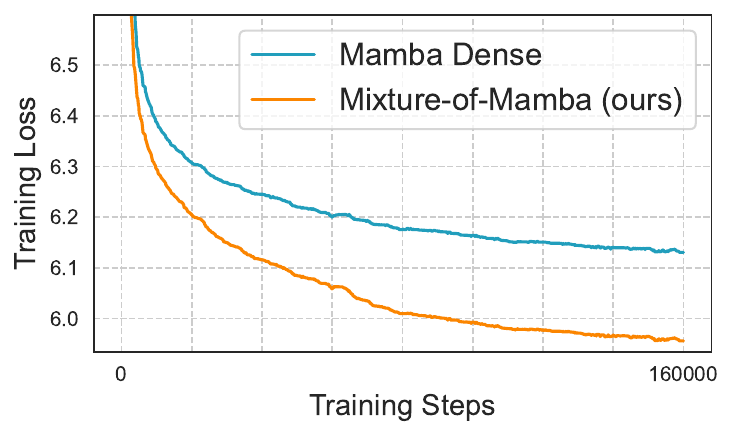}
        \caption{\textbf{37M} Image Training Loss}
    \end{subfigure}
    \hfill
    \begin{subfigure}[b]{0.24\textwidth}
       \centering
       \includegraphics[width=\textwidth]{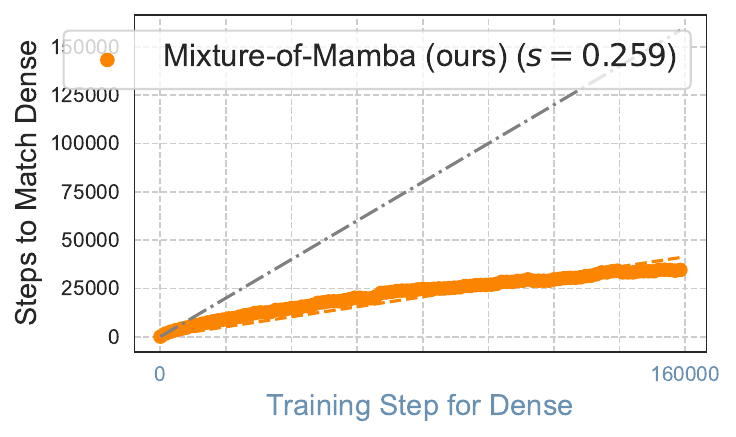}
       \caption{Image Loss Matching}
   \end{subfigure}
   \hfill
   \begin{subfigure}[b]{0.24\textwidth}
        \centering
        \includegraphics[width=\textwidth]{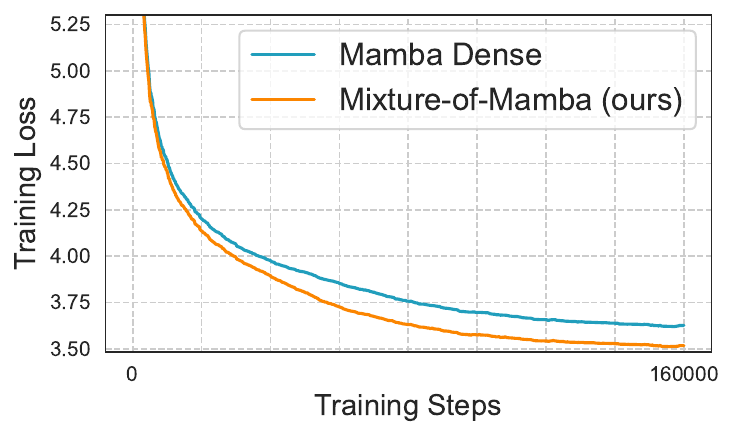}
        \caption{Text Training Loss}
    \end{subfigure}
    \hfill
    \begin{subfigure}[b]{0.24\textwidth}
       \centering
       \includegraphics[width=\textwidth]{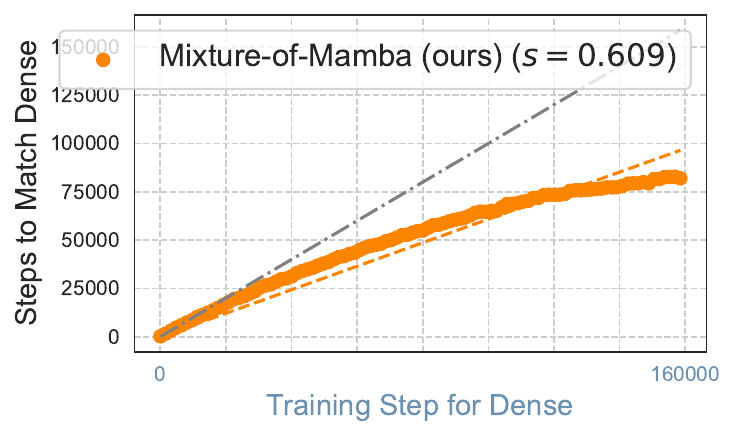}
       \caption{Text Loss Matching}
   \end{subfigure}
    \begin{subfigure}[b]{0.24\textwidth}
        \centering
        \includegraphics[width=\textwidth]{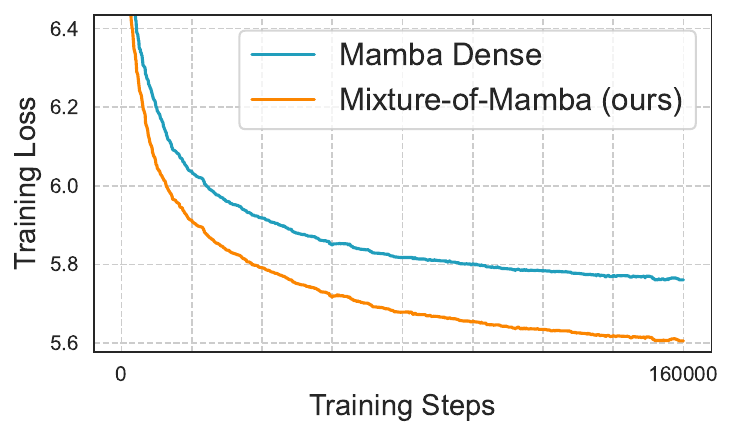}
        \caption{\textbf{94M} Image Training Loss}
    \end{subfigure}
    \hfill
    \begin{subfigure}[b]{0.24\textwidth}
       \centering
       \includegraphics[width=\textwidth]{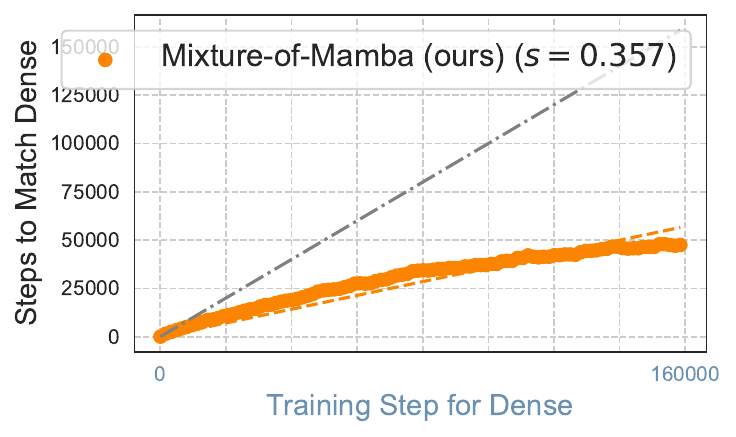}
       \caption{Image Loss Matching}
   \end{subfigure}
   \hfill
   \begin{subfigure}[b]{0.24\textwidth}
        \centering
        \includegraphics[width=\textwidth]{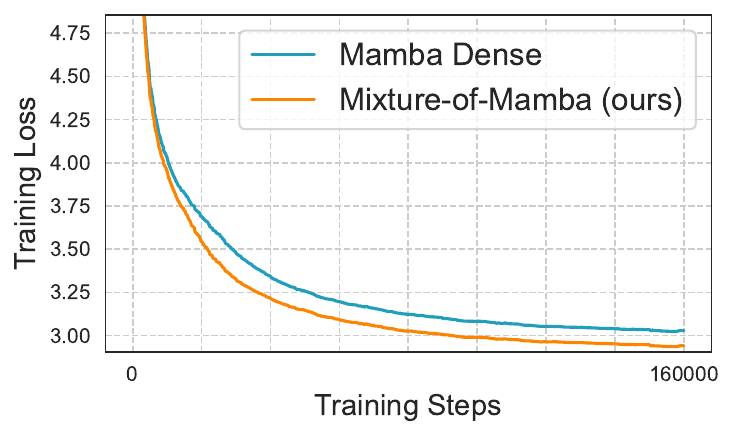}
        \caption{Text Training Loss}
    \end{subfigure}
    \hfill
    \begin{subfigure}[b]{0.24\textwidth}
       \centering
       \includegraphics[width=\textwidth]{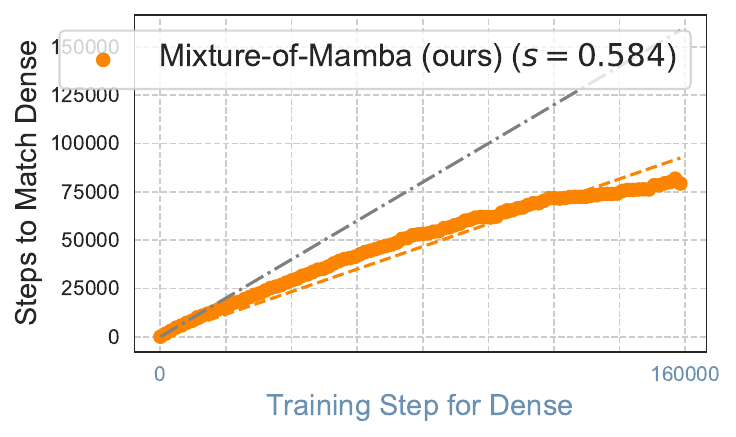}
       \caption{Text Loss Matching}
   \end{subfigure}
    \begin{subfigure}[b]{0.24\textwidth}
        \centering
        \includegraphics[width=\textwidth]{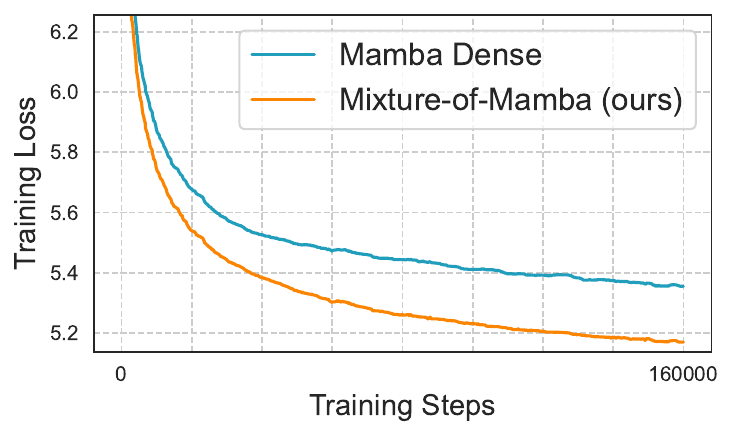}
        \caption{\textbf{443M} Image Training Loss}
    \end{subfigure}
    \hfill
    \begin{subfigure}[b]{0.24\textwidth}
       \centering
       \includegraphics[width=\textwidth]{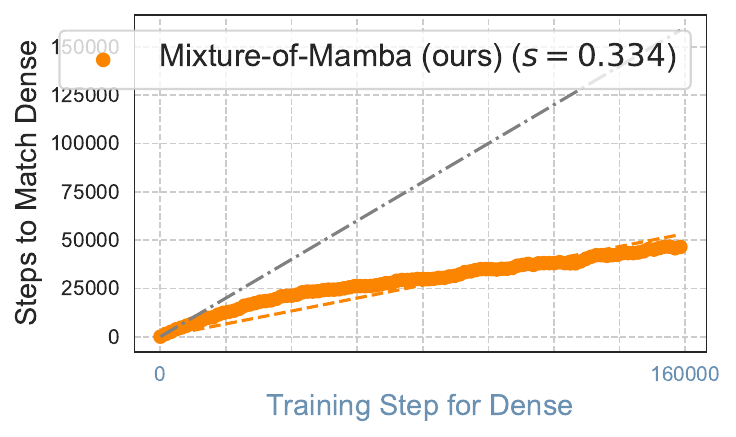}
       \caption{Image Loss Matching}
   \end{subfigure}
   \hfill
   \begin{subfigure}[b]{0.24\textwidth}
        \centering
        \includegraphics[width=\textwidth]{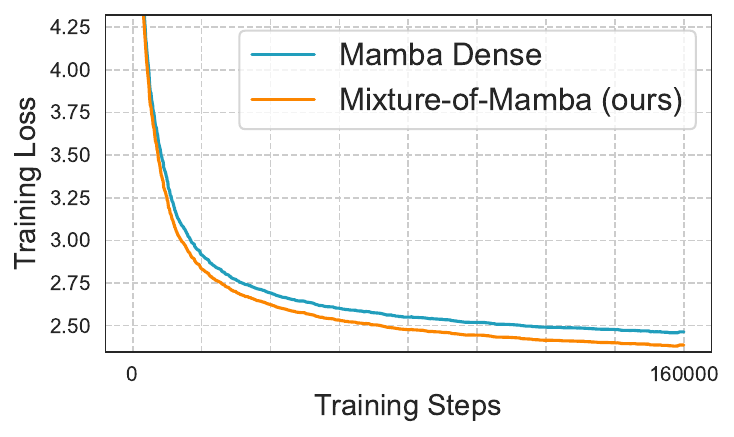}
        \caption{Text Training Loss}
    \end{subfigure}
    \hfill
    \begin{subfigure}[b]{0.24\textwidth}
       \centering
       \includegraphics[width=\textwidth]{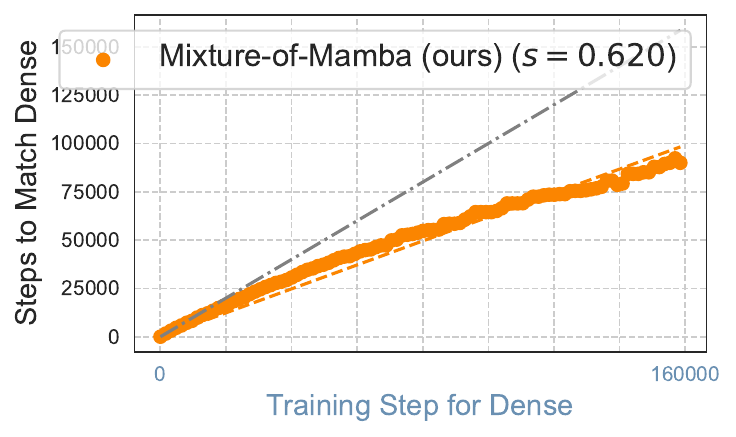}
       \caption{Text Loss Matching}
   \end{subfigure}
    
    \begin{subfigure}[b]{0.24\textwidth}
        \centering
        \includegraphics[width=\textwidth]{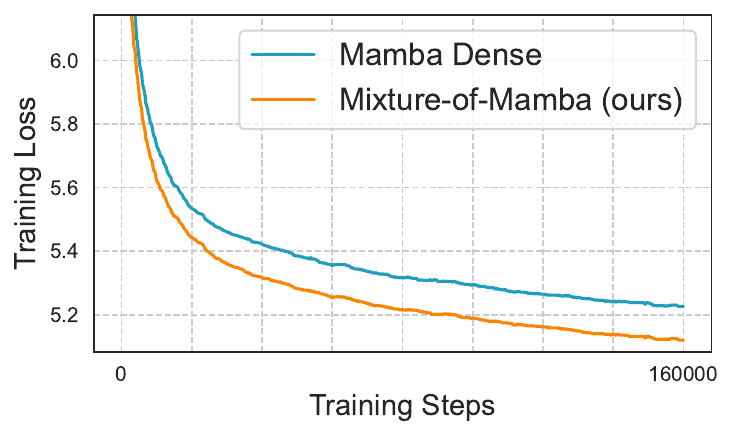}
        \caption{\textbf{880M} Image Training Loss}
    \end{subfigure}
    \hfill
    \begin{subfigure}[b]{0.24\textwidth}
       \centering
       \includegraphics[width=\textwidth]{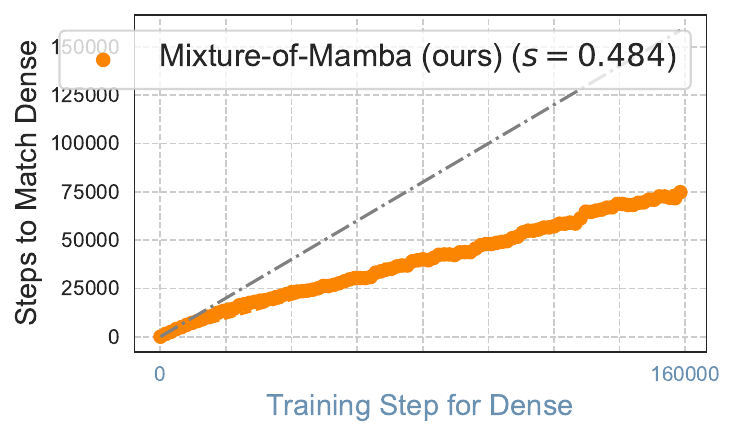}
       \caption{Image Loss Matching}
   \end{subfigure}
   \hfill
   \begin{subfigure}[b]{0.24\textwidth}
        \centering
        \includegraphics[width=\textwidth]{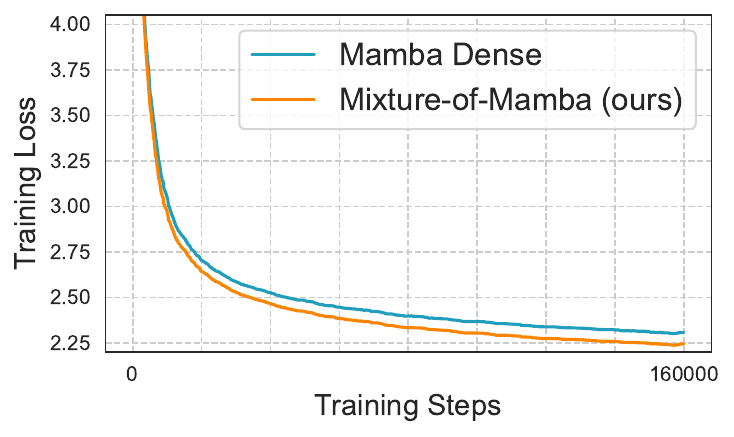}
        \caption{Text Training Loss}
    \end{subfigure}
    \hfill
    \begin{subfigure}[b]{0.24\textwidth}
       \centering
       \includegraphics[width=\textwidth]{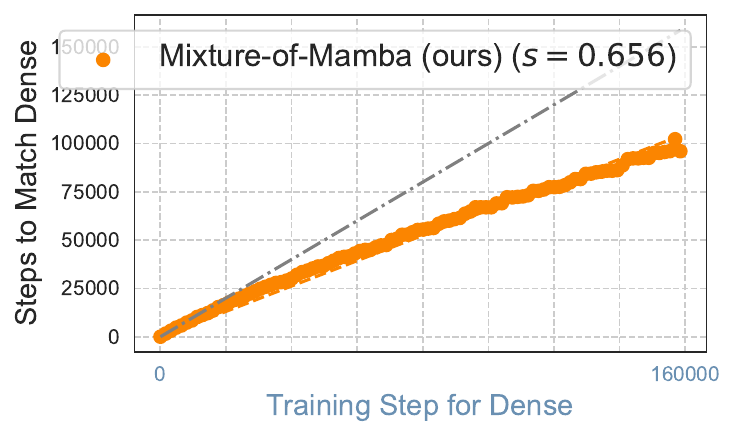}
       \caption{Text Loss Matching}
   \end{subfigure}

    \begin{subfigure}[b]{0.24\textwidth}
        \centering
        \includegraphics[width=\textwidth]{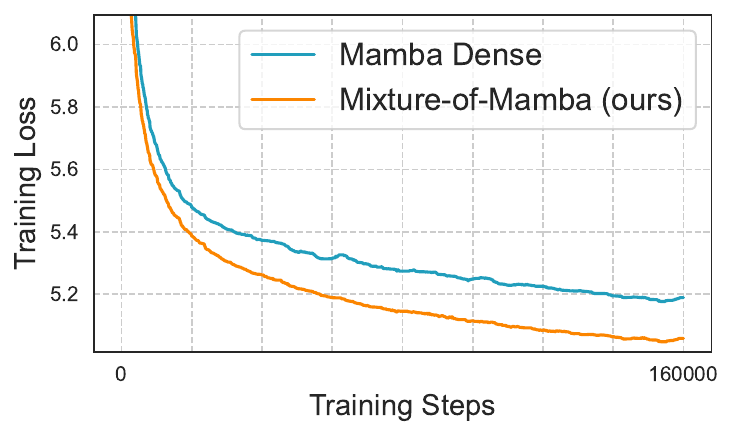}
        \caption{\textbf{1.5B} Image Training Loss}
    \end{subfigure}
    \hfill
    \begin{subfigure}[b]{0.24\textwidth}
       \centering
       \includegraphics[width=\textwidth]{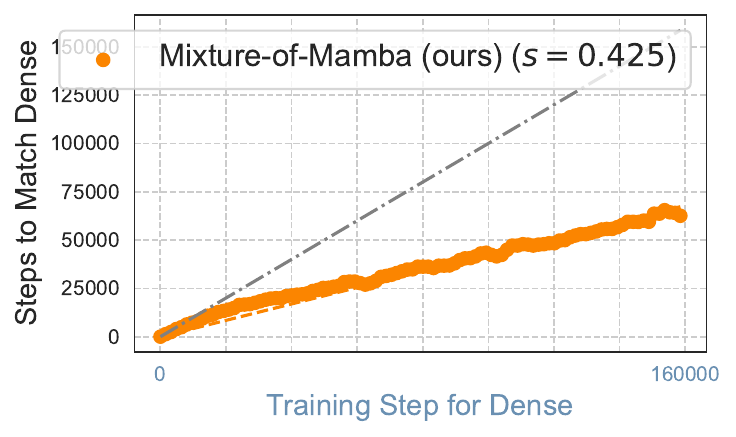}
       \caption{Image Loss Matching}
   \end{subfigure}
   \hfill
   \begin{subfigure}[b]{0.24\textwidth}
        \centering
        \includegraphics[width=\textwidth]{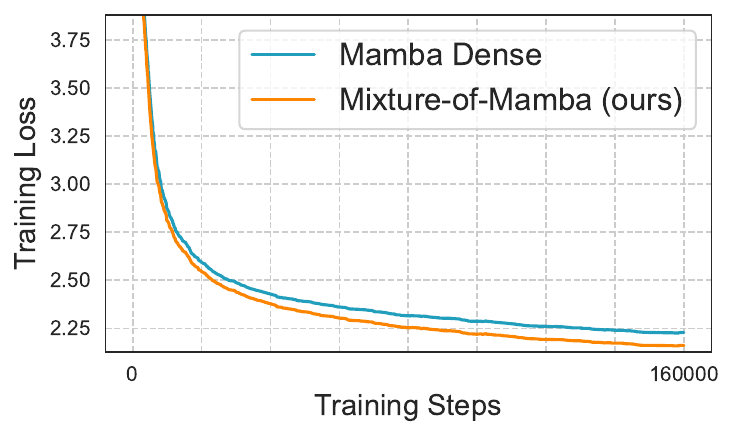}
        \caption{Text Training Loss}
    \end{subfigure}
    \hfill
    \begin{subfigure}[b]{0.24\textwidth}
       \centering
       \includegraphics[width=\textwidth]{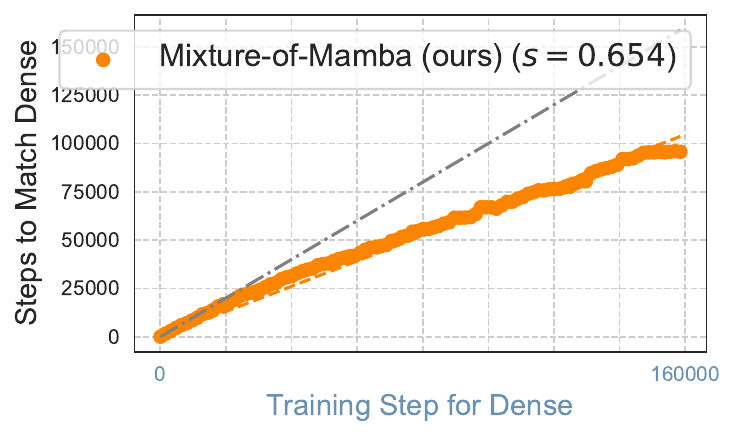}
       \caption{Text Loss Matching}
   \end{subfigure}

\caption{\textbf{Modality-specific pre-training loss and step matching plots across model scales (Chameleon setting).} 
Training loss and loss matching are reported for image and text modalities across five model scales: \textbf{37M}, \textbf{94M}, \textbf{443M}, \textbf{880M}, and \textbf{1.5B}.  
\textbf{(a, e, i, m, q)} Image training loss shows significant improvements for Mixture-of-Mamba (\textcolor{orange}{orange}), which consistently achieves lower loss compared to Mamba Dense (\textcolor{cyan}{cyan}) across all scales.  
\textbf{(b, f, j, n, r)} Image loss matching compares the training dynamics and shows that Mixture-of-Mamba reaches the same loss values at earlier training steps compared to Mamba Dense, highlighting its improved efficiency.  
\textbf{(c, g, k, o, s)} Text training loss demonstrates competitive performance, with Mixture-of-Mamba achieving slightly lower loss values compared to Mamba Dense.  
\textbf{(d, h, l, p, t)} Text loss matching illustrates that Mixture-of-Mamba reaches the same loss values at earlier training steps compared to Mamba Dense, reflecting its efficient training dynamics.  
Overall, in the \textbf{Chameleon setting}, Mixture-of-Mamba achieves consistent improvements in the image modality, with substantial computational savings, while also demonstrating meaningful gains in the text modality.}
\label{fig:chameleon_training}

\end{figure*}

\begin{figure*}[t]
    \centering
    \begin{subfigure}[b]{0.24\textwidth}
        \centering
        \includegraphics[width=\textwidth]{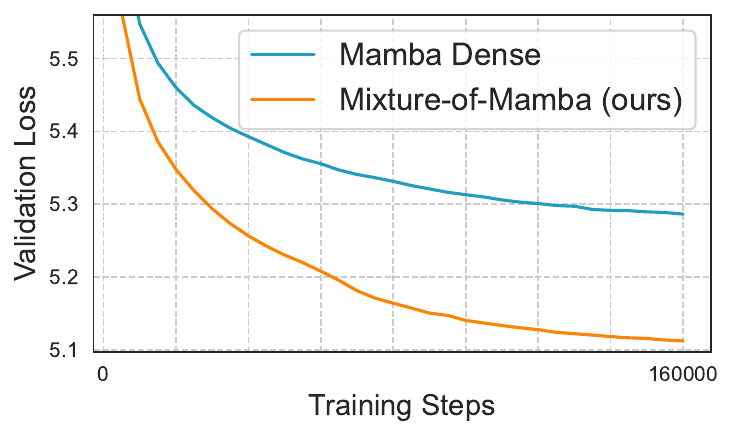}
        \caption{\textbf{37M} Image Eval Loss}
    \end{subfigure}
    \hfill
    \begin{subfigure}[b]{0.24\textwidth}
       \centering
       \includegraphics[width=\textwidth]{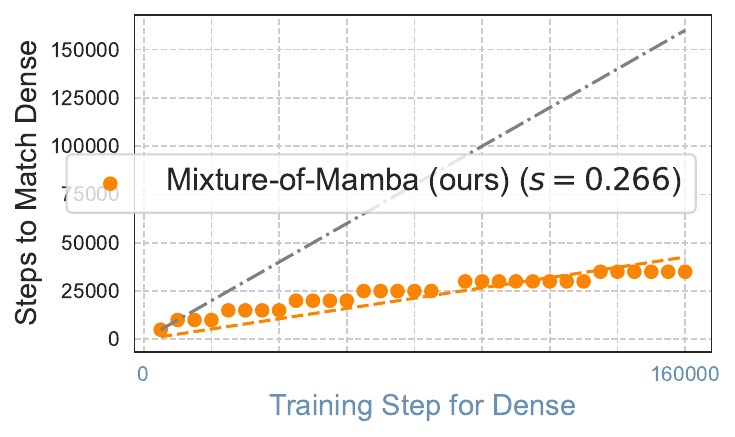}
       \caption{Image Loss Matching}
   \end{subfigure}
   \hfill
   \begin{subfigure}[b]{0.24\textwidth}
        \centering
        \includegraphics[width=\textwidth]{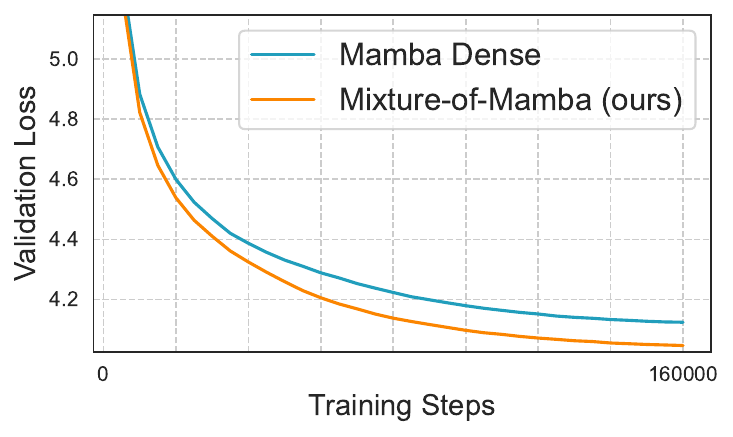}
        \caption{Text Eval Loss}
    \end{subfigure}
    \hfill
    \begin{subfigure}[b]{0.24\textwidth}
       \centering
       \includegraphics[width=\textwidth]{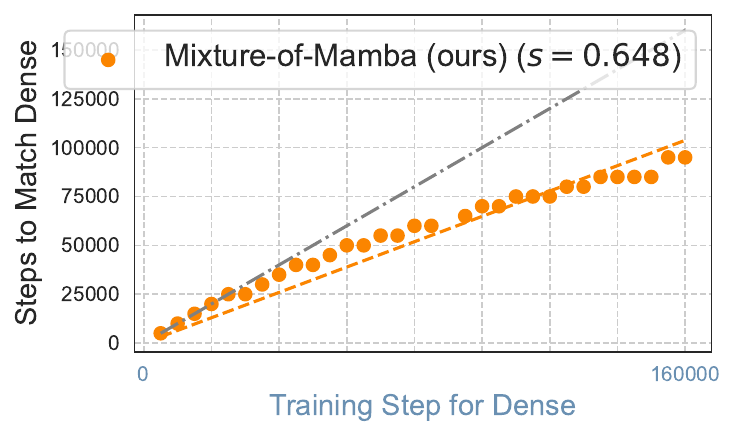}
       \caption{Text Loss Matching}
   \end{subfigure}
    \begin{subfigure}[b]{0.24\textwidth}
        \centering
        \includegraphics[width=\textwidth]{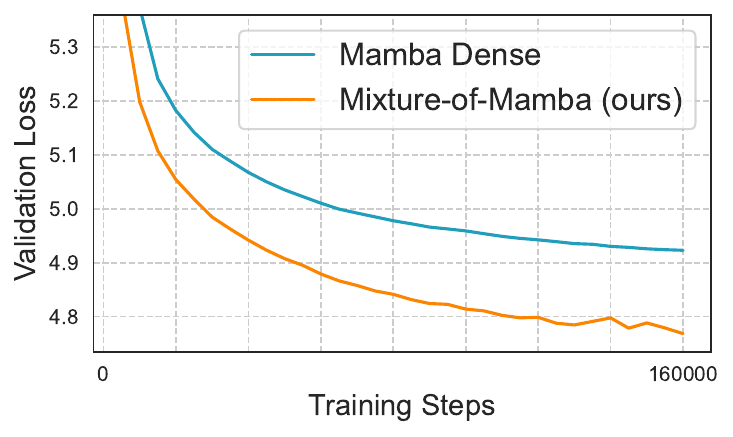}
        \caption{\textbf{94M} Image Eval Loss}
    \end{subfigure}
    \hfill
    \begin{subfigure}[b]{0.24\textwidth}
       \centering
       \includegraphics[width=\textwidth]{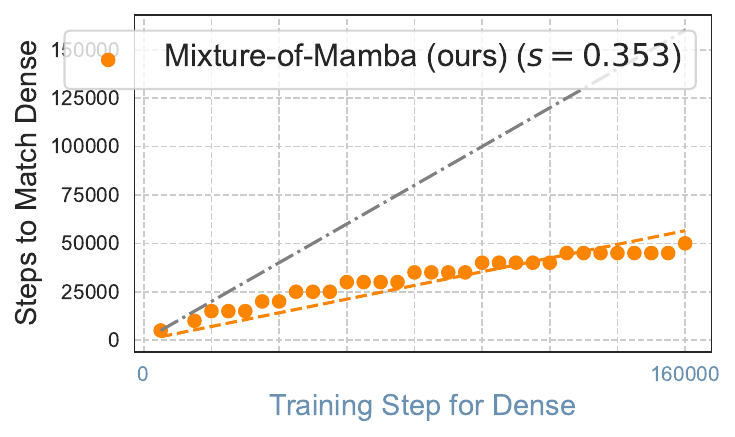}
       \caption{Image Loss Matching}
   \end{subfigure}
   \hfill
   \begin{subfigure}[b]{0.24\textwidth}
        \centering
        \includegraphics[width=\textwidth]{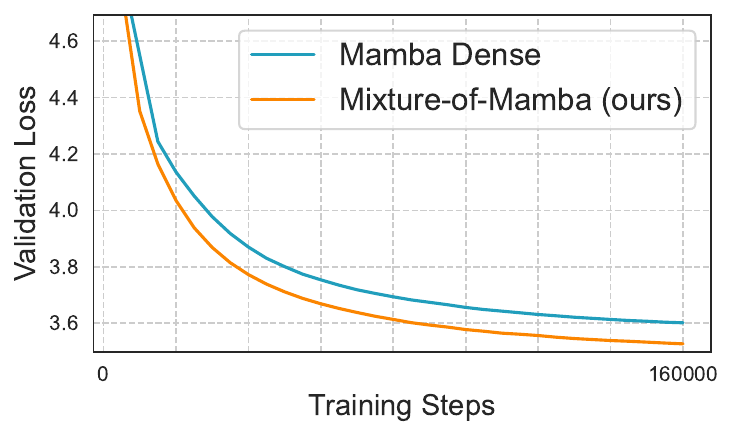}
        \caption{Text Eval Loss}
    \end{subfigure}
    \hfill
    \begin{subfigure}[b]{0.24\textwidth}
       \centering
       \includegraphics[width=\textwidth]{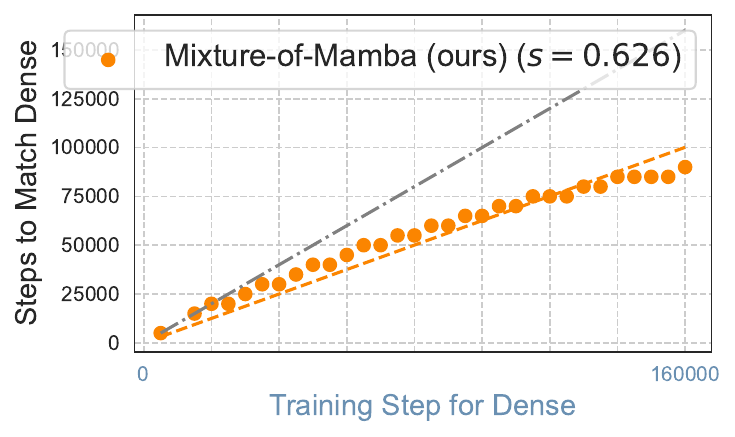}
       \caption{Text Loss Matching}
   \end{subfigure}
    \begin{subfigure}[b]{0.24\textwidth}
        \centering
        \includegraphics[width=\textwidth]{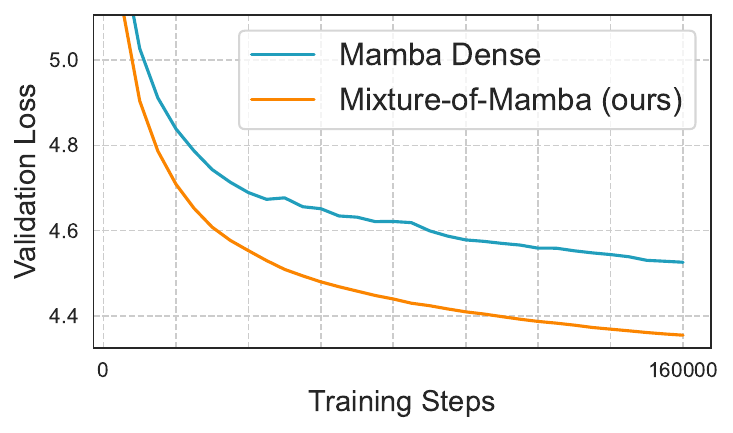}
        \caption{\textbf{443M} Image Eval Loss}
    \end{subfigure}
    \hfill
    \begin{subfigure}[b]{0.24\textwidth}
       \centering
       \includegraphics[width=\textwidth]{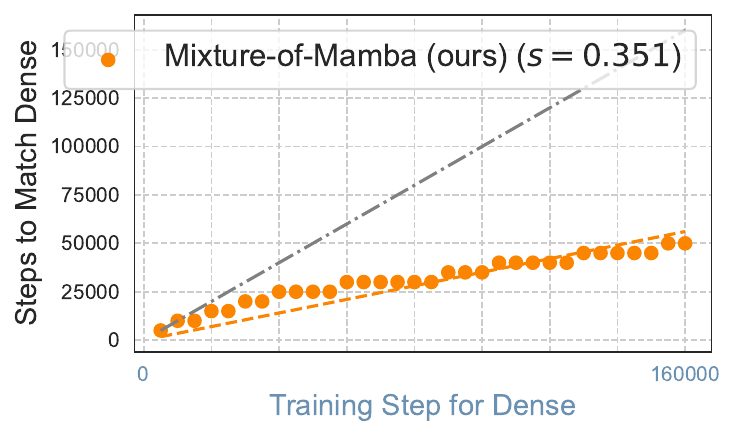}
       \caption{Image Loss Matching}
   \end{subfigure}
   \hfill
   \begin{subfigure}[b]{0.24\textwidth}
        \centering
        \includegraphics[width=\textwidth]{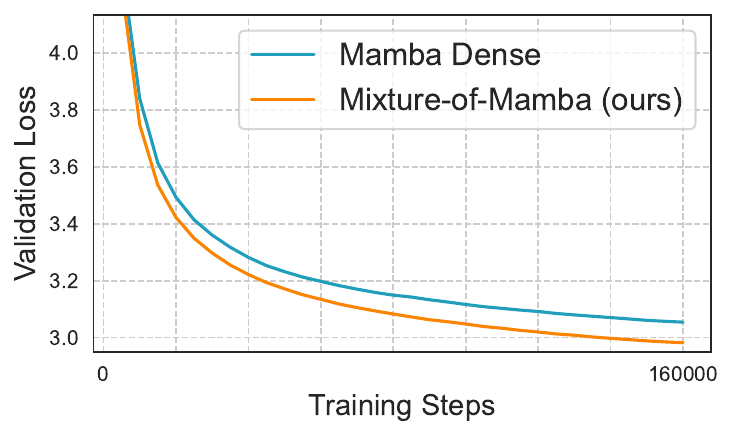}
        \caption{Text Eval Loss}
    \end{subfigure}
    \hfill
    \begin{subfigure}[b]{0.24\textwidth}
       \centering
       \includegraphics[width=\textwidth]{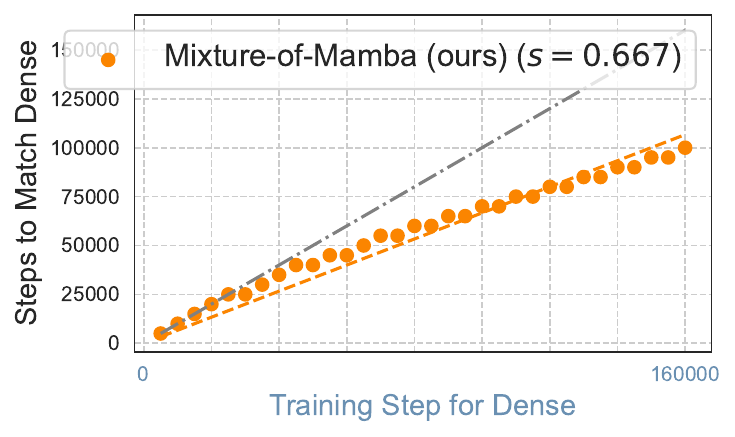}
       \caption{Text Loss Matching}
   \end{subfigure}

    \begin{subfigure}[b]{0.24\textwidth}
        \centering
        \includegraphics[width=\textwidth]{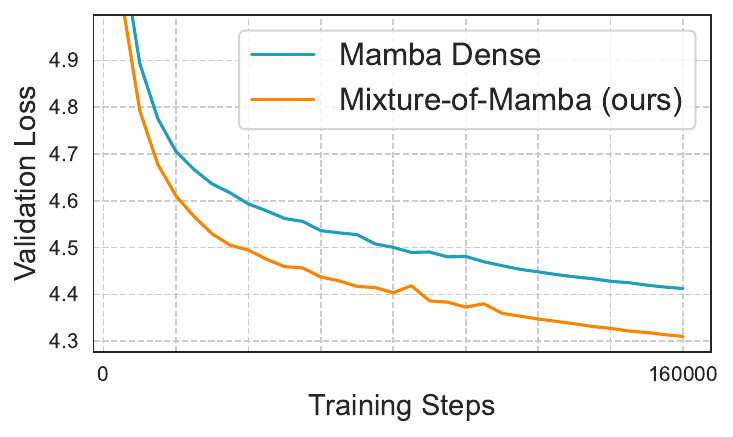}
        \caption{\textbf{880M} Image Eval Loss}
    \end{subfigure}
    \hfill
    \begin{subfigure}[b]{0.24\textwidth}
       \centering
       \includegraphics[width=\textwidth]{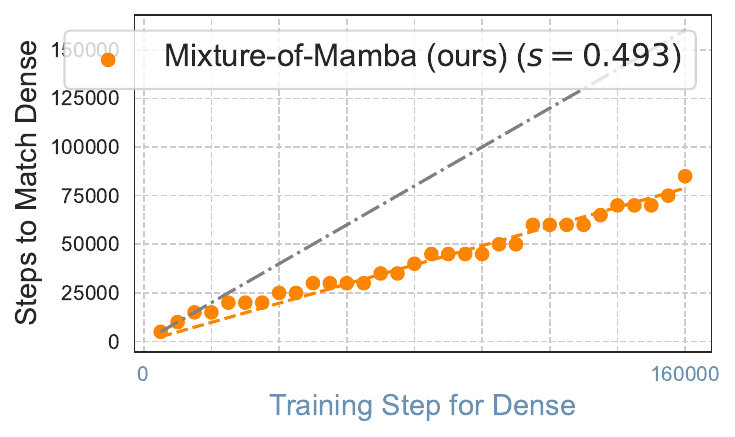}
       \caption{Image Loss Matching}
   \end{subfigure}
   \hfill
   \begin{subfigure}[b]{0.24\textwidth}
        \centering
        \includegraphics[width=\textwidth]{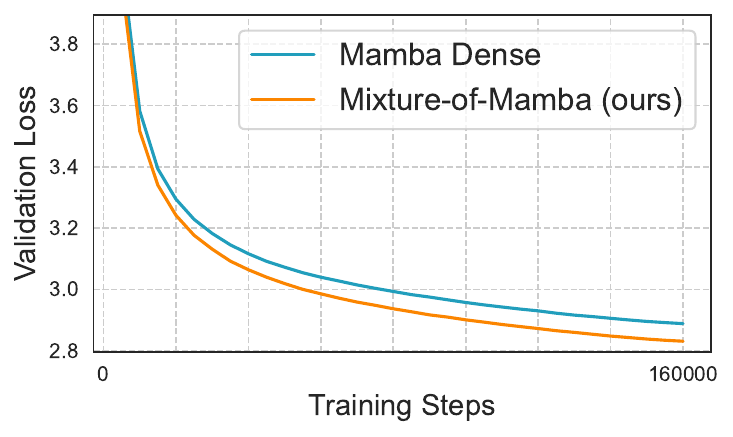}
        \caption{Text Eval Loss}
    \end{subfigure}
    \hfill
    \begin{subfigure}[b]{0.24\textwidth}
       \centering
       \includegraphics[width=\textwidth]{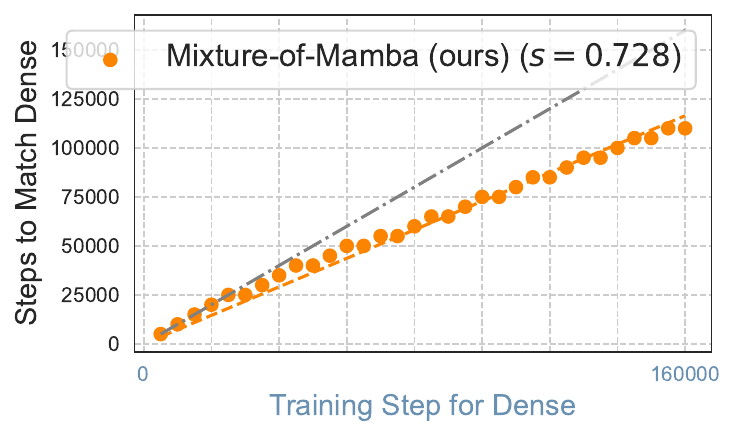}
       \caption{Text Loss Matching}
   \end{subfigure}

    \begin{subfigure}[b]{0.24\textwidth}
        \centering
        \includegraphics[width=\textwidth]{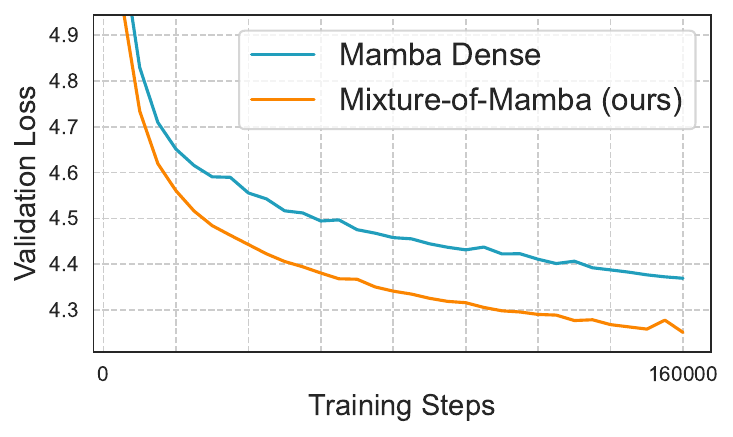}
        \caption{\textbf{1.5B} Image Eval Loss}
    \end{subfigure}
    \hfill
    \begin{subfigure}[b]{0.24\textwidth}
       \centering
       \includegraphics[width=\textwidth]{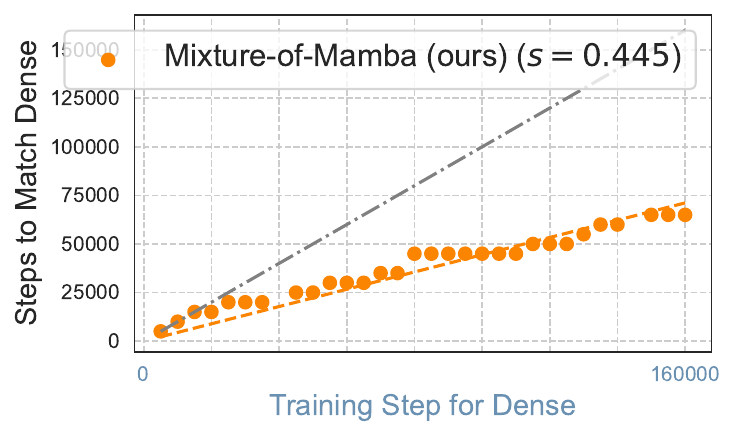}
       \caption{Image Loss Matching}
   \end{subfigure}
   \hfill
   \begin{subfigure}[b]{0.24\textwidth}
        \centering
        \includegraphics[width=\textwidth]{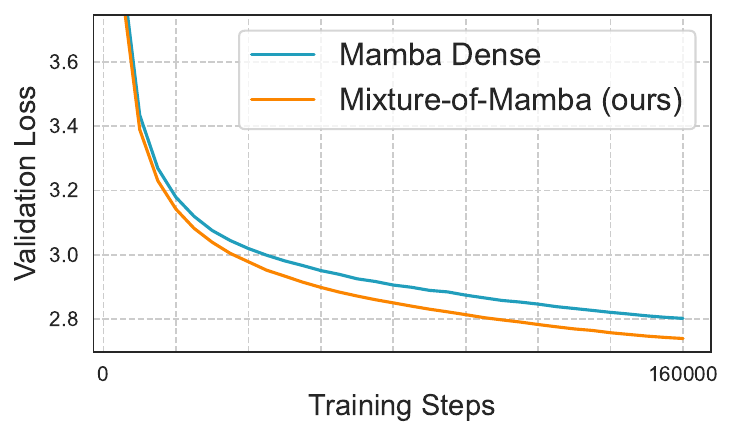}
        \caption{Text Eval Loss}
    \end{subfigure}
    \hfill
    \begin{subfigure}[b]{0.24\textwidth}
       \centering
       \includegraphics[width=\textwidth]{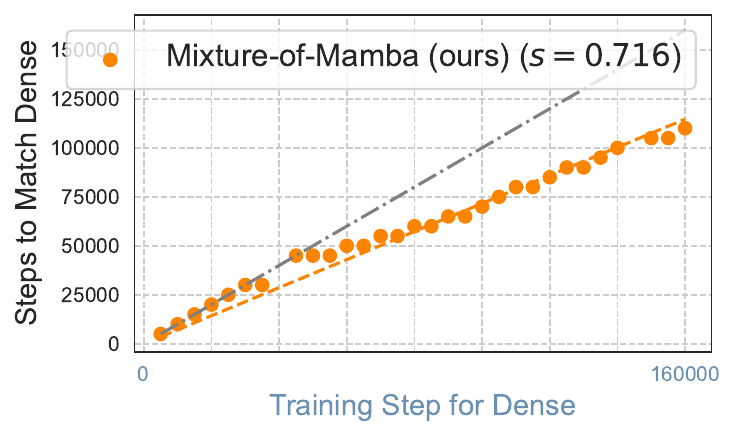}
       \caption{Text Loss Matching}
   \end{subfigure}

\caption{\textbf{Training and evaluation losses for image and text modalities across model scales in the Chameleon setting on the Obelisc dataset.} 
Results are shown for Mixture-of-Mamba and Mamba Dense across five model scales: \textbf{37M}, \textbf{94M}, \textbf{443M}, \textbf{880M}, and \textbf{1.5B}.  
\textbf{(a, e, i, m, q)} Image evaluation loss demonstrates consistent improvements for Mixture-of-Mamba (\textcolor{orange}{orange}), achieving lower loss compared to Mamba Dense (\textcolor{cyan}{cyan}) across all scales.  
\textbf{(b, f, j, n, r)} Image loss matching shows that Mixture-of-Mamba {reaches the same loss values at earlier training steps} compared to Mamba Dense, reflecting its improved training efficiency.  
\textbf{(c, g, k, o, s)} Text evaluation loss indicates competitive results for Mixture-of-Mamba, achieving lower losses relative to Mamba Dense.  
\textbf{(d, h, l, p, t)} Text loss matching highlights that Mixture-of-Mamba reaches the same loss values at earlier training steps, further demonstrating its efficiency in the text modality.  
Overall, Mixture-of-Mamba achieves strong and consistent improvements in both image and text modalities across all model scales in the Chameleon setting evaluated on the Obelisc dataset.}
\label{fig:appendix_chameleon_obelisc}

\end{figure*}

\begin{figure*}[t]
    \centering
    \begin{subfigure}[b]{0.24\textwidth}
        \centering
        \includegraphics[width=\textwidth]{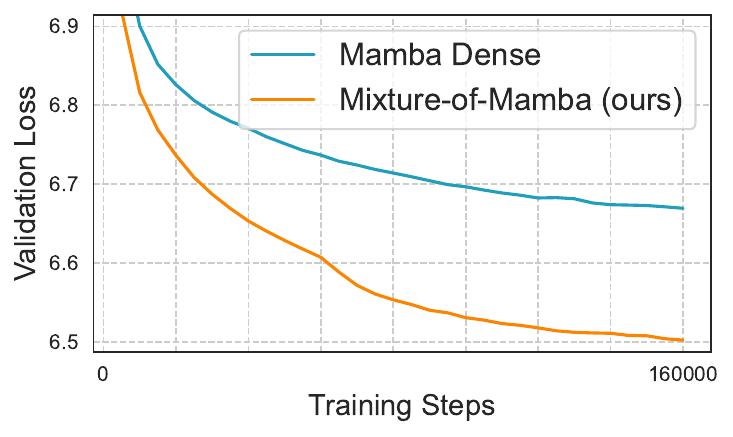}
        \caption{\textbf{37M} Image Eval Loss}
    \end{subfigure}
    \hfill
    \begin{subfigure}[b]{0.24\textwidth}
       \centering
       \includegraphics[width=\textwidth]{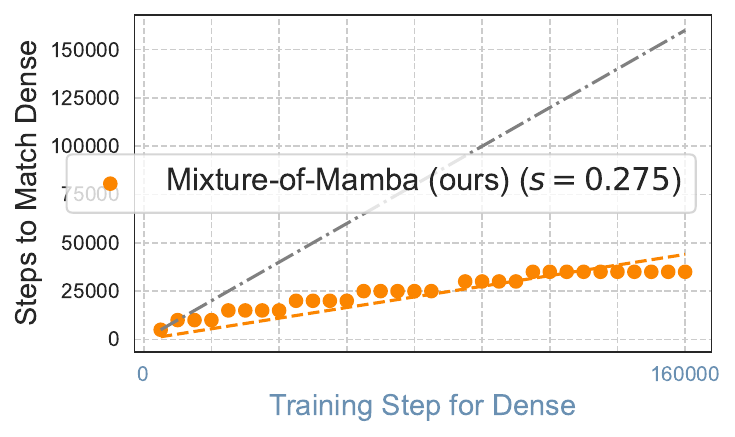}
       \caption{Image Loss Matching}
   \end{subfigure}
   \hfill
   \begin{subfigure}[b]{0.24\textwidth}
        \centering
        \includegraphics[width=\textwidth]{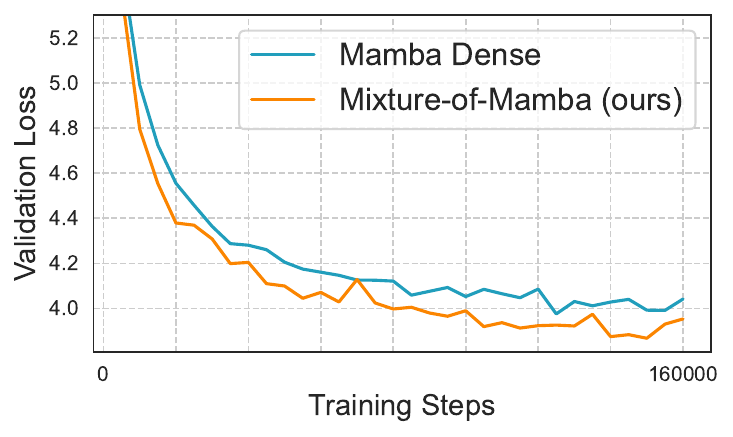}
        \caption{Text Eval Loss}
    \end{subfigure}
    \hfill
    \begin{subfigure}[b]{0.24\textwidth}
       \centering
       \includegraphics[width=\textwidth]{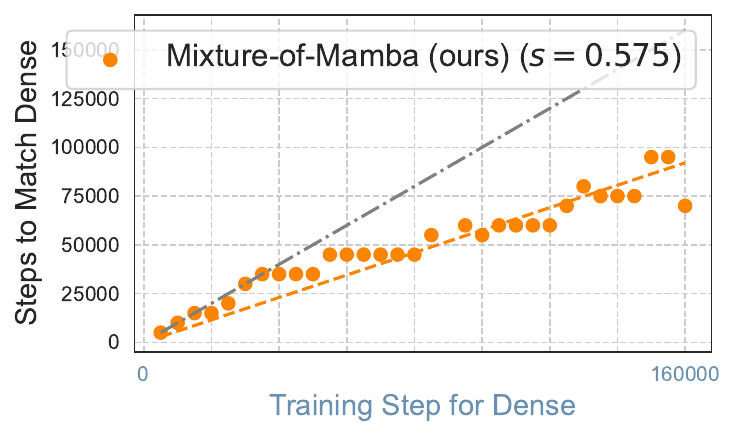}
       \caption{Text Loss Matching}
   \end{subfigure}
    \begin{subfigure}[b]{0.24\textwidth}
        \centering
        \includegraphics[width=\textwidth]{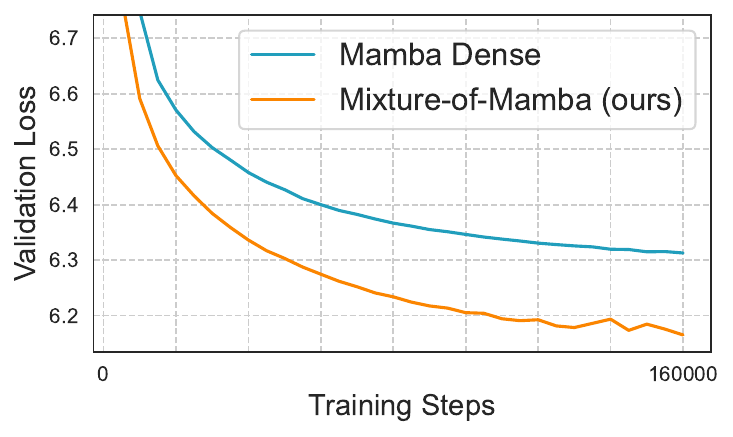}
        \caption{\textbf{94M} Image Eval Loss}
    \end{subfigure}
    \hfill
    \begin{subfigure}[b]{0.24\textwidth}
       \centering
       \includegraphics[width=\textwidth]{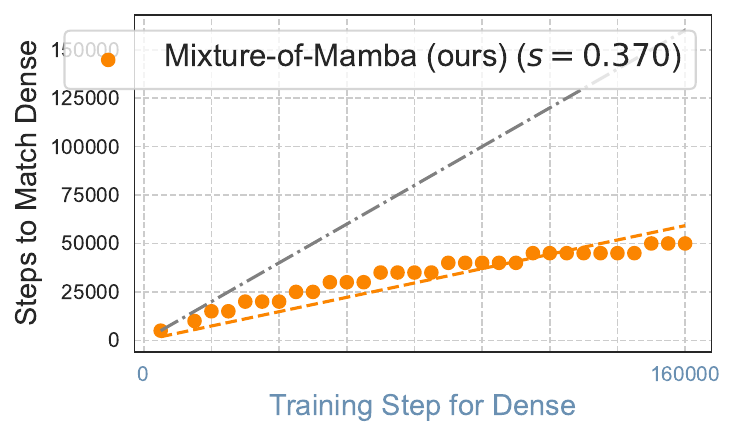}
       \caption{Image Loss Matching}
   \end{subfigure}
   \hfill
   \begin{subfigure}[b]{0.24\textwidth}
        \centering
        \includegraphics[width=\textwidth]{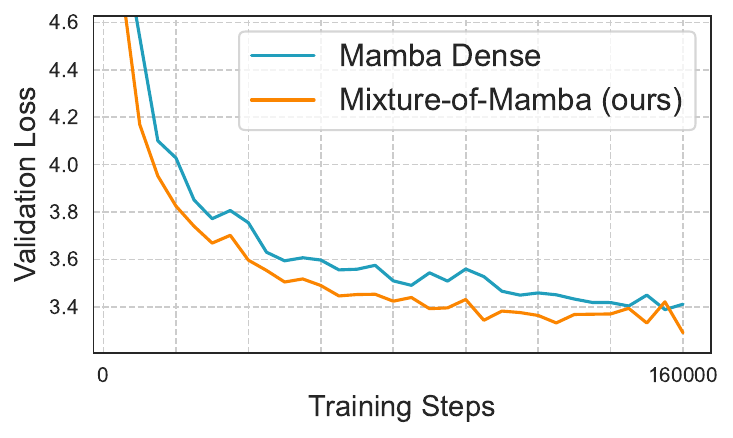}
        \caption{Text Eval Loss}
    \end{subfigure}
    \hfill
    \begin{subfigure}[b]{0.24\textwidth}
       \centering
       \includegraphics[width=\textwidth]{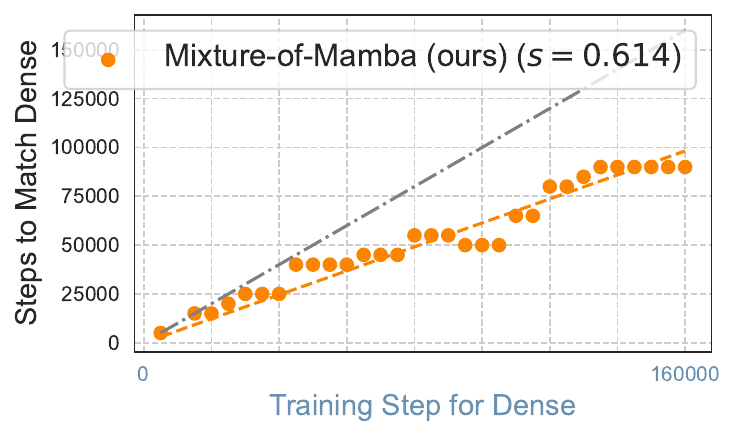}
       \caption{Text Loss Matching}
   \end{subfigure}
    \begin{subfigure}[b]{0.24\textwidth}
        \centering
        \includegraphics[width=\textwidth]{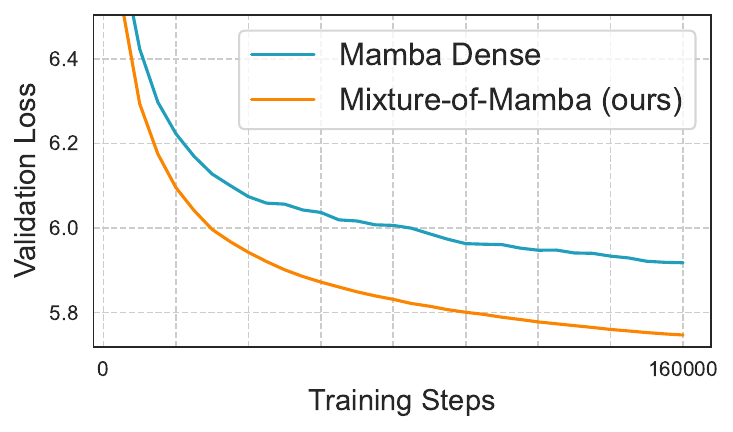}
        \caption{\textbf{443M} Image Eval Loss}
    \end{subfigure}
    \hfill
    \begin{subfigure}[b]{0.24\textwidth}
       \centering
       \includegraphics[width=\textwidth]{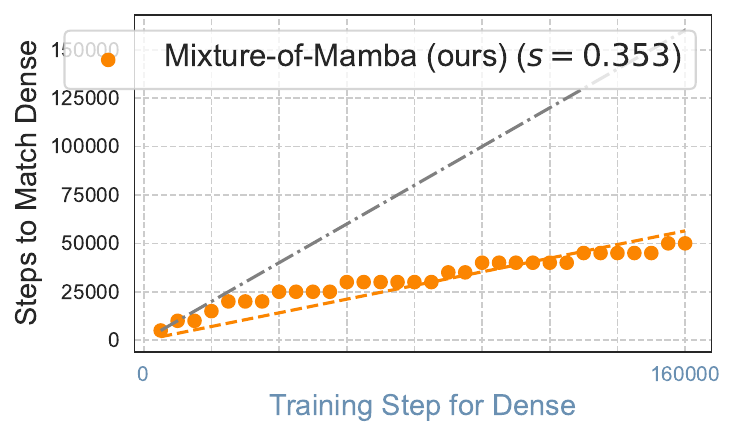}
       \caption{Image Loss Matching}
   \end{subfigure}
   \hfill
   \begin{subfigure}[b]{0.24\textwidth}
        \centering
        \includegraphics[width=\textwidth]{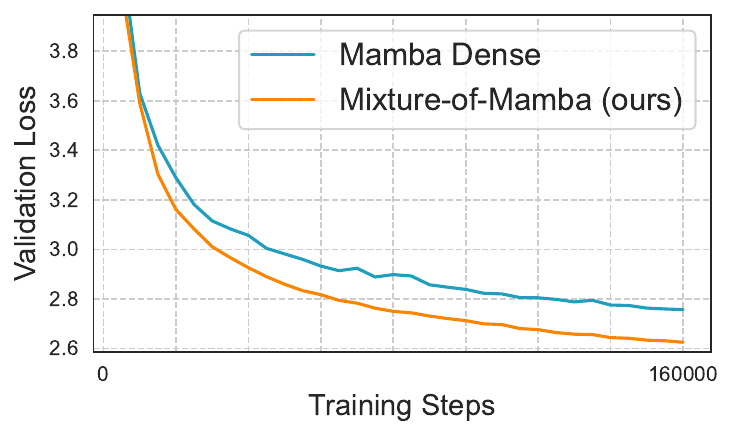}
        \caption{Text Eval Loss}
    \end{subfigure}
    \hfill
    \begin{subfigure}[b]{0.24\textwidth}
       \centering
       \includegraphics[width=\textwidth]{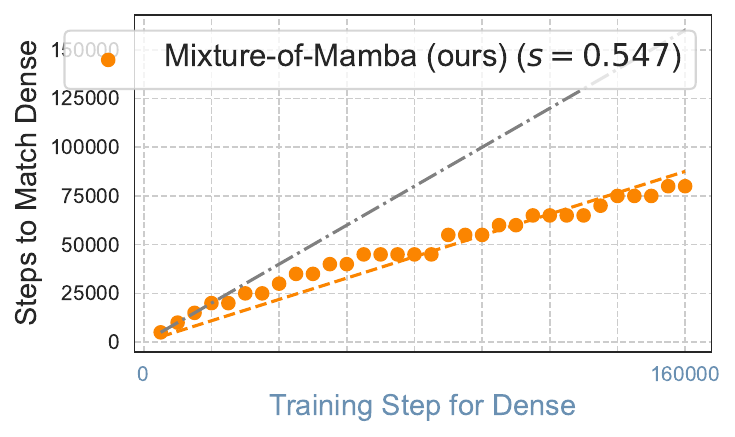}
       \caption{Text Loss Matching}
   \end{subfigure}

    \begin{subfigure}[b]{0.24\textwidth}
        \centering
        \includegraphics[width=\textwidth]{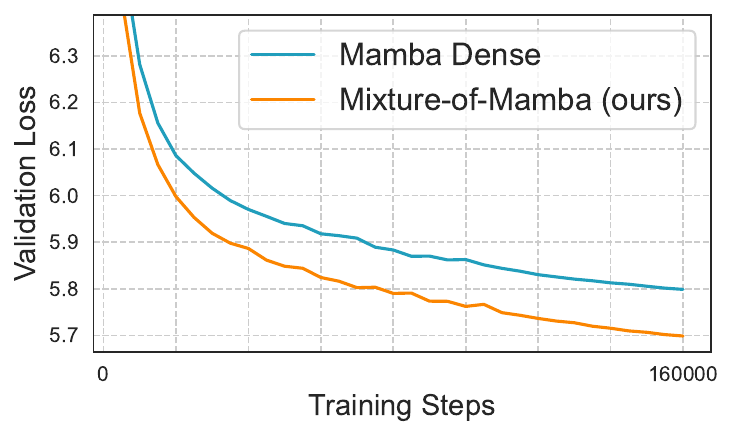}
        \caption{\textbf{880M} Image Eval Loss}
    \end{subfigure}
    \hfill
    \begin{subfigure}[b]{0.24\textwidth}
       \centering
       \includegraphics[width=\textwidth]{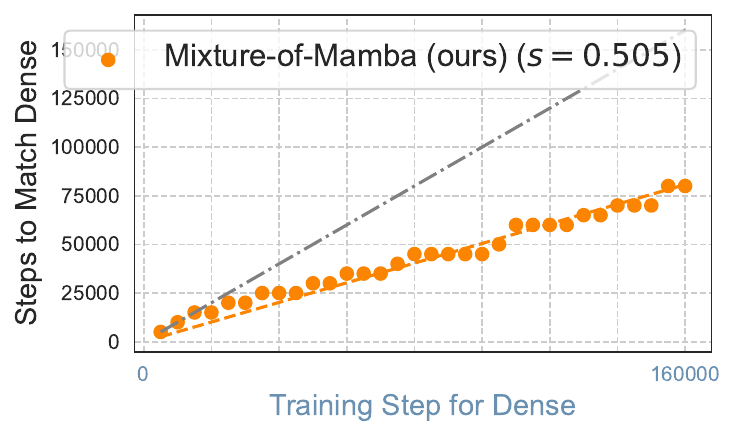}
       \caption{Image Loss Matching}
   \end{subfigure}
   \hfill
   \begin{subfigure}[b]{0.24\textwidth}
        \centering
        \includegraphics[width=\textwidth]{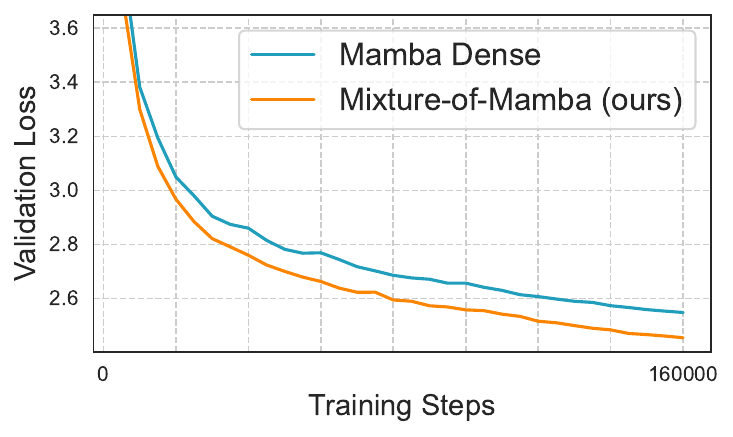}
        \caption{Text Eval Loss}
    \end{subfigure}
    \hfill
    \begin{subfigure}[b]{0.24\textwidth}
       \centering
       \includegraphics[width=\textwidth]{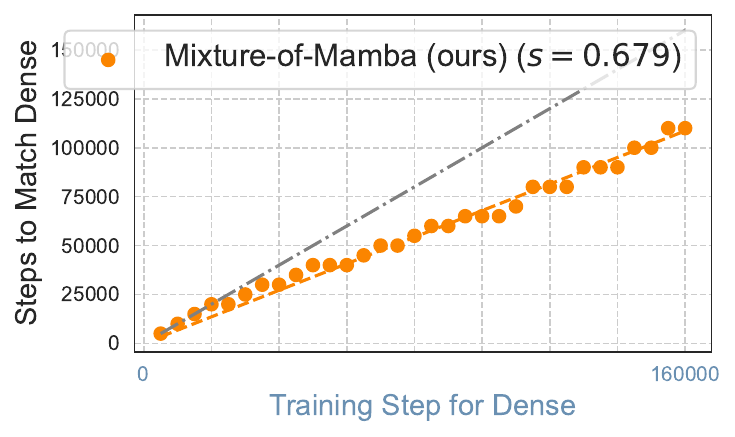}
       \caption{Text Loss Matching}
   \end{subfigure}

    \begin{subfigure}[b]{0.24\textwidth}
        \centering
        \includegraphics[width=\textwidth]{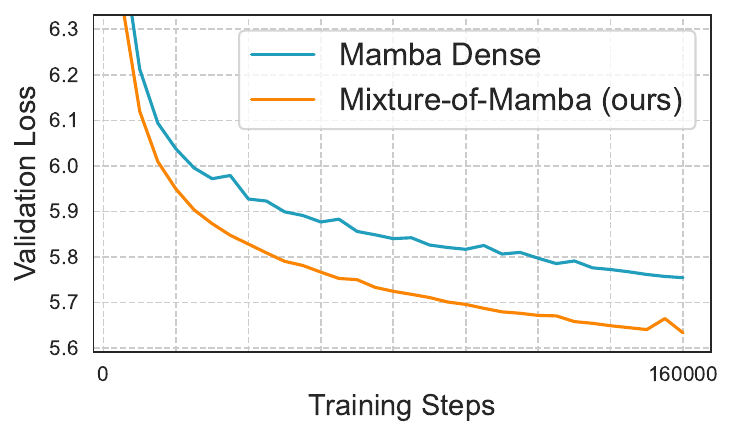}
        \caption{\textbf{1.5B} Image Eval Loss}
    \end{subfigure}
    \hfill
    \begin{subfigure}[b]{0.24\textwidth}
       \centering
       \includegraphics[width=\textwidth]{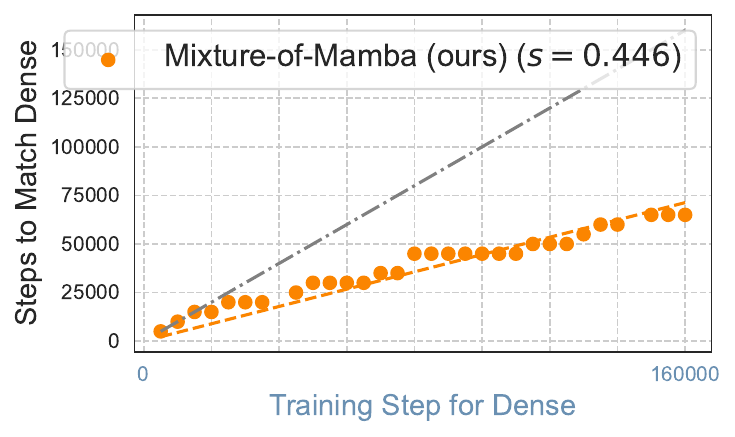}
       \caption{Image Loss Matching}
   \end{subfigure}
   \hfill
   \begin{subfigure}[b]{0.24\textwidth}
        \centering
        \includegraphics[width=\textwidth]{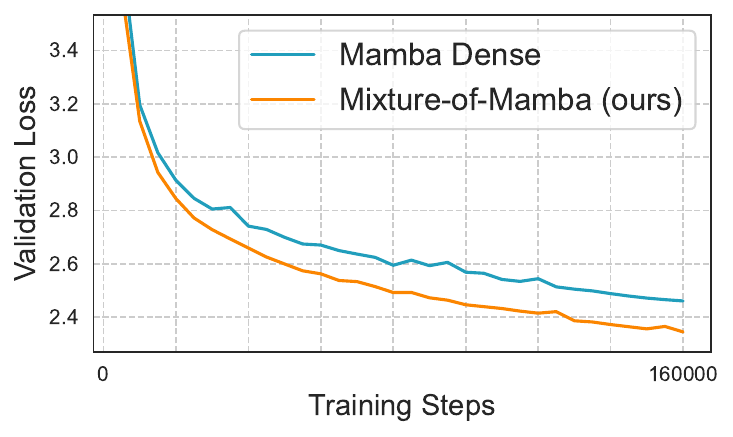}
        \caption{Text Eval Loss}
    \end{subfigure}
    \hfill
    \begin{subfigure}[b]{0.24\textwidth}
       \centering
       \includegraphics[width=\textwidth]{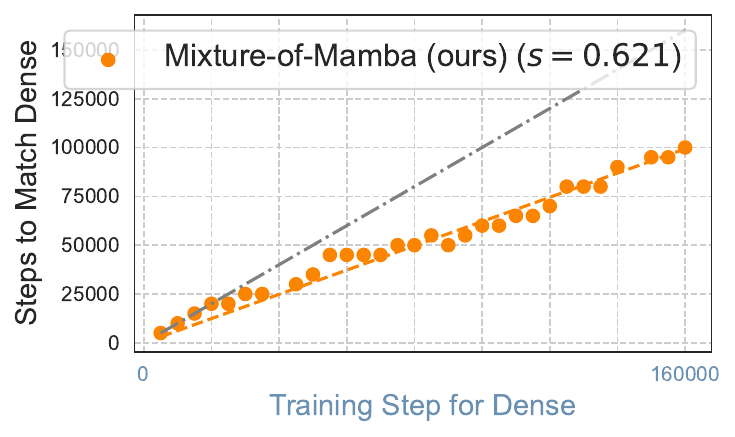}
       \caption{Text Loss Matching}
   \end{subfigure}

\caption{\textbf{Training and evaluation losses for image and text modalities across model scales in the Chameleon setting on the Shutterstock dataset.} 
Results are shown for Mixture-of-Mamba and Mamba Dense across five model scales: \textbf{37M}, \textbf{94M}, \textbf{443M}, \textbf{880M}, and \textbf{1.5B}.  
\textbf{(a, e, i, m, q)} Image evaluation loss demonstrates consistent improvements for Mixture-of-Mamba (\textcolor{orange}{orange}), achieving lower loss compared to Mamba Dense (\textcolor{cyan}{cyan}) across all scales.  
\textbf{(b, f, j, n, r)} Image loss matching shows that Mixture-of-Mamba {reaches the same loss values at earlier training steps} compared to Mamba Dense, reflecting its improved training efficiency.  
\textbf{(c, g, k, o, s)} Text evaluation loss indicates competitive results for Mixture-of-Mamba, achieving lower losses relative to Mamba Dense.  
\textbf{(d, h, l, p, t)} Text loss matching highlights that Mixture-of-Mamba reaches the same loss values at earlier training steps, further demonstrating its efficiency in the text modality.  
Overall, Mixture-of-Mamba achieves strong and consistent improvements in both image and text modalities across all model scales in the Chameleon setting evaluated on the Shutterstock dataset.}
\label{fig:appendix_chameleon_shutterstock}   
\end{figure*}

\begin{figure*}[t]
    \centering
    \begin{subfigure}[b]{0.24\textwidth}
        \centering
        \includegraphics[width=\textwidth]{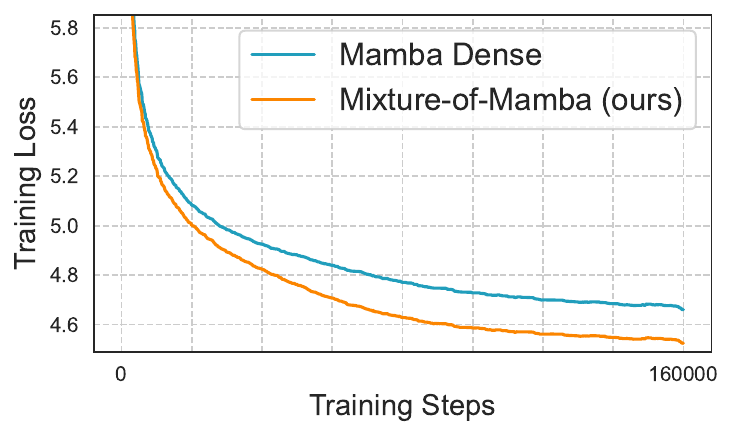}
        \caption{\textbf{37M} Avg Training Loss}
    \end{subfigure}
    \begin{subfigure}[b]{0.24\textwidth}
       \centering
       \includegraphics[width=\textwidth]{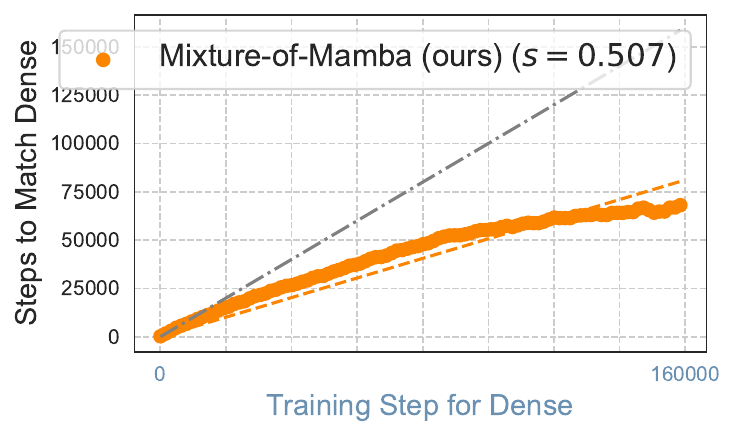}
       \caption{Avg Loss Matching}
   \end{subfigure}

    \begin{subfigure}[b]{0.24\textwidth}
        \centering
        \includegraphics[width=\textwidth]{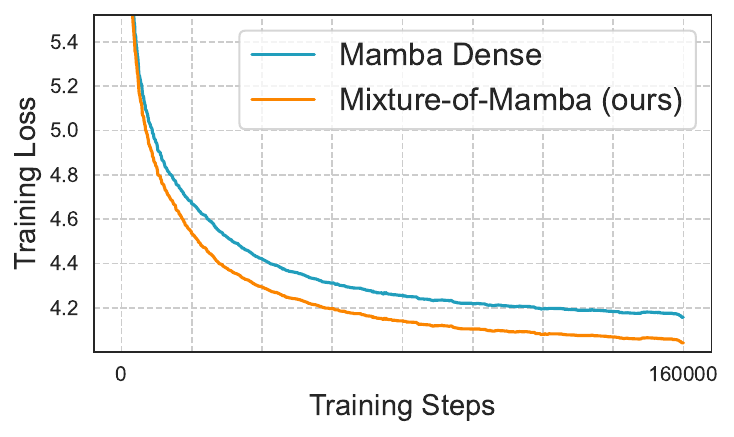}
        \caption{\textbf{94M} Avg Training Loss}
    \end{subfigure}
    \begin{subfigure}[b]{0.24\textwidth}
       \centering
       \includegraphics[width=\textwidth]{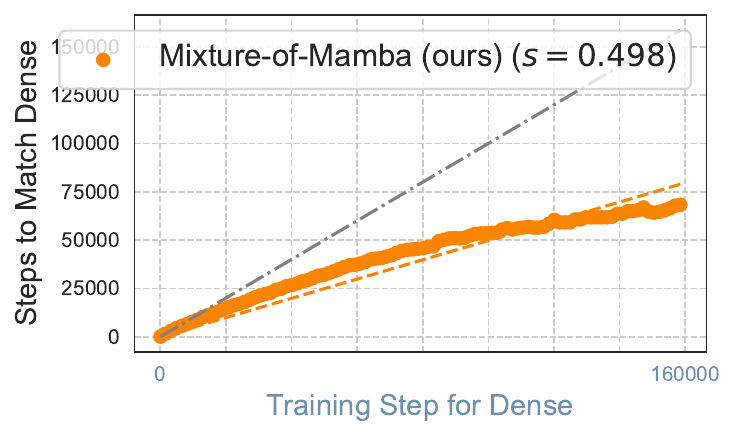}
       \caption{Avg Loss Matching}
   \end{subfigure}

   \begin{subfigure}[b]{0.24\textwidth}
        \centering
        \includegraphics[width=\textwidth]{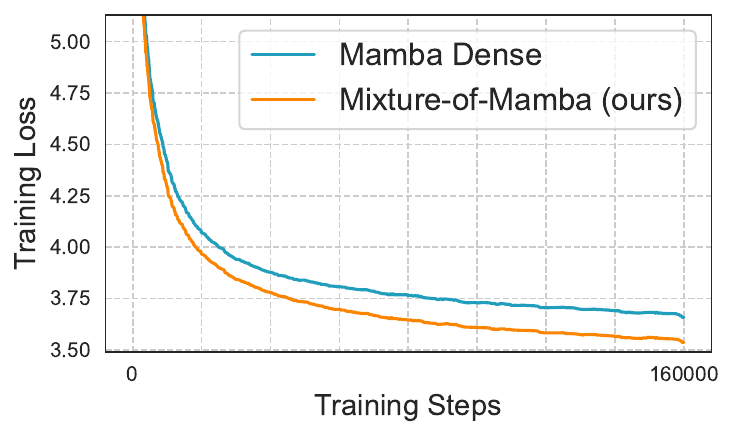}
        \caption{\textbf{443M} Avg Training Loss}
    \end{subfigure}
    \begin{subfigure}[b]{0.24\textwidth}
       \centering
       \includegraphics[width=\textwidth]{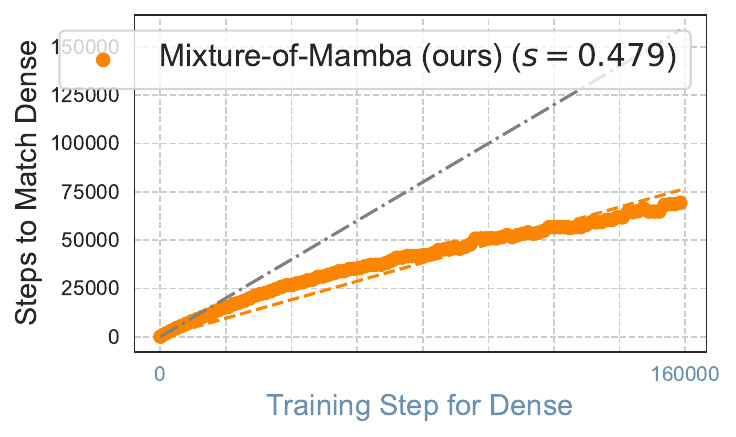}
       \caption{Avg Loss Matching}
   \end{subfigure}

    \begin{subfigure}[b]{0.24\textwidth}
        \centering
        \includegraphics[width=\textwidth]{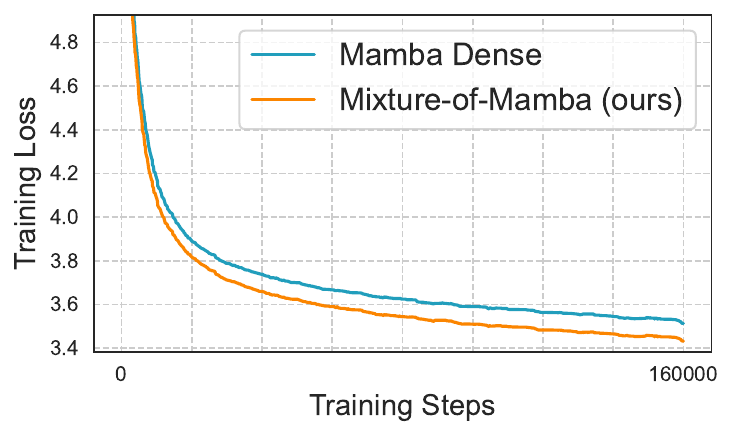}
        \caption{\textbf{880M} Avg Training Loss}
    \end{subfigure}
    \begin{subfigure}[b]{0.24\textwidth}
       \centering
       \includegraphics[width=\textwidth]{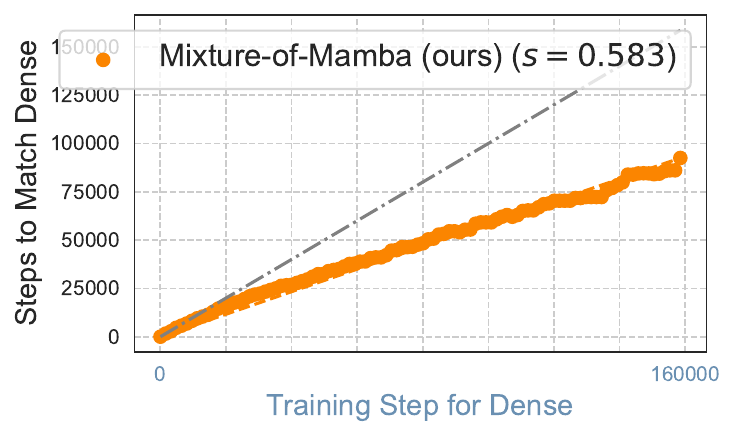}
       \caption{Avg Loss Matching}
   \end{subfigure}

    \begin{subfigure}[b]{0.24\textwidth}
        \centering
        \includegraphics[width=\textwidth]{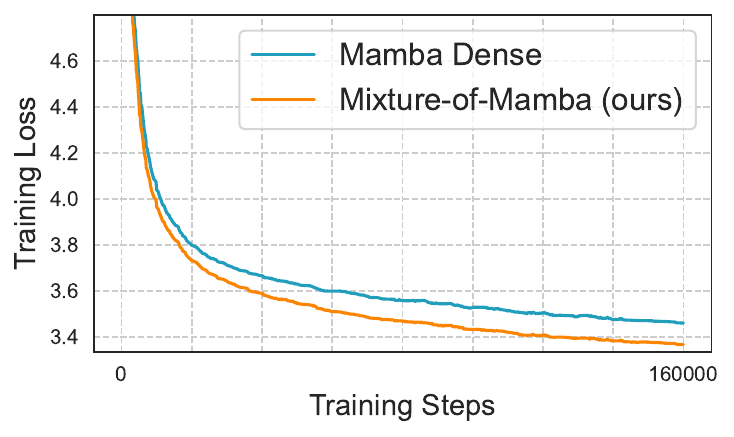}
        \caption{\textbf{1.5B} Avg Training Loss}
    \end{subfigure}
    \begin{subfigure}[b]{0.24\textwidth}
       \centering
       \includegraphics[width=\textwidth]{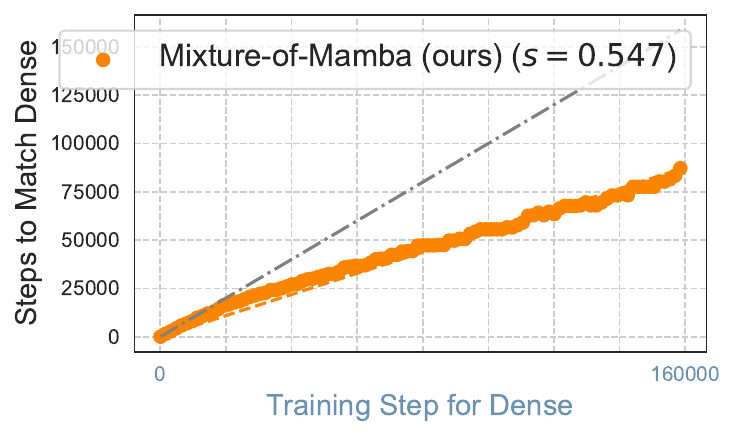}
       \caption{Avg Loss Matching}
   \end{subfigure}

\caption{\textbf{Average training loss and step matching plots across model scales in the Chameleon setting.} 
Results are shown for Mixture-of-Mamba and Mamba Dense across five model scales: \textbf{37M}, \textbf{94M}, \textbf{443M}, \textbf{880M}, and \textbf{1.5B}.  
\textbf{(a, c, e, g, i)} Average training loss (across image and text modalities) demonstrates consistent reductions for Mixture-of-Mamba (\textcolor{orange}{orange}), achieving lower loss values compared to Mamba Dense (\textcolor{cyan}{cyan}) at all model scales.  
\textbf{(b, d, f, h, j)} Average loss matching plots highlight that Mixture-of-Mamba {reaches the same loss values at earlier training steps} compared to Mamba Dense, reflecting improved training efficiency.  
Overall, Mixture-of-Mamba consistently reduces average training loss and achieves more efficient convergence across all model scales in the Chameleon setting.}
\label{fig:appendix_avg_training_loss_Chameleon}

\end{figure*}

\begin{figure*}
    \centering

    \begin{subfigure}[b]{0.24\textwidth}
        \centering
        \includegraphics[width=\textwidth]{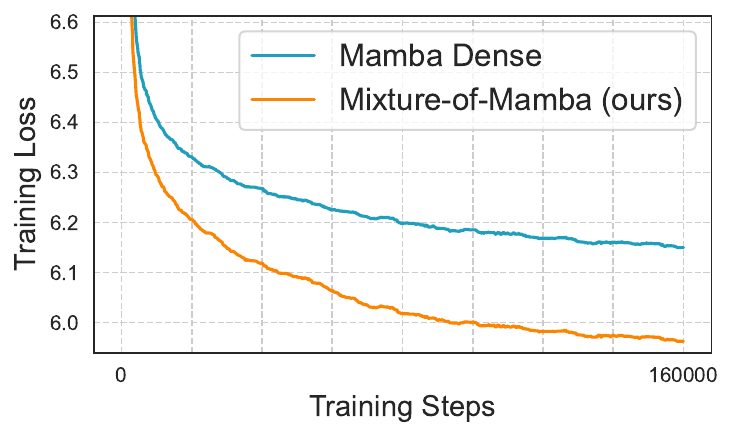}
        
        \caption{\footnotesize \textbf{37M} Image Training Loss}
    \end{subfigure}
    \hfill
    \begin{subfigure}[b]{0.24\textwidth}
        \centering
        \includegraphics[width=\textwidth]{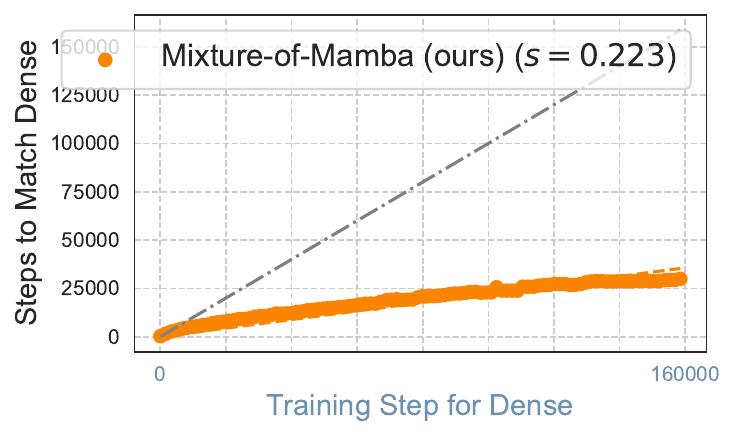}
        \caption{\footnotesize Image Loss Matching}
    \end{subfigure}
    \hfill 
    \begin{subfigure}[b]{0.24\textwidth}
        \centering
        \includegraphics[width=\textwidth]{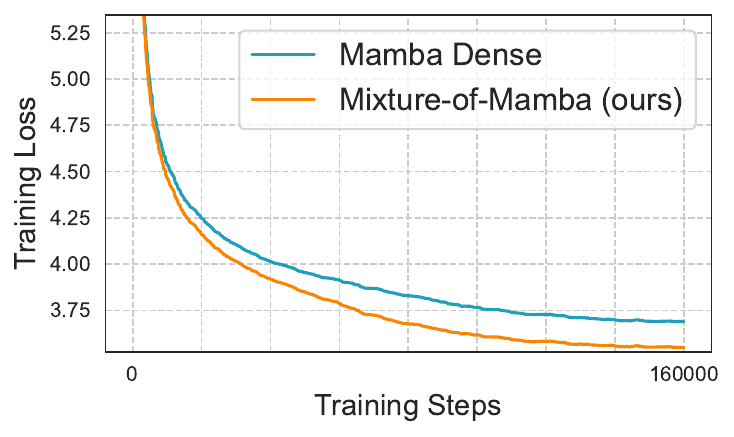}
        \caption{\footnotesize Text Training Loss}
    \end{subfigure}
    \hfill
    \begin{subfigure}[b]{0.24\textwidth}
        \centering
        \includegraphics[width=\textwidth]{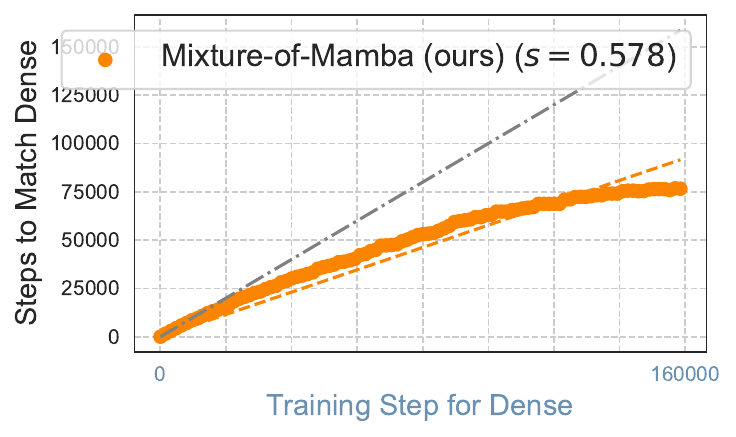}
        \caption{\footnotesize Text Loss Matching}
    \end{subfigure}

   \begin{subfigure}[b]{0.24\textwidth}
       \centering
       \includegraphics[width=\textwidth]{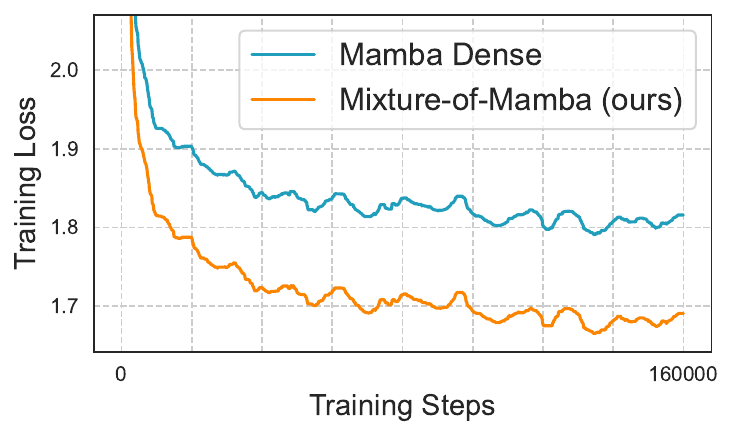}
       \caption{\footnotesize \textbf{37M} Speech Training Loss}
   \end{subfigure}
   \hfill
   \begin{subfigure}[b]{0.24\textwidth}
       \centering
       \includegraphics[width=\textwidth]{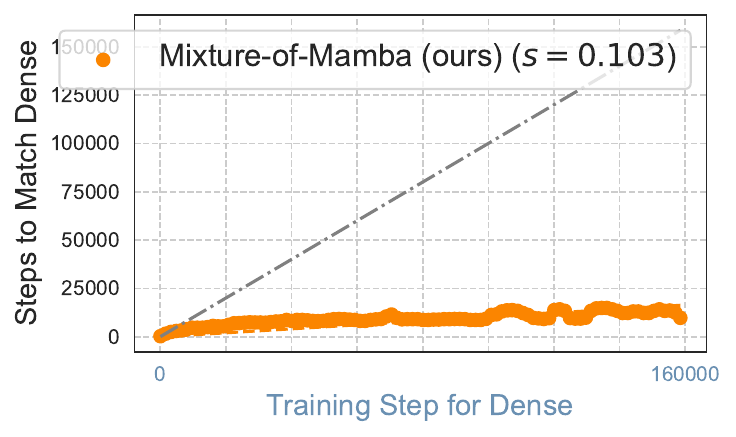}
       \caption{\footnotesize Speech Loss Matching}
   \end{subfigure}
   \hfill 
    \begin{subfigure}[b]{0.24\textwidth}
        \centering
        \includegraphics[width=\textwidth]{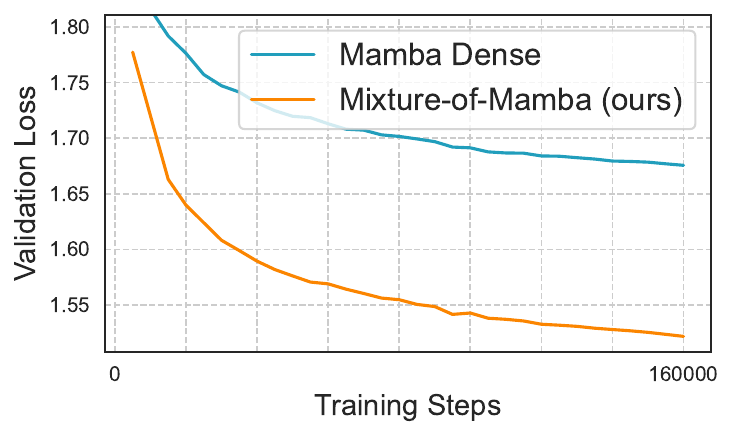}
        \caption{\footnotesize Speech Eval Loss: LL60K}
    \end{subfigure}
    \hfill
    \begin{subfigure}[b]{0.24\textwidth}
        \centering
        \includegraphics[width=\textwidth]{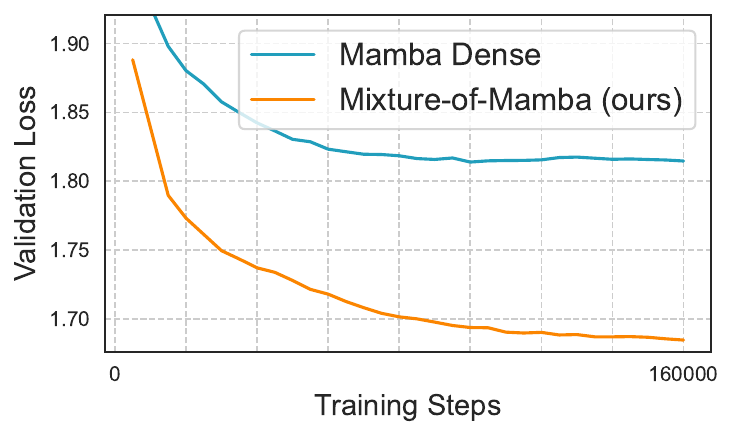}
        \caption{\footnotesize Speech Eval Loss: PPL30K}
    \end{subfigure}

    \begin{subfigure}[b]{0.24\textwidth}
        \centering
        \includegraphics[width=\textwidth]{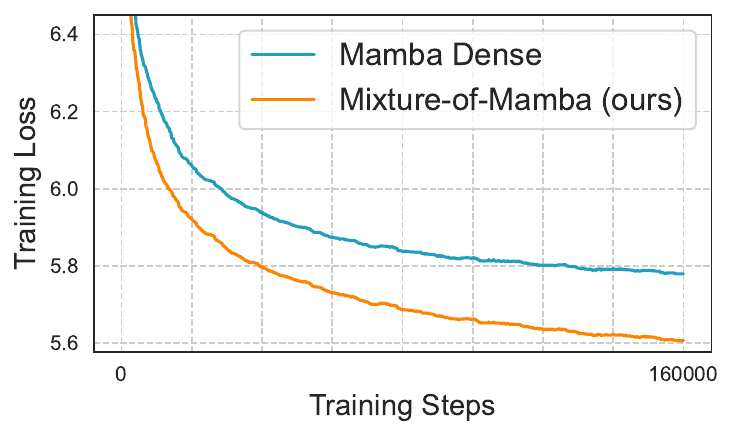}
        
        \caption{\footnotesize \textbf{94M} Image Training Loss}
    \end{subfigure}
    \hfill
    \begin{subfigure}[b]{0.24\textwidth}
        \centering
        \includegraphics[width=\textwidth]{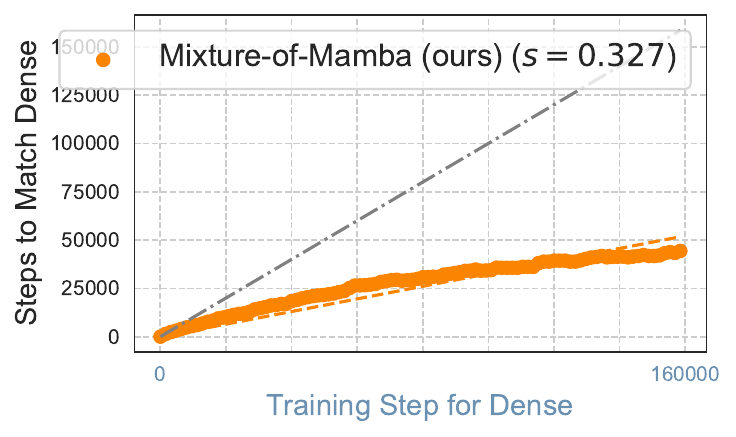}
        \caption{\footnotesize Image Loss Matching}
    \end{subfigure}
    \hfill 
    \begin{subfigure}[b]{0.24\textwidth}
        \centering
        \includegraphics[width=\textwidth]{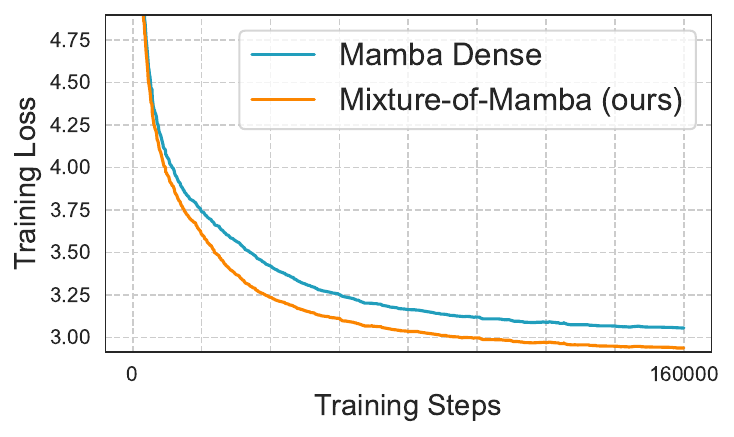}
        \caption{\footnotesize Text Training Loss}
    \end{subfigure}
    \hfill
    \begin{subfigure}[b]{0.24\textwidth}
        \centering
        \includegraphics[width=\textwidth]{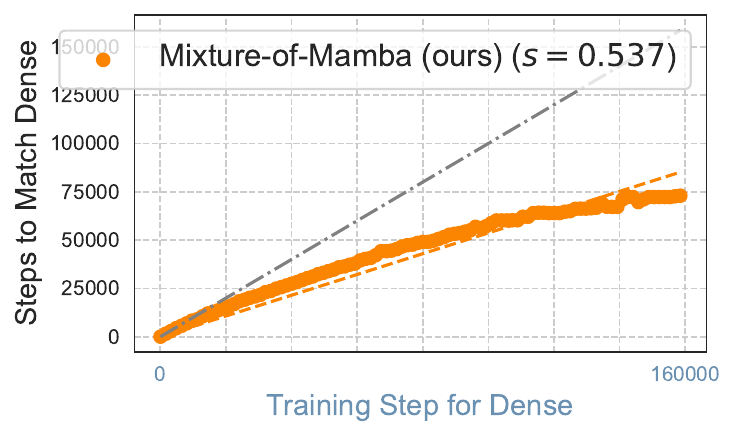}
        \caption{\footnotesize Text Loss Matching}
    \end{subfigure}

   \begin{subfigure}[b]{0.24\textwidth}
       \centering
       \includegraphics[width=\textwidth]{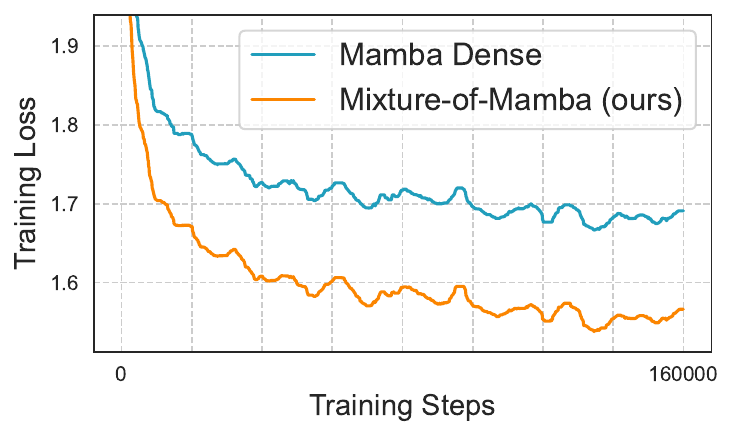}
       \caption{\footnotesize \textbf{94M} Speech Training Loss}
   \end{subfigure}
   \hfill
   \begin{subfigure}[b]{0.24\textwidth}
       \centering
       \includegraphics[width=\textwidth]{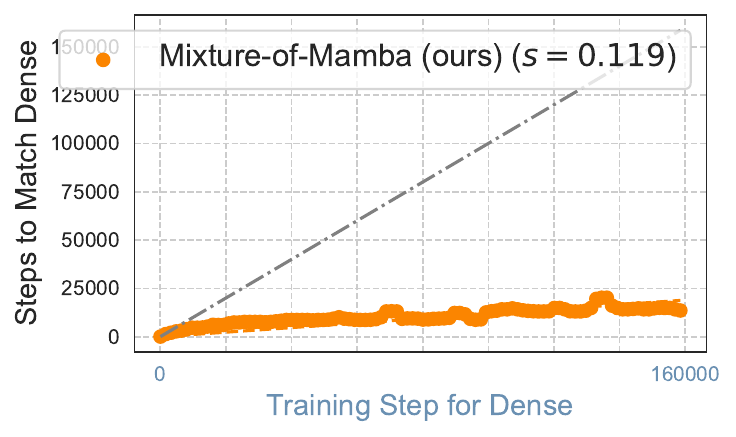}
       \caption{\footnotesize Speech Loss Matching}
   \end{subfigure}
   \hfill 
    \begin{subfigure}[b]{0.24\textwidth}
        \centering
        \includegraphics[width=\textwidth]{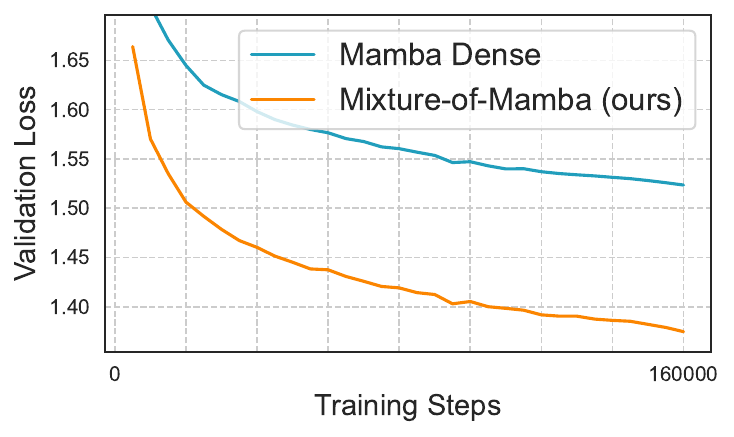}
        \caption{\footnotesize Speech Eval Loss: LL60K}
    \end{subfigure}
    \hfill
    \begin{subfigure}[b]{0.24\textwidth}
        \centering
        \includegraphics[width=\textwidth]{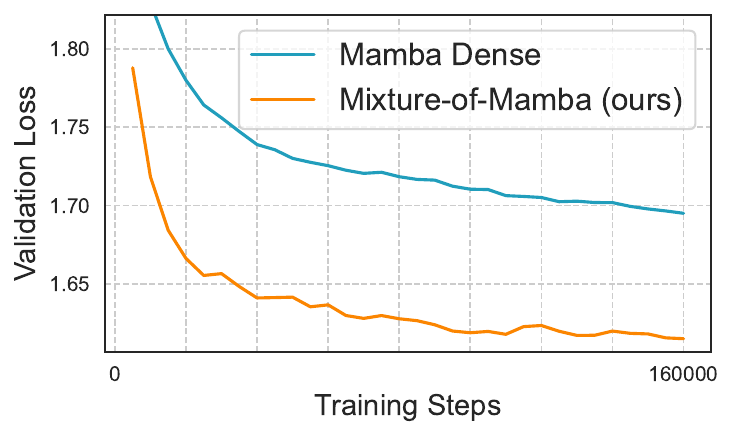}
        \caption{\footnotesize Speech Eval Loss: PPL30K}
    \end{subfigure}

\caption{\textbf{Training and evaluation losses for image, text, and speech modalities (37M and 94M scales) in the Chameleon+Speech setting.} 
Results are reported for Mixture-of-Mamba and Mamba Dense.  
\textbf{(a, e, i)} Image training loss demonstrates that Mixture-of-Mamba (\textcolor{orange}{orange}) achieves consistently lower loss compared to Mamba Dense (\textcolor{cyan}{cyan}).  
\textbf{(b, f, j)} Image loss matching highlights Mixture-of-Mamba’s ability to reach the same loss values at earlier training steps, showing improved training efficiency.  
\textbf{(c, g, k)} Text training loss shows competitive results for Mixture-of-Mamba, improving over Mamba Dense.  
\textbf{(d, h, l)} Text loss matching confirms Mixture-of-Mamba’s ability to reach the same loss values at earlier training steps, showing improved training efficiency.   
\textbf{(e, m)} Speech training loss highlights significant improvements in speech modality performance.
\textbf{(f, n)} Speech loss matching shows efficient learning dynamics for Mixture-of-Mamba.  
\textbf{(g, o)} Speech evaluation loss on LL60K confirms notable performance gains, and  
\textbf{(h, p)} Speech evaluation loss on PPL30K further highlights the efficiency of Mixture-of-Mamba.}
\label{fig:appendix_chameleon_speech_part1}
\end{figure*}

\begin{figure*}
    \centering

    \begin{subfigure}[b]{0.24\textwidth}
        \centering
        \includegraphics[width=\textwidth]{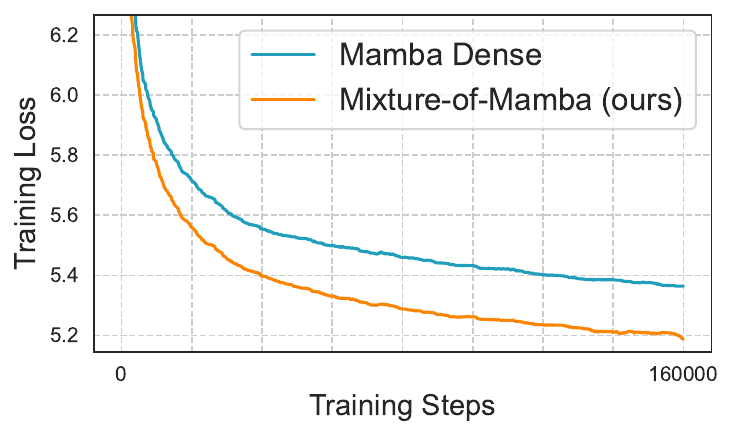}
        
        \caption{\footnotesize \textbf{443M} Image Training Loss}
    \end{subfigure}
    \hfill
    \begin{subfigure}[b]{0.24\textwidth}
        \centering
        \includegraphics[width=\textwidth]{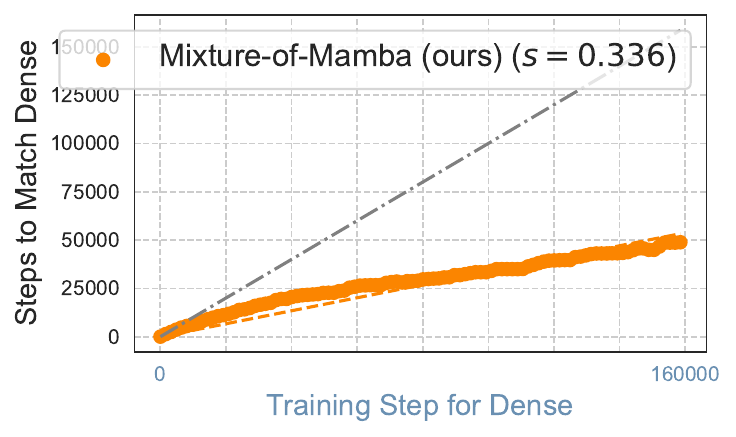}
        \caption{\footnotesize Image Loss Matching}
    \end{subfigure}
    \hfill 
    \begin{subfigure}[b]{0.24\textwidth}
        \centering
        \includegraphics[width=\textwidth]{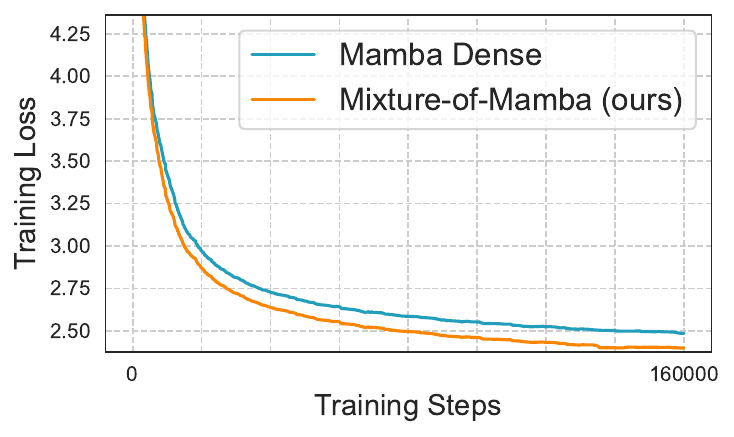}
        \caption{\footnotesize Text Training Loss}
    \end{subfigure}
    \hfill
    \begin{subfigure}[b]{0.24\textwidth}
        \centering
        \includegraphics[width=\textwidth]{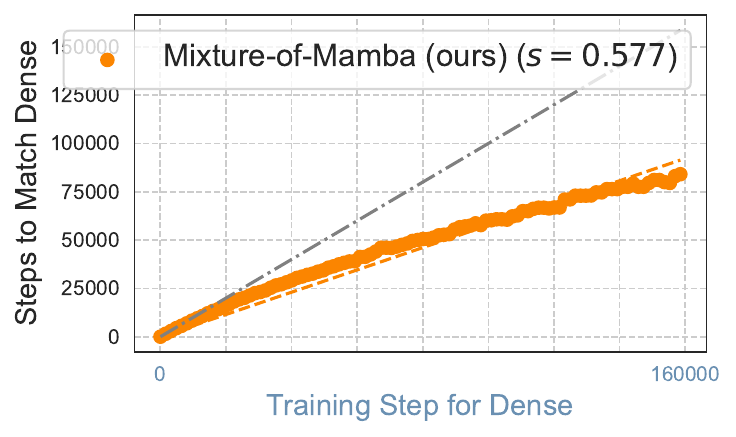}
        \caption{\footnotesize Text Loss Matching}
    \end{subfigure}

   \begin{subfigure}[b]{0.24\textwidth}
       \centering
       \includegraphics[width=\textwidth]{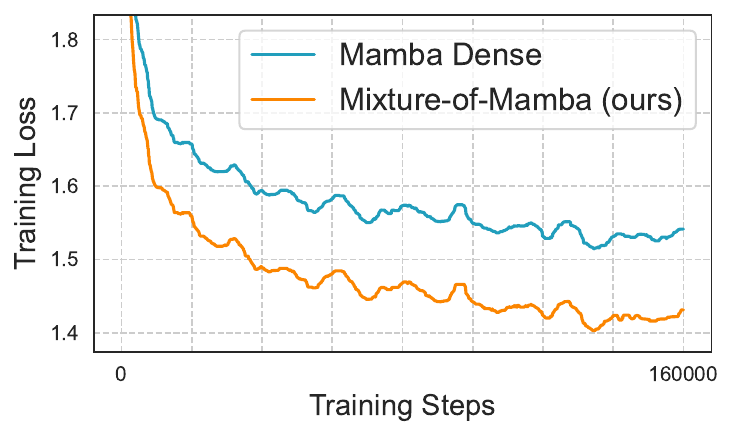}
       \caption{\footnotesize \textbf{443M} Speech Training Loss}
   \end{subfigure}
   \hfill
   \begin{subfigure}[b]{0.24\textwidth}
       \centering
       \includegraphics[width=\textwidth]{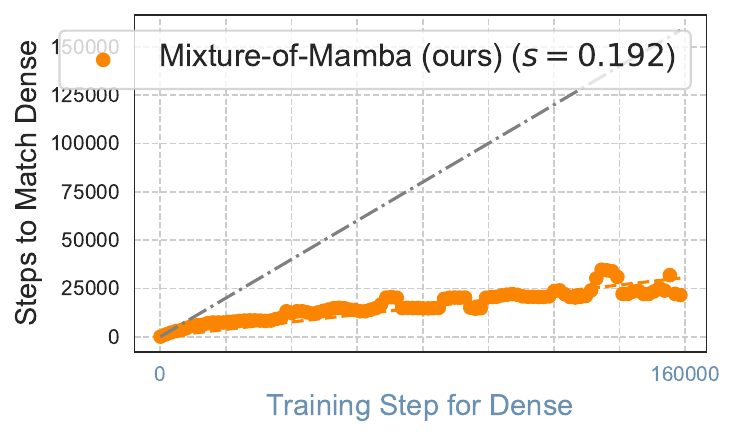}
       \caption{\footnotesize Speech Loss Matching}
   \end{subfigure}
   \hfill 
    \begin{subfigure}[b]{0.24\textwidth}
        \centering
        \includegraphics[width=\textwidth]{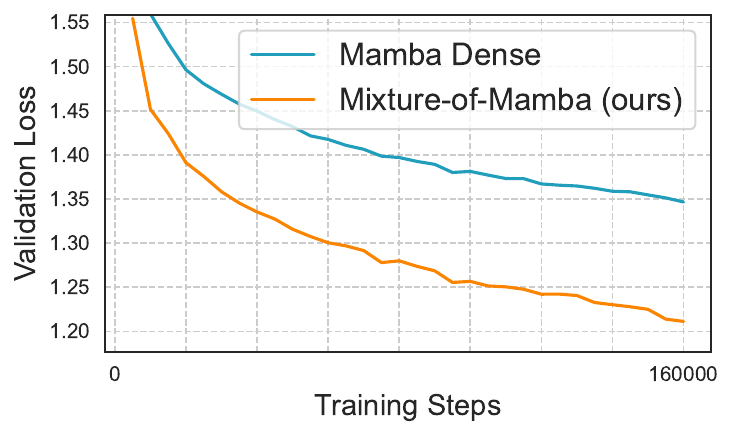}
        \caption{\footnotesize Speech Eval Loss: LL60K}
    \end{subfigure}
    \hfill
    \begin{subfigure}[b]{0.24\textwidth}
        \centering
        \includegraphics[width=\textwidth]{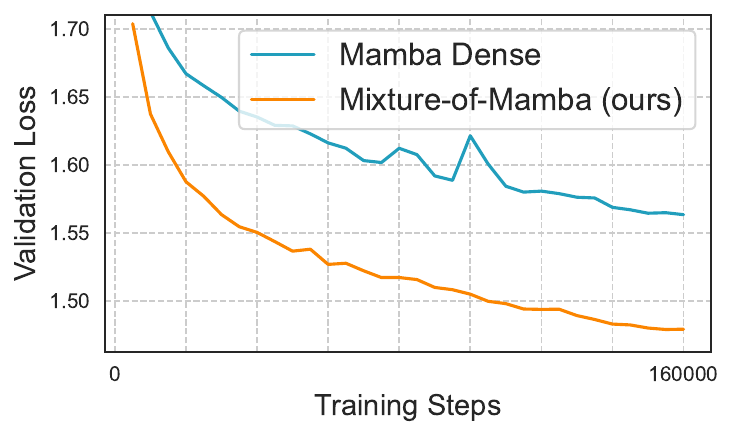}
        \caption{\footnotesize Speech Eval Loss: PPL30K}
    \end{subfigure}

    \begin{subfigure}[b]{0.24\textwidth}
        \centering
        \includegraphics[width=\textwidth]{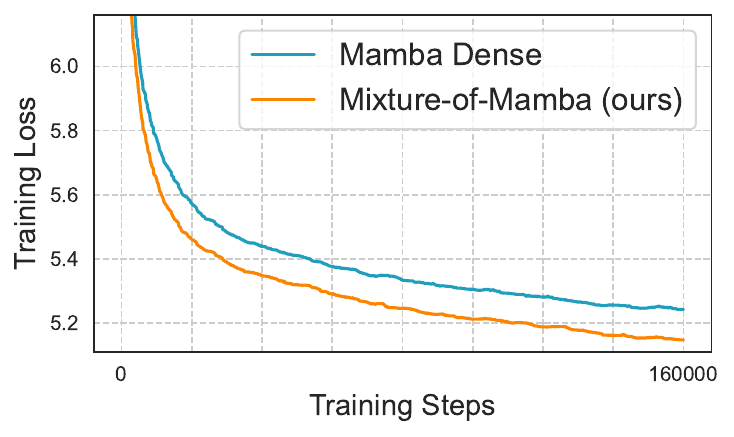}
        
        \caption{\footnotesize \textbf{880M} Image Training Loss}
    \end{subfigure}
    \hfill
    \begin{subfigure}[b]{0.24\textwidth}
        \centering
        \includegraphics[width=\textwidth]{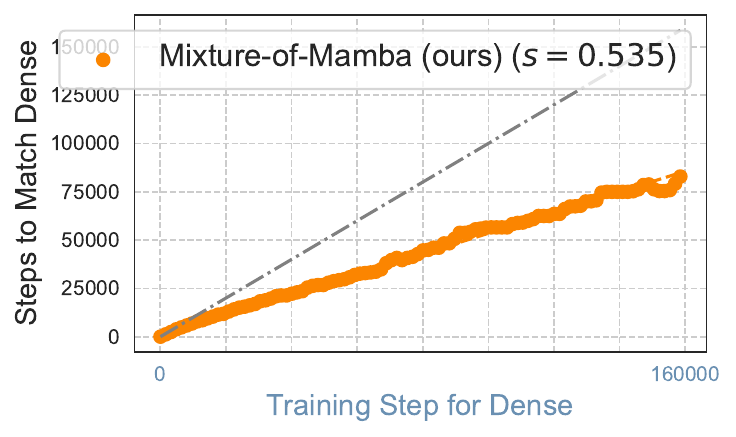}
        \caption{\footnotesize Image Loss Matching}
    \end{subfigure}
    \hfill 
    \begin{subfigure}[b]{0.24\textwidth}
        \centering
        \includegraphics[width=\textwidth]{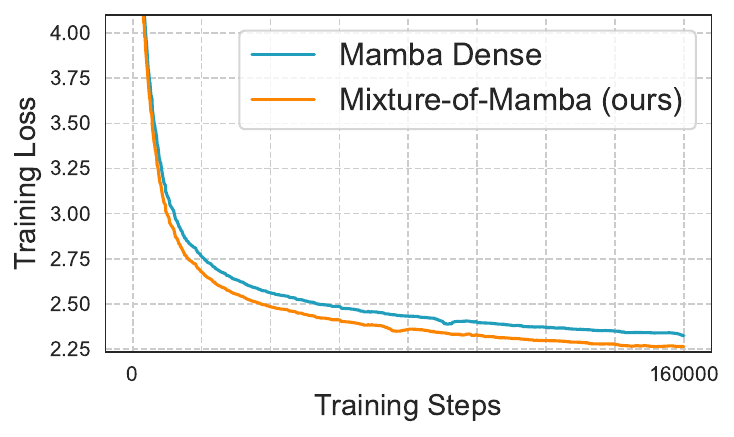}
        \caption{\footnotesize Text Training Loss}
    \end{subfigure}
    \hfill
    \begin{subfigure}[b]{0.24\textwidth}
        \centering
        \includegraphics[width=\textwidth]{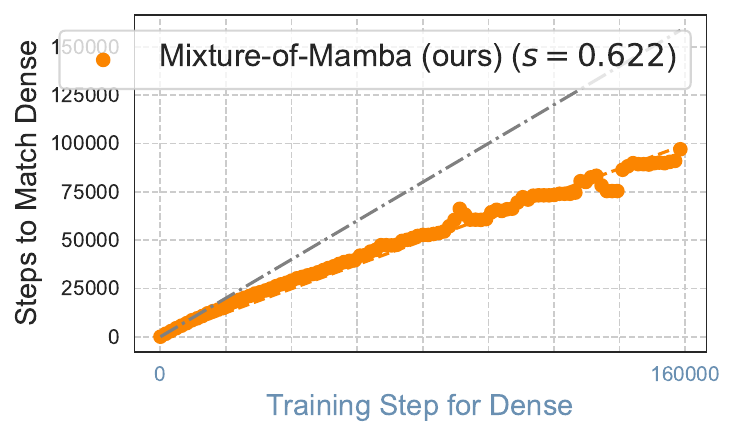}
        \caption{\footnotesize Text Loss Matching}
    \end{subfigure}

   \begin{subfigure}[b]{0.24\textwidth}
       \centering
       \includegraphics[width=\textwidth]{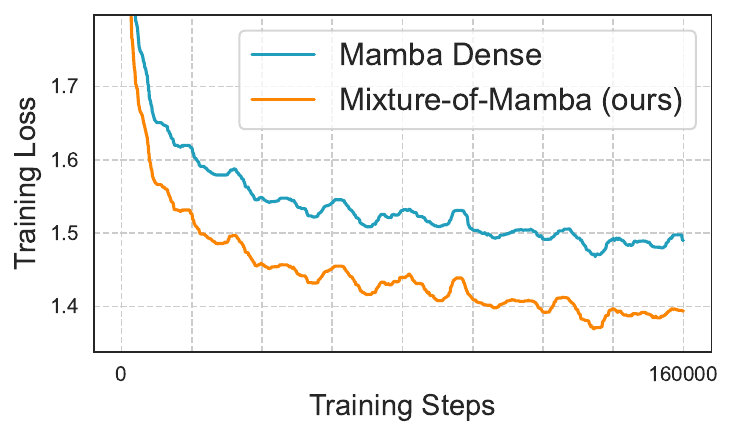}
       \caption{\footnotesize \textbf{880M} Speech Training Loss}
   \end{subfigure}
   \hfill
   \begin{subfigure}[b]{0.24\textwidth}
       \centering
       \includegraphics[width=\textwidth]{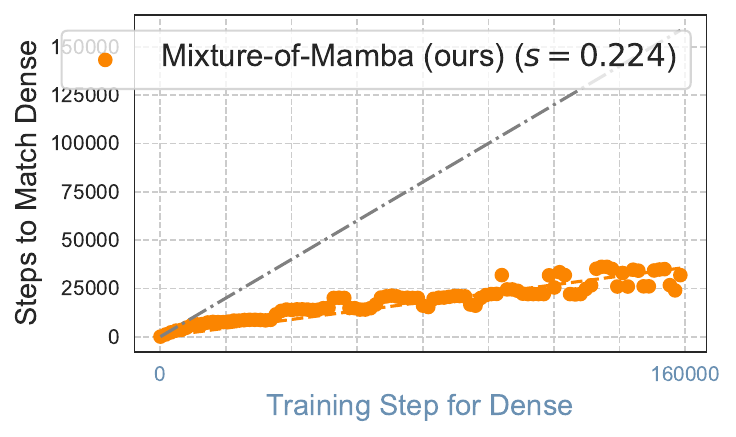}
       \caption{\footnotesize Speech Loss Matching}
   \end{subfigure}
   \hfill 
    \begin{subfigure}[b]{0.24\textwidth}
        \centering
        \includegraphics[width=\textwidth]{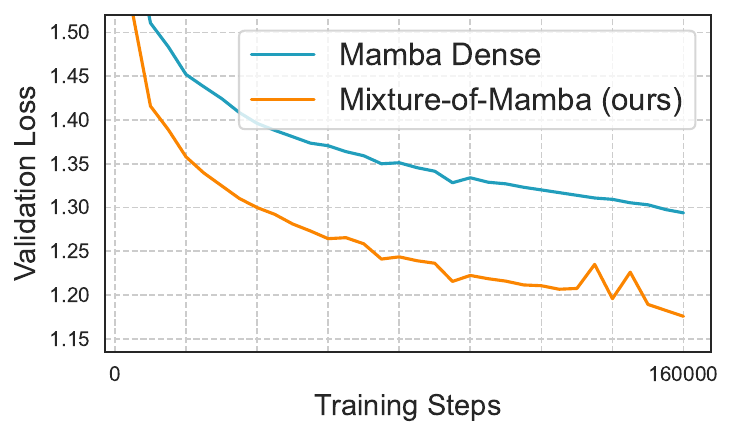}
        \caption{\footnotesize Speech Eval Loss: LL60K}
    \end{subfigure}
    \hfill
    \begin{subfigure}[b]{0.24\textwidth}
        \centering
        \includegraphics[width=\textwidth]{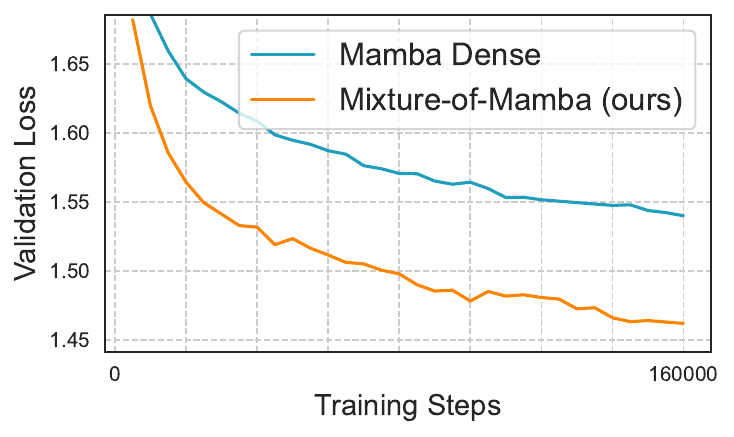}
        \caption{\footnotesize Speech Eval Loss: PPL30K}
    \end{subfigure}

    \begin{subfigure}[b]{0.24\textwidth}
        \centering
        \includegraphics[width=\textwidth]{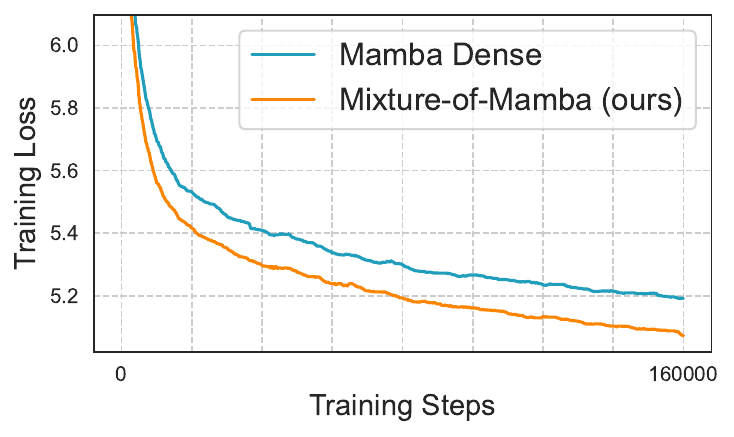}
        
        \caption{\footnotesize \textbf{1.5B} Image Training Loss}
    \end{subfigure}
    \hfill
    \begin{subfigure}[b]{0.24\textwidth}
        \centering
        \includegraphics[width=\textwidth]{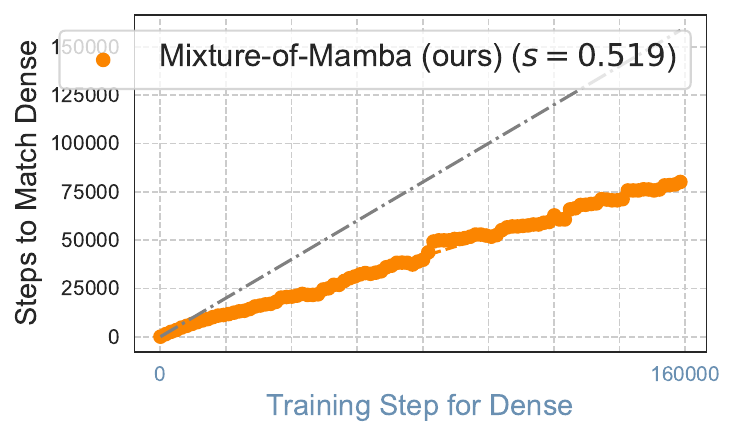}
        \caption{\footnotesize Image Loss Matching}
    \end{subfigure}
    \hfill 
    \begin{subfigure}[b]{0.24\textwidth}
        \centering
        \includegraphics[width=\textwidth]{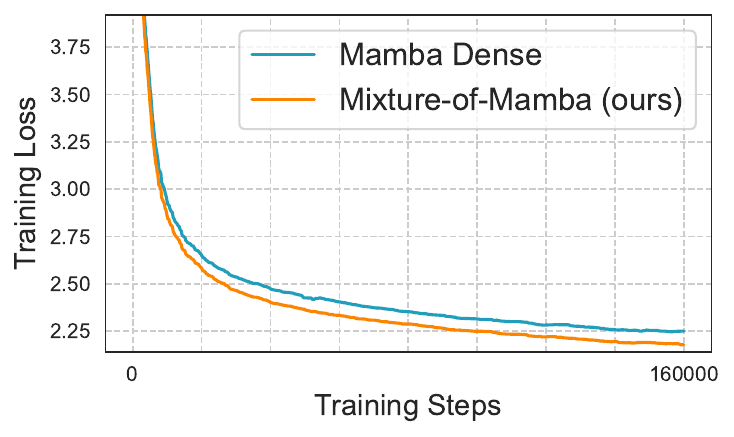}
        \caption{\footnotesize Text Training Loss}
    \end{subfigure}
    \hfill
    \begin{subfigure}[b]{0.24\textwidth}
        \centering
        \includegraphics[width=\textwidth]{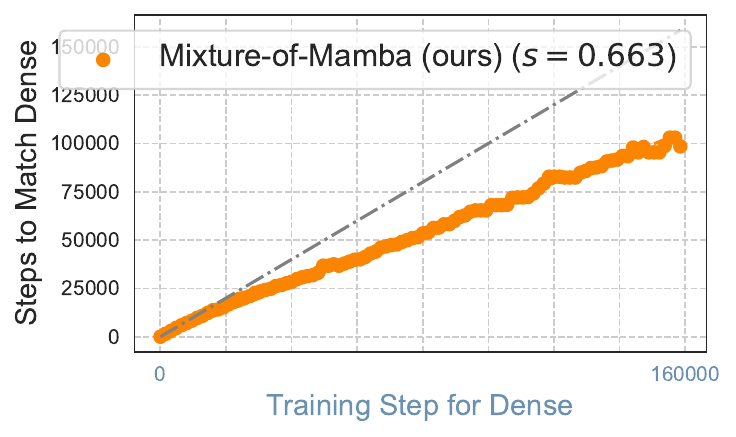}
        \caption{\footnotesize Text Loss Matching}
    \end{subfigure}

   \begin{subfigure}[b]{0.24\textwidth}
       \centering
       \includegraphics[width=\textwidth]{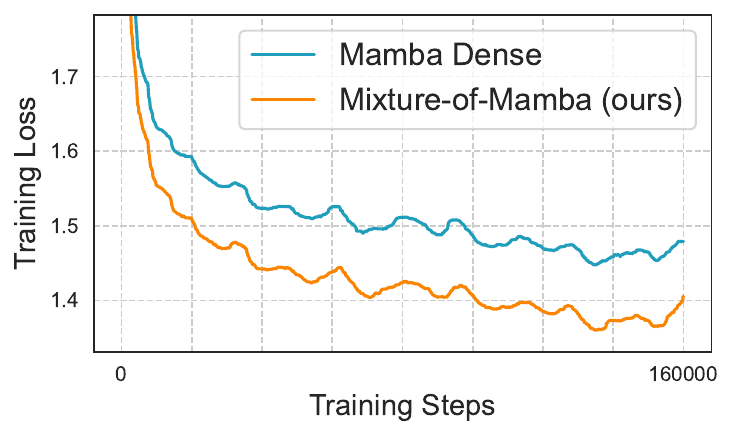}
       \caption{\footnotesize \textbf{1.5B} Speech Training Loss}
   \end{subfigure}
   \hfill
   \begin{subfigure}[b]{0.24\textwidth}
       \centering
       \includegraphics[width=\textwidth]{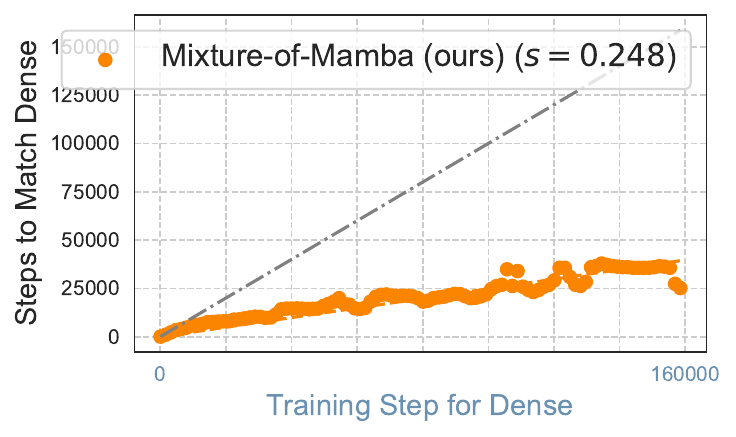}
       \caption{\footnotesize Speech Loss Matching}
   \end{subfigure}
   \hfill 
    \begin{subfigure}[b]{0.24\textwidth}
        \centering
        \includegraphics[width=\textwidth]{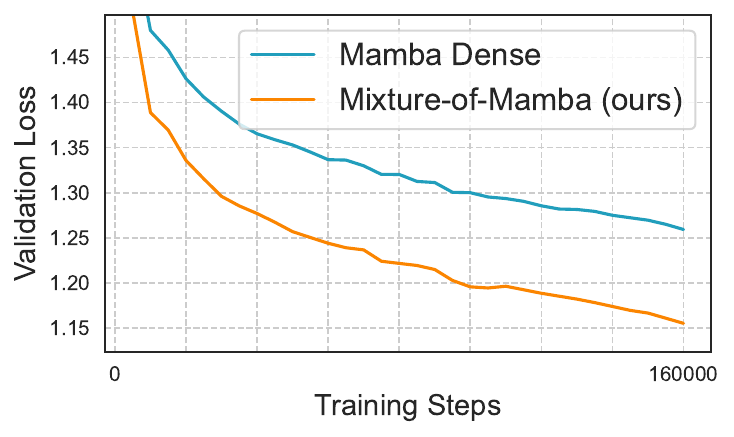}
        \caption{\footnotesize Speech Eval Loss: LL60K}
    \end{subfigure}
    \hfill
    \begin{subfigure}[b]{0.24\textwidth}
        \centering
        \includegraphics[width=\textwidth]{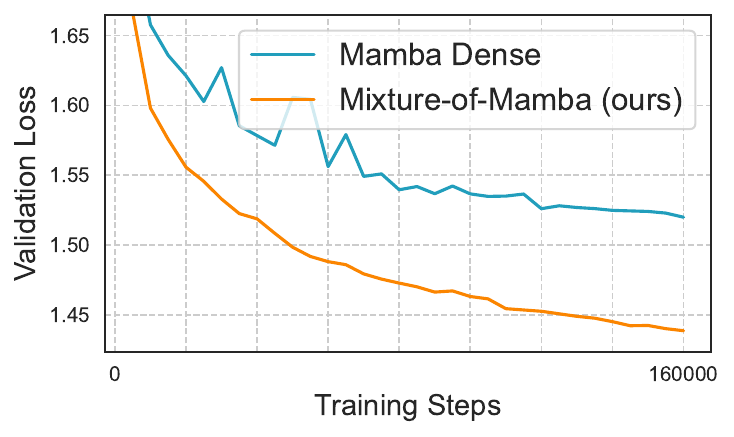}
        \caption{\footnotesize Speech Eval Loss: PPL30K}
    \end{subfigure}

\caption{\textbf{Training and evaluation losses for image, text, and speech modalities (443M, 880M, and 1.5B scales) in the Chameleon+Speech setting.} 
Results are reported for Mixture-of-Mamba and Mamba Dense.  
\textbf{(a, i, q)} Image training loss demonstrates that Mixture-of-Mamba (\textcolor{orange}{orange}) consistently outperforms Mamba Dense (\textcolor{cyan}{cyan}) across larger scales.  
\textbf{(b, j, r)} Image loss matching highlights improved training efficiency for Mixture-of-Mamba, reaching the same loss values at earlier training steps.  
\textbf{(c, k, s)} Text training loss shows Mixture-of-Mamba achieving better performance.  
\textbf{(d, l, t)} Text loss matching further demonstrates efficient learning dynamics.  
\textbf{(e, m, u)} Speech training loss confirms substantial gains for Mixture-of-Mamba in the speech modality, consistent across model scales.  
\textbf{(f, n, v)} Speech loss matching illustrates the improved efficiency of Mixture-of-Mamba across scales.  
\textbf{(g, o, w)} Speech evaluation loss on LL60K highlights consistent improvements, while  
\textbf{(h, p, x)} Speech evaluation loss on PPL30K demonstrates notable gains and efficient performance across scales.}
\label{fig:appendix_chameleon_speech_part2}

\end{figure*}

\begin{figure*}[t]
    \centering
    \begin{subfigure}[b]{0.24\textwidth}
        \centering
        \includegraphics[width=\textwidth]{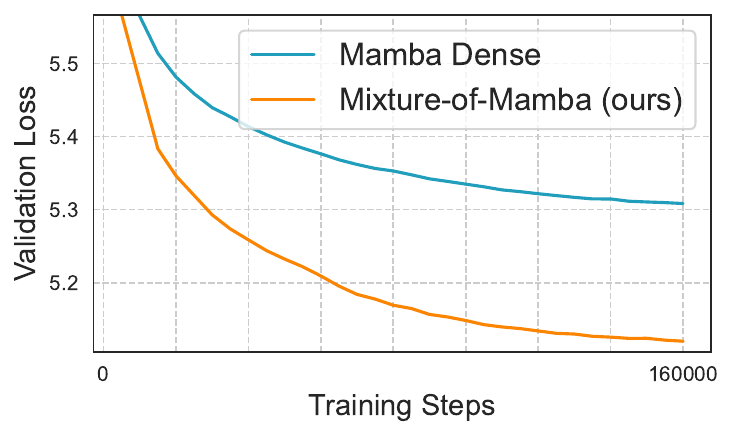}
        \caption{\textbf{37M} Image Eval Loss}
    \end{subfigure}
    \hfill
    \begin{subfigure}[b]{0.24\textwidth}
       \centering
       \includegraphics[width=\textwidth]{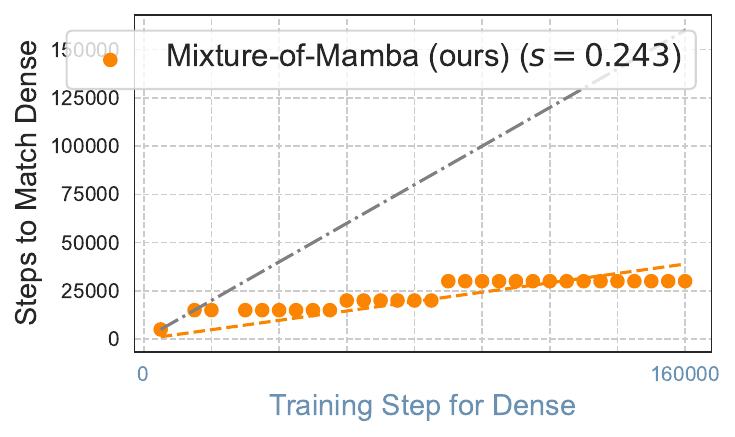}
       \caption{Image Loss Matching}
   \end{subfigure}
   \hfill
   \begin{subfigure}[b]{0.24\textwidth}
        \centering
        \includegraphics[width=\textwidth]{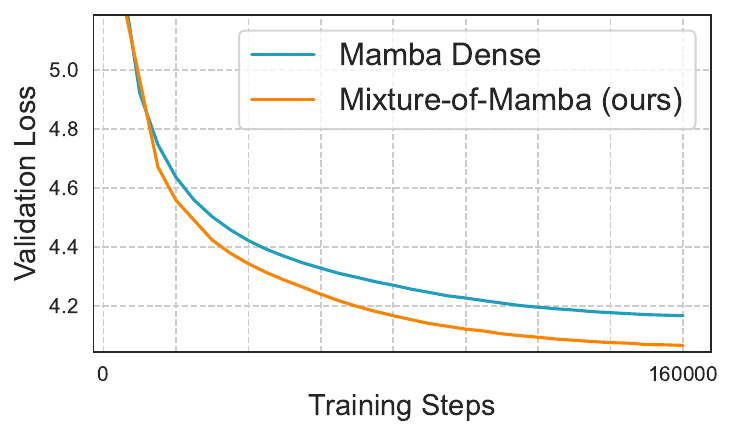}
        \caption{Text Eval Loss}
    \end{subfigure}
    \hfill
    \begin{subfigure}[b]{0.24\textwidth}
       \centering
       \includegraphics[width=\textwidth]{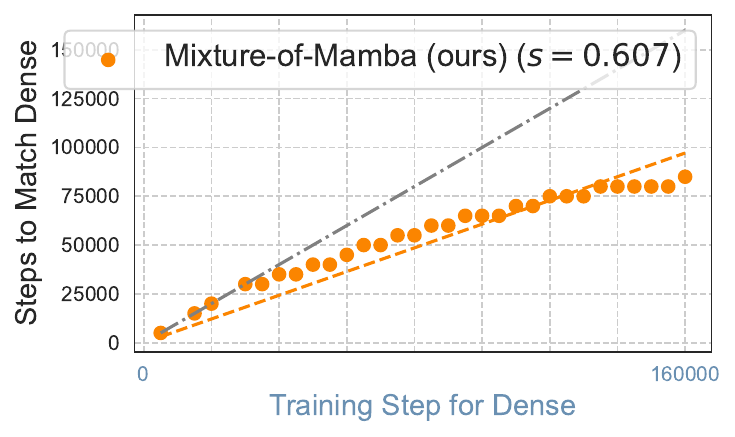}
       \caption{Text Loss Matching}
   \end{subfigure}
    \begin{subfigure}[b]{0.24\textwidth}
        \centering
        \includegraphics[width=\textwidth]{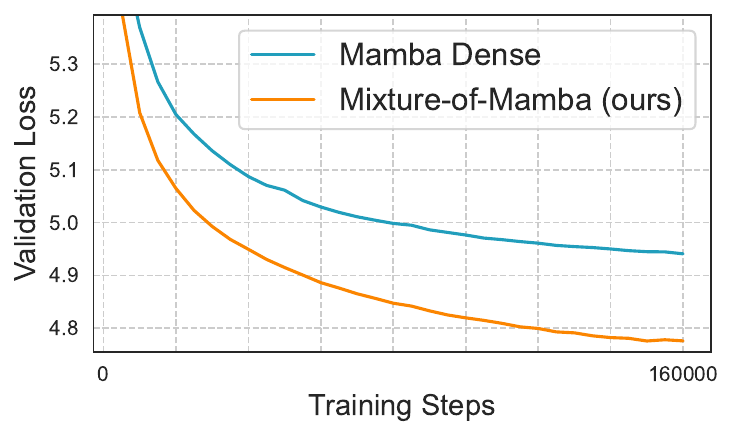}
        \caption{\textbf{94M} Image Eval Loss}
    \end{subfigure}
    \hfill
    \begin{subfigure}[b]{0.24\textwidth}
       \centering
       \includegraphics[width=\textwidth]{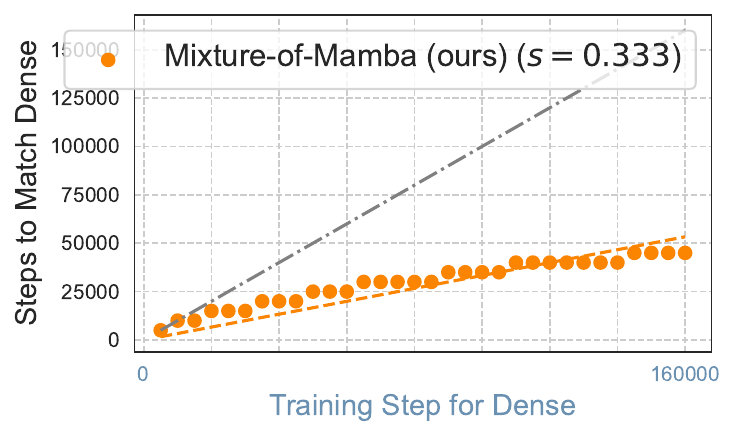}
       \caption{Image Loss Matching}
   \end{subfigure}
   \hfill
   \begin{subfigure}[b]{0.24\textwidth}
        \centering
        \includegraphics[width=\textwidth]{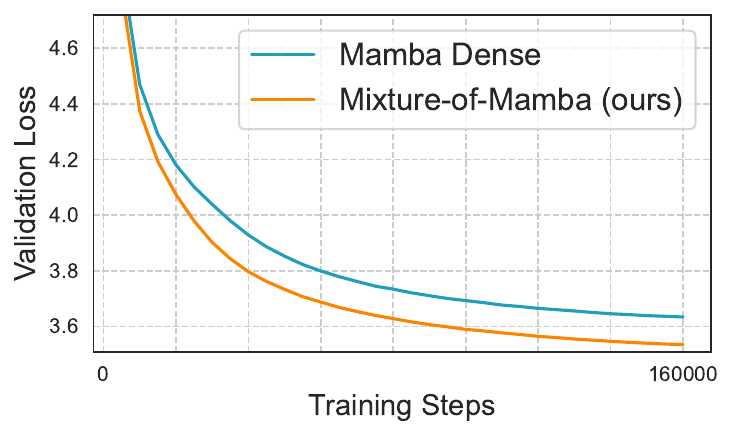}
        \caption{Text Eval Loss}
    \end{subfigure}
    \hfill
    \begin{subfigure}[b]{0.24\textwidth}
       \centering
       \includegraphics[width=\textwidth]{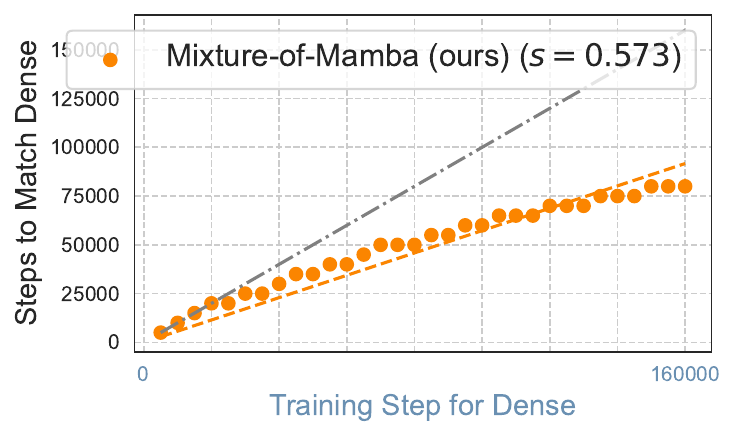}
       \caption{Text Loss Matching}
   \end{subfigure}
    \begin{subfigure}[b]{0.24\textwidth}
        \centering
        \includegraphics[width=\textwidth]{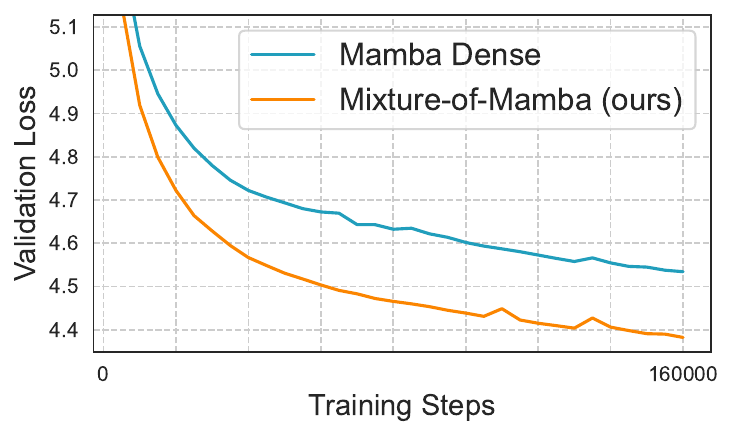}
        \caption{\textbf{443M} Image Eval Loss}
    \end{subfigure}
    \hfill
    \begin{subfigure}[b]{0.24\textwidth}
       \centering
       \includegraphics[width=\textwidth]{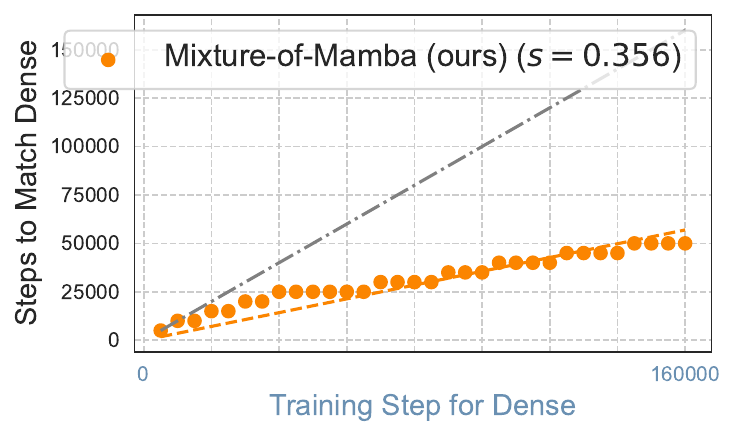}
       \caption{Image Loss Matching}
   \end{subfigure}
   \hfill
   \begin{subfigure}[b]{0.24\textwidth}
        \centering
        \includegraphics[width=\textwidth]{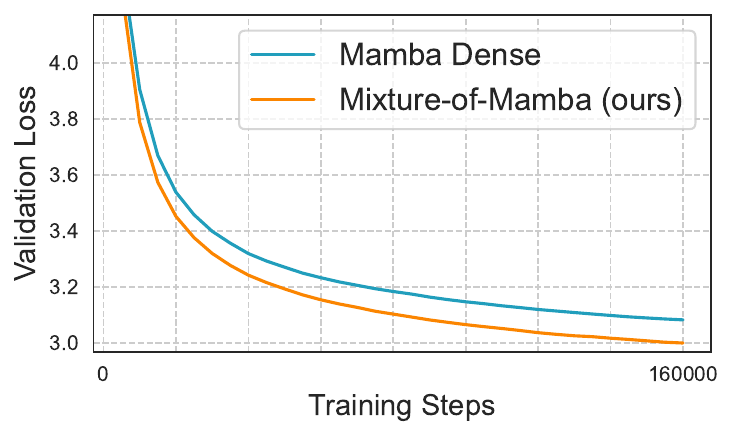}
        \caption{Text Eval Loss}
    \end{subfigure}
    \hfill
    \begin{subfigure}[b]{0.24\textwidth}
       \centering
       \includegraphics[width=\textwidth]{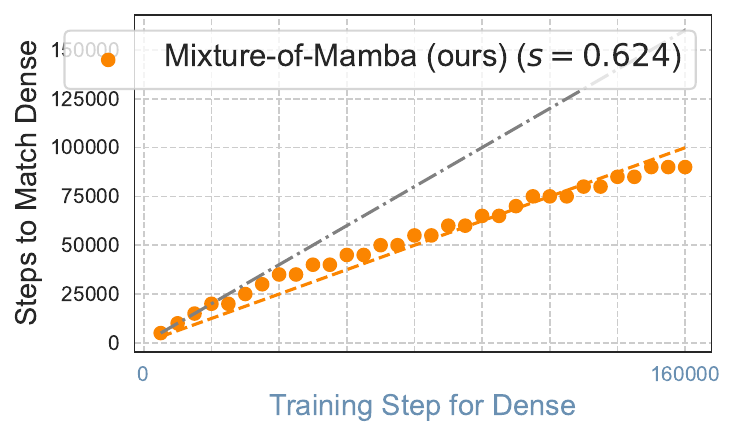}
       \caption{Text Loss Matching}
   \end{subfigure}
    \begin{subfigure}[b]{0.24\textwidth}
        \centering
        \includegraphics[width=\textwidth]{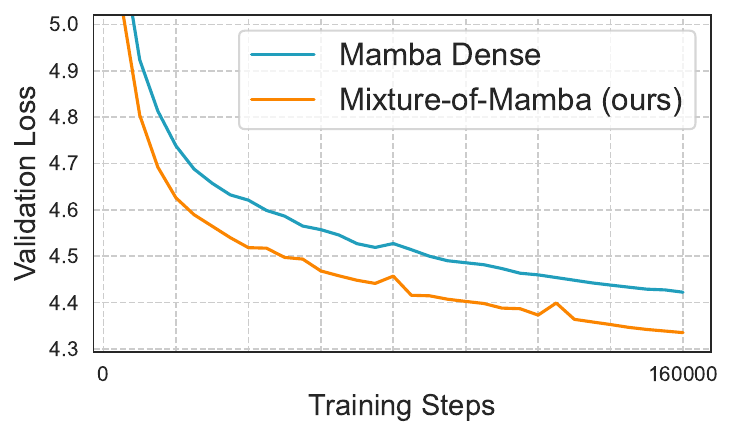}
        \caption{\textbf{880M} Image Eval Loss}
    \end{subfigure}
    \hfill
    \begin{subfigure}[b]{0.24\textwidth}
       \centering
       \includegraphics[width=\textwidth]{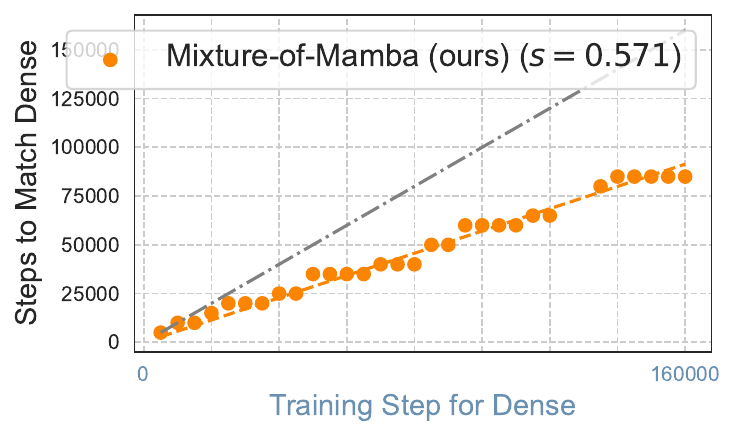}
       \caption{Image Loss Matching}
   \end{subfigure}
   \hfill
   \begin{subfigure}[b]{0.24\textwidth}
        \centering
        \includegraphics[width=\textwidth]{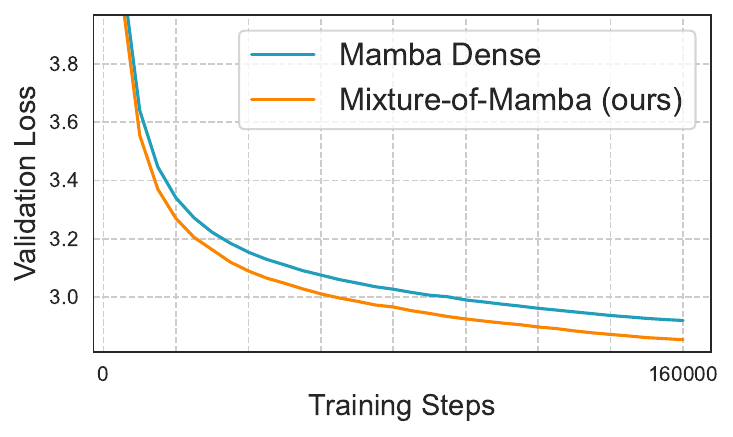}
        \caption{Text Eval Loss}
    \end{subfigure}
    \hfill
    \begin{subfigure}[b]{0.24\textwidth}
       \centering
       \includegraphics[width=\textwidth]{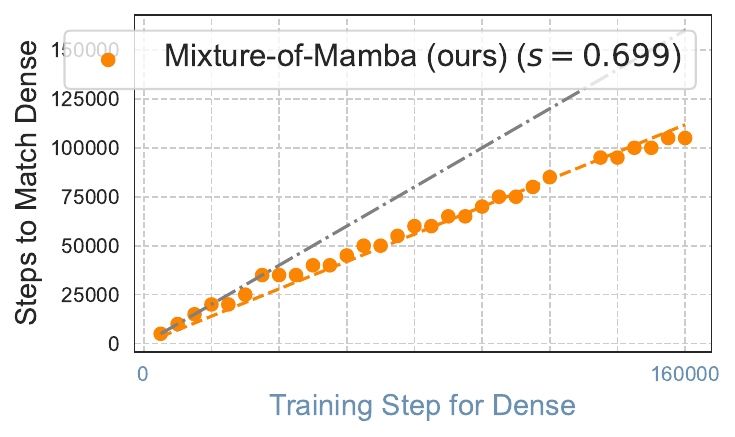}
       \caption{Text Loss Matching}
   \end{subfigure}
    \begin{subfigure}[b]{0.24\textwidth}
        \centering
        \includegraphics[width=\textwidth]{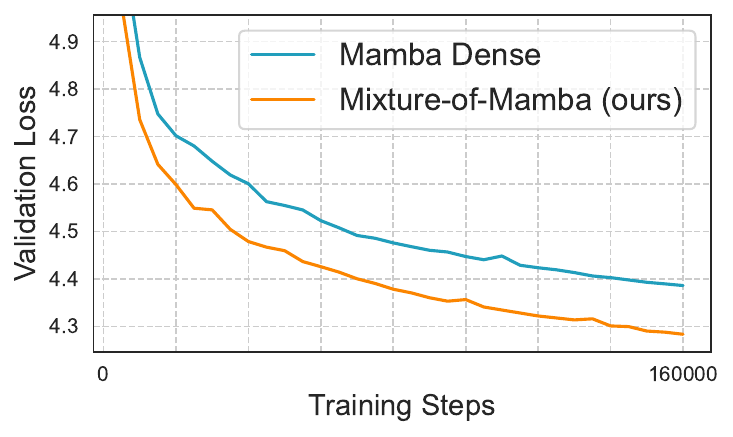}
        \caption{\textbf{1.5B} Image Eval Loss}
    \end{subfigure}
    \hfill
    \begin{subfigure}[b]{0.24\textwidth}
       \centering
       \includegraphics[width=\textwidth]{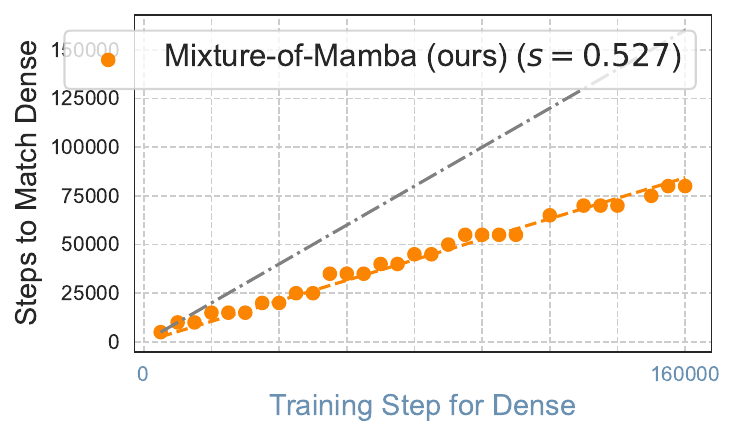}
       \caption{Image Loss Matching}
   \end{subfigure}
   \hfill
   \begin{subfigure}[b]{0.24\textwidth}
        \centering
        \includegraphics[width=\textwidth]{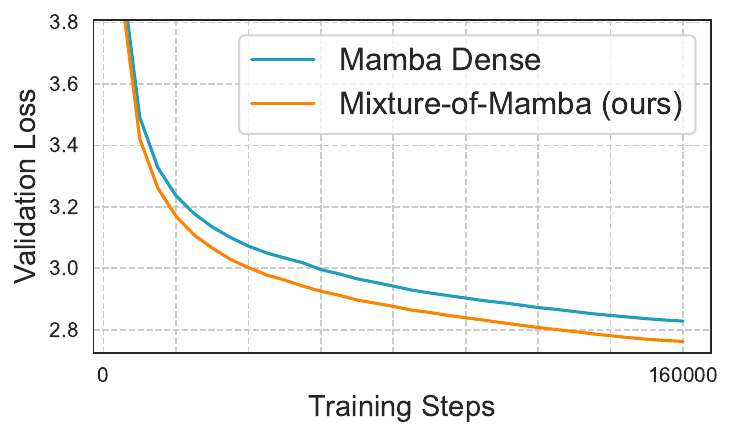}
        \caption{Text Eval Loss}
    \end{subfigure}
    \hfill
    \begin{subfigure}[b]{0.24\textwidth}
       \centering
       \includegraphics[width=\textwidth]{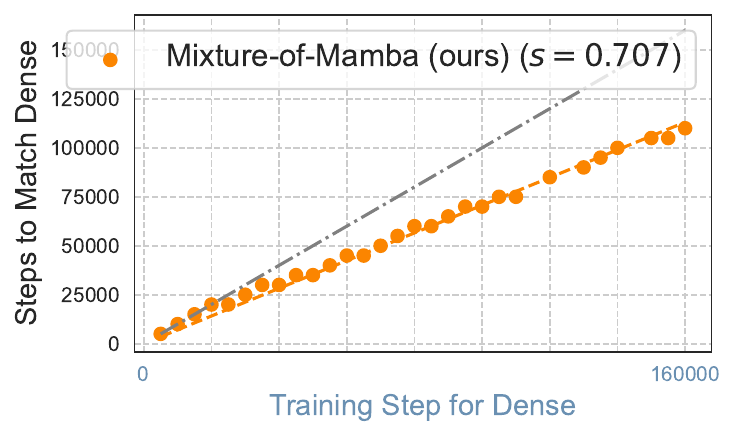}
       \caption{Text Loss Matching}
   \end{subfigure}

\caption{\textbf{Training and validation losses for image and text modalities across model scales in the Chameleon+Speech setting evaluated on the Obelisc dataset.}  
Results are shown for Mixture-of-Mamba and Mamba Dense across five model scales: \textbf{37M}, \textbf{94M}, \textbf{443M}, \textbf{880M}, and \textbf{1.5B}.  
\textbf{(a, e, i, m, q)} Image evaluation loss demonstrates consistent gains for Mixture-of-Mamba (\textcolor{orange}{orange}) over Mamba Dense (\textcolor{cyan}{cyan}), even with the inclusion of the speech modality.  
\textbf{(b, f, j, n, r)} Image loss matching shows that Mixture-of-Mamba reaches the same loss values at earlier training steps compared to Mamba Dense, highlighting improved efficiency.  
\textbf{(c, g, k, o, s)} Text evaluation loss indicates consistent reductions for Mixture-of-Mamba relative to Mamba Dense across all scales.  
\textbf{(d, h, l, p, t)} Text loss matching illustrates that Mixture-of-Mamba reaches the same loss values at earlier training steps compared to Mamba Dense, maintaining its efficiency in the text modality.  
Overall, Mixture-of-Mamba achieves consistent improvements in both image and text modalities while maintaining its efficiency, even with the addition of the \textbf{speech modality}. These results confirm the robustness of Mixture-of-Mamba in multi-modal settings.}
\label{fig:appendix_chameleon_speech_obelisc}

\end{figure*}

\begin{figure*}[t]
    \centering
    \begin{subfigure}[b]{0.24\textwidth}
        \centering
        \includegraphics[width=\textwidth]{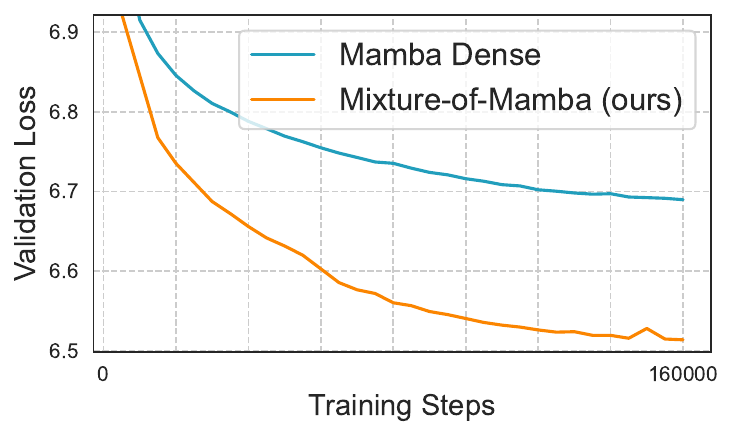}
        \caption{\textbf{37M} Image Eval Loss}
    \end{subfigure}
    \hfill
    \begin{subfigure}[b]{0.24\textwidth}
       \centering
       \includegraphics[width=\textwidth]{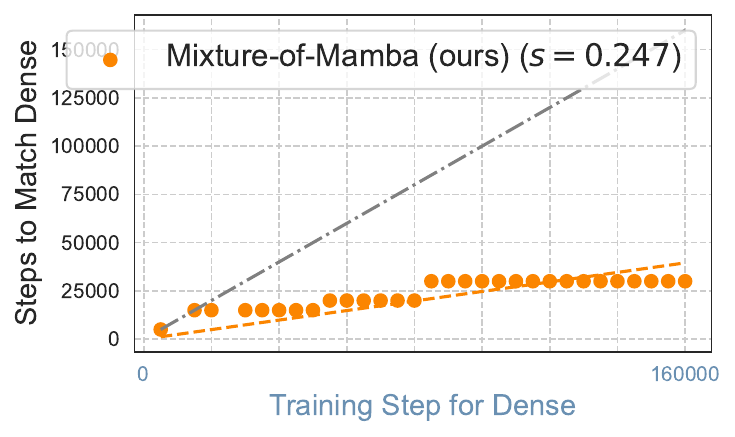}
       \caption{Image Loss Matching}
   \end{subfigure}
   \hfill
   \begin{subfigure}[b]{0.24\textwidth}
        \centering
        \includegraphics[width=\textwidth]{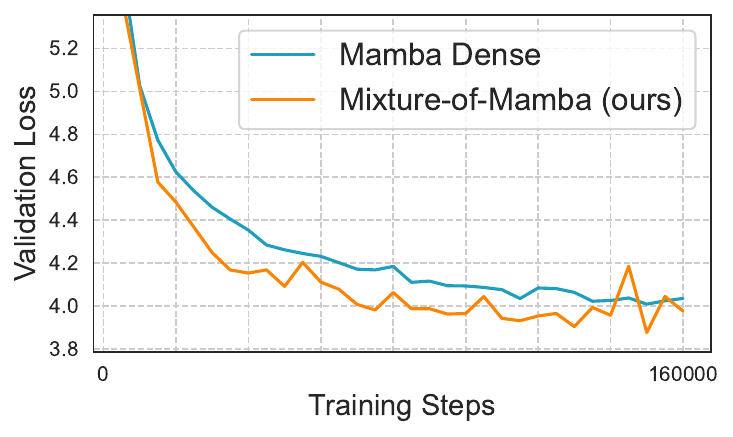}
        \caption{Text Eval Loss}
    \end{subfigure}
    \hfill
    \begin{subfigure}[b]{0.24\textwidth}
       \centering
       \includegraphics[width=\textwidth]{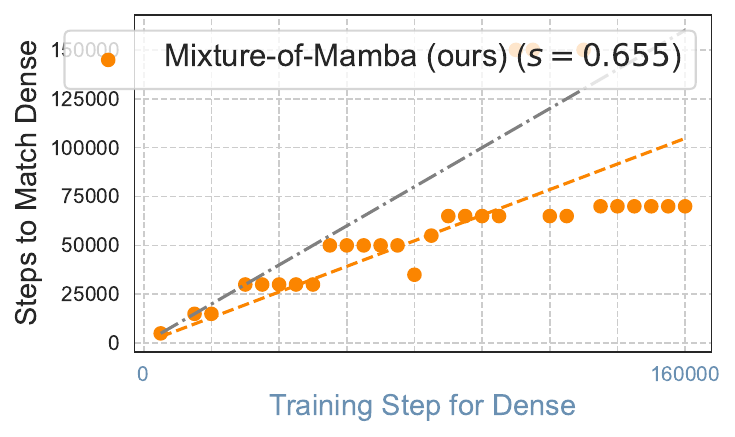}
       \caption{Text Loss Matching}
   \end{subfigure}
    \begin{subfigure}[b]{0.24\textwidth}
        \centering
        \includegraphics[width=\textwidth]{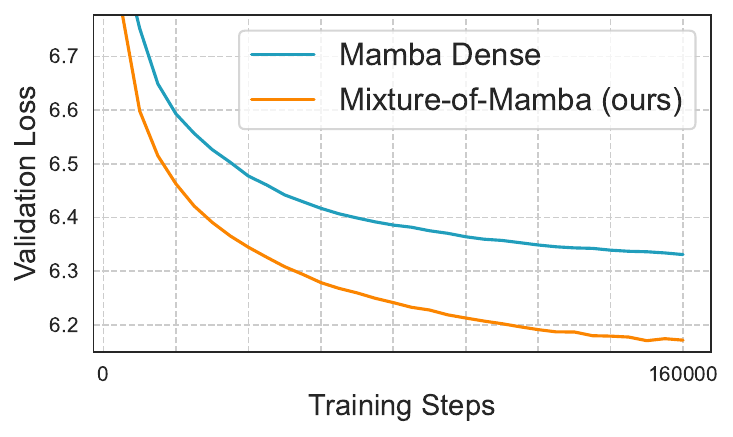}
        \caption{\textbf{94M} Image Eval Loss}
    \end{subfigure}
    \hfill
    \begin{subfigure}[b]{0.24\textwidth}
       \centering
       \includegraphics[width=\textwidth]{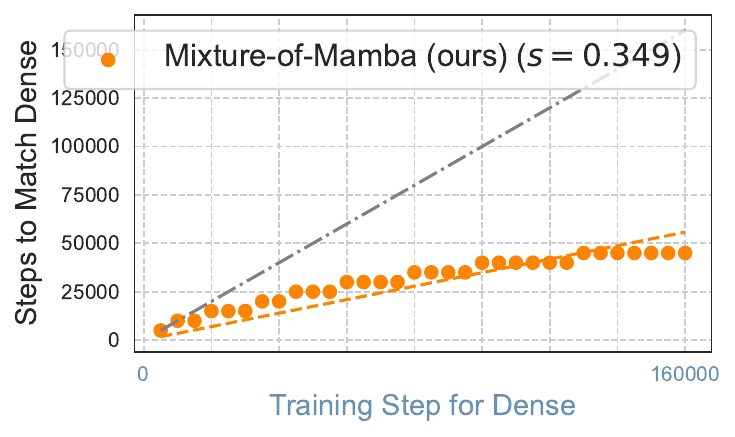}
       \caption{Image Loss Matching}
   \end{subfigure}
   \hfill
   \begin{subfigure}[b]{0.24\textwidth}
        \centering
        \includegraphics[width=\textwidth]{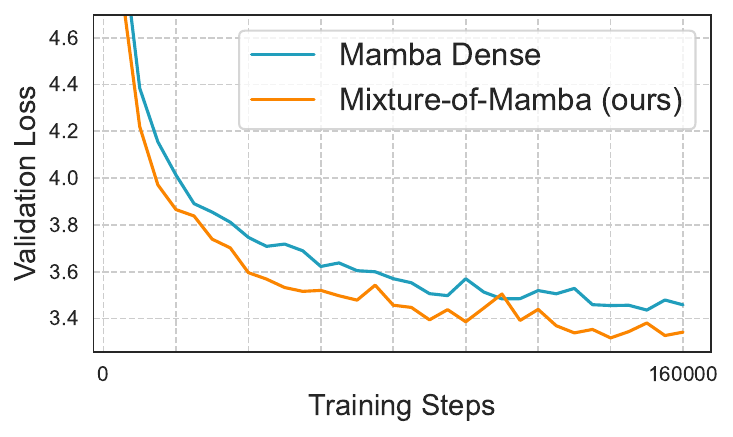}
        \caption{Text Eval Loss}
    \end{subfigure}
    \hfill
    \begin{subfigure}[b]{0.24\textwidth}
       \centering
       \includegraphics[width=\textwidth]{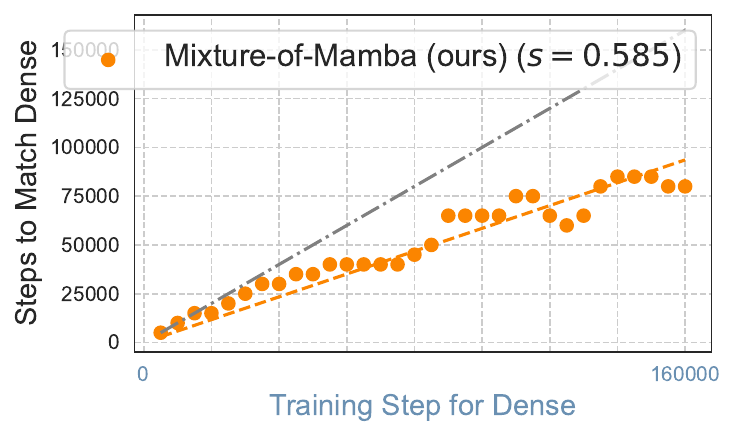}
       \caption{Text Loss Matching}
   \end{subfigure}
    \begin{subfigure}[b]{0.24\textwidth}
        \centering
        \includegraphics[width=\textwidth]{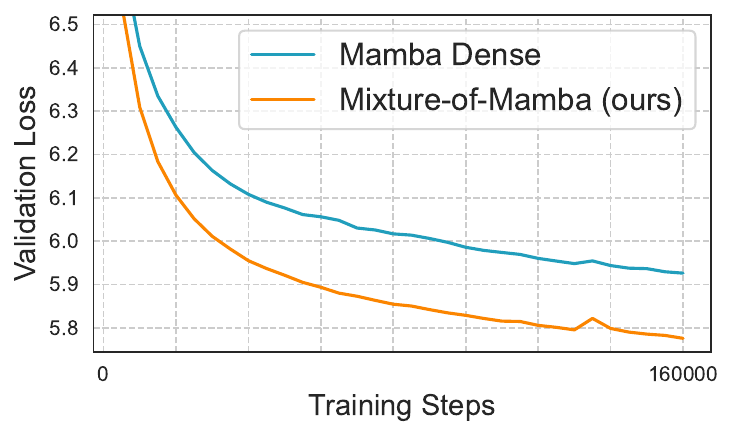}
        \caption{\textbf{443M} Image Eval Loss}
    \end{subfigure}
    \hfill
    \begin{subfigure}[b]{0.24\textwidth}
       \centering
       \includegraphics[width=\textwidth]{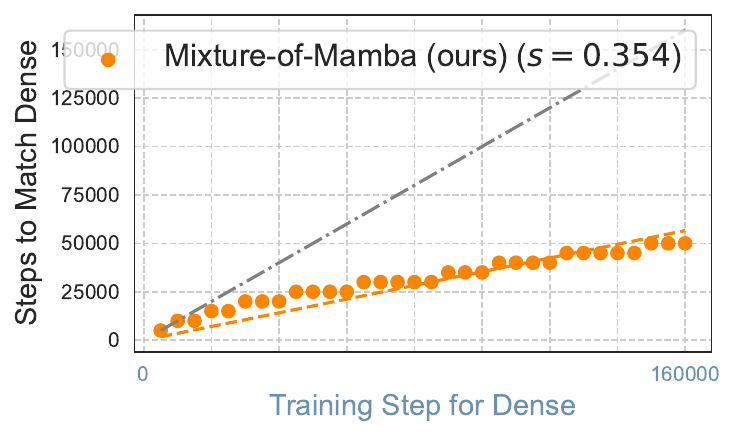}
       \caption{Image Loss Matching}
   \end{subfigure}
   \hfill
   \begin{subfigure}[b]{0.24\textwidth}
        \centering
        \includegraphics[width=\textwidth]{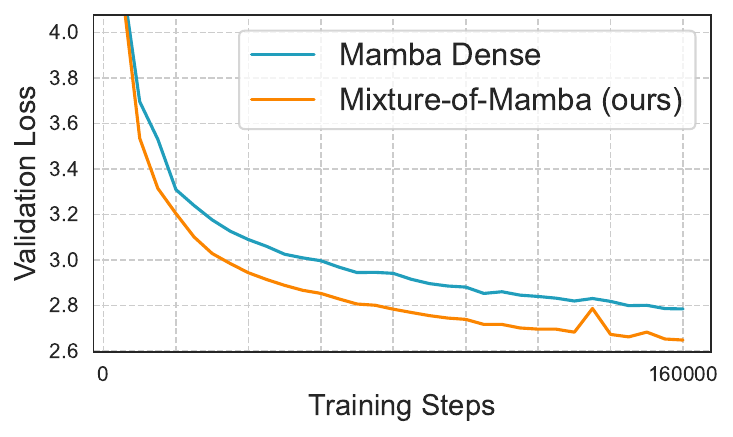}
        \caption{Text Eval Loss}
    \end{subfigure}
    \hfill
    \begin{subfigure}[b]{0.24\textwidth}
       \centering
       \includegraphics[width=\textwidth]{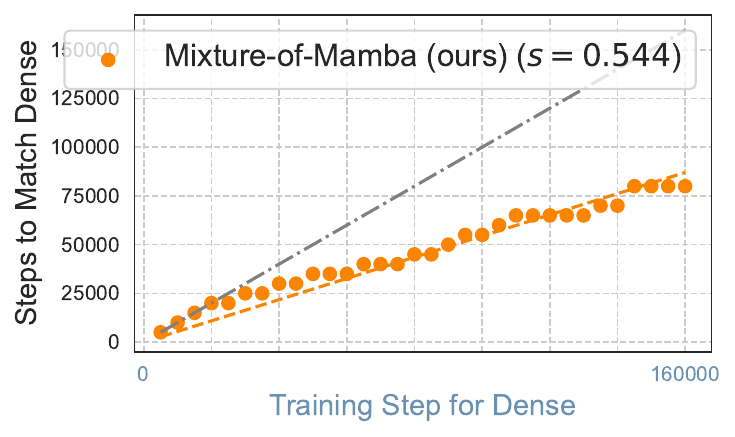}
       \caption{Text Loss Matching}
   \end{subfigure}
    \begin{subfigure}[b]{0.24\textwidth}
        \centering
        \includegraphics[width=\textwidth]{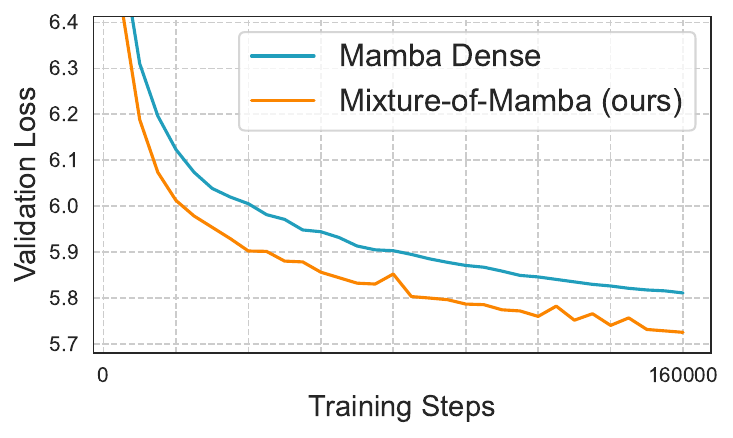}
        \caption{\textbf{880M} Image Eval Loss}
    \end{subfigure}
    \hfill
    \begin{subfigure}[b]{0.24\textwidth}
       \centering
       \includegraphics[width=\textwidth]{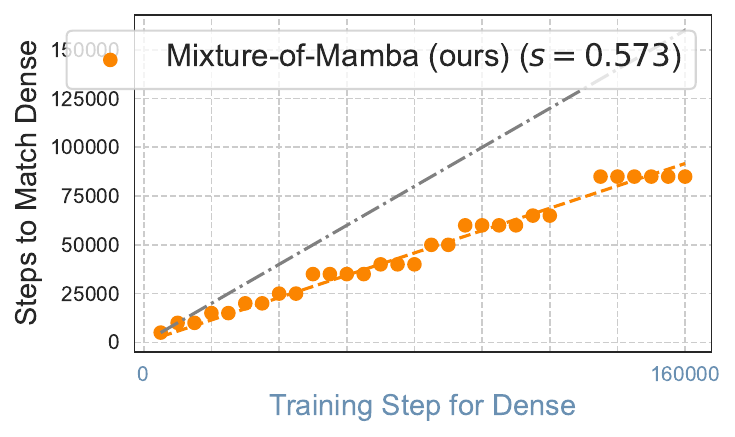}
       \caption{Image Loss Matching}
   \end{subfigure}
   \hfill
   \begin{subfigure}[b]{0.24\textwidth}
        \centering
        \includegraphics[width=\textwidth]{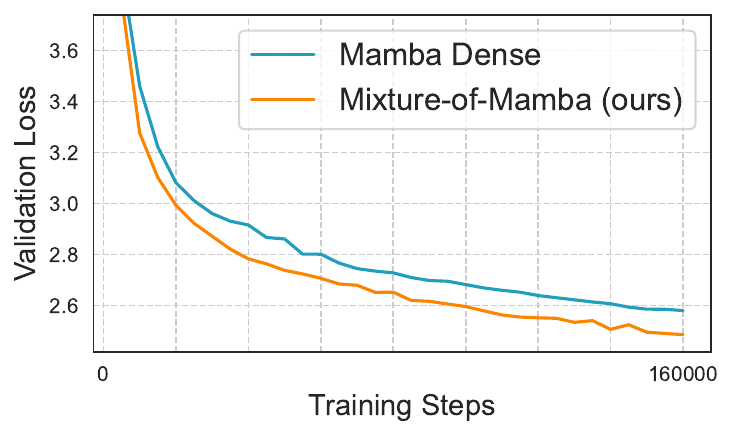}
        \caption{Text Eval Loss}
    \end{subfigure}
    \hfill
    \begin{subfigure}[b]{0.24\textwidth}
       \centering
       \includegraphics[width=\textwidth]{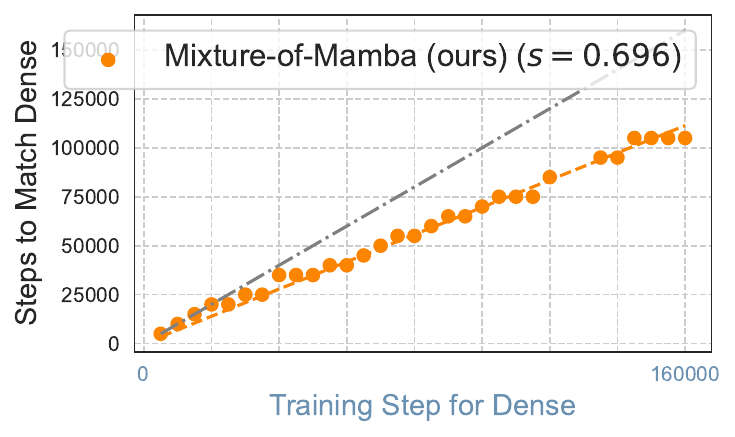}
       \caption{Text Loss Matching}
   \end{subfigure}
    \begin{subfigure}[b]{0.24\textwidth}
        \centering
        \includegraphics[width=\textwidth]{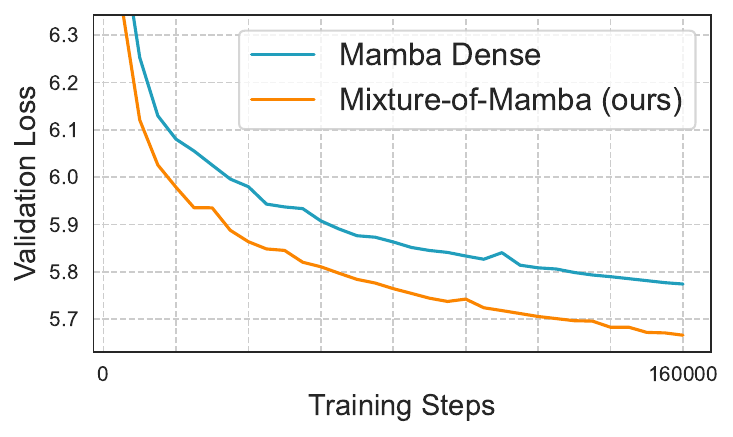}
        \caption{\textbf{1.5B} Image Eval Loss}
    \end{subfigure}
    \hfill
    \begin{subfigure}[b]{0.24\textwidth}
       \centering
       \includegraphics[width=\textwidth]{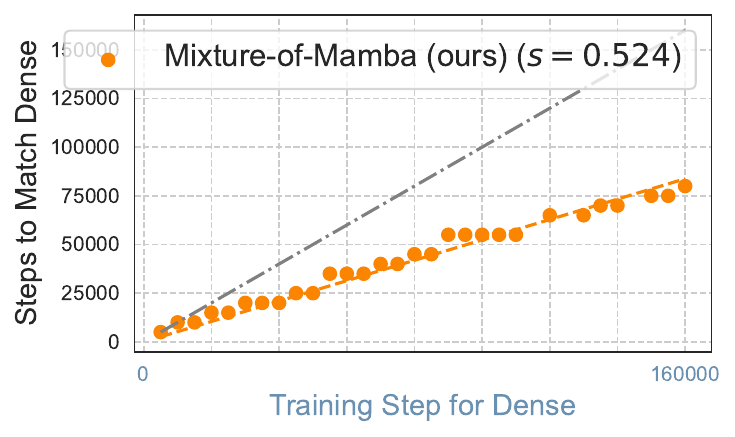}
       \caption{Image Loss Matching}
   \end{subfigure}
   \hfill
   \begin{subfigure}[b]{0.24\textwidth}
        \centering
        \includegraphics[width=\textwidth]{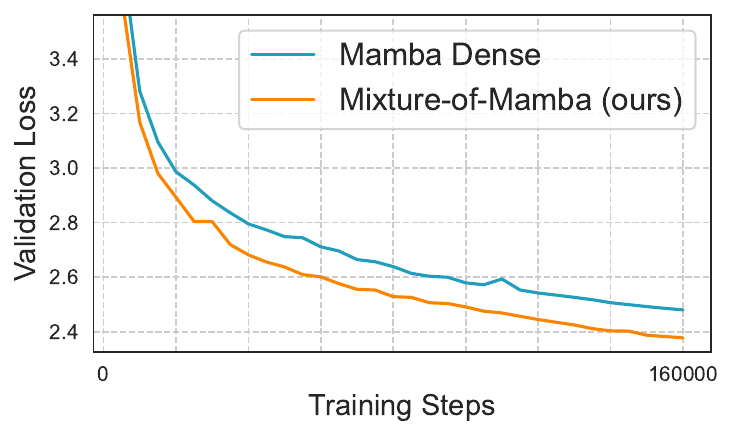}
        \caption{Text Eval Loss}
    \end{subfigure}
    \hfill
    \begin{subfigure}[b]{0.24\textwidth}
       \centering
       \includegraphics[width=\textwidth]{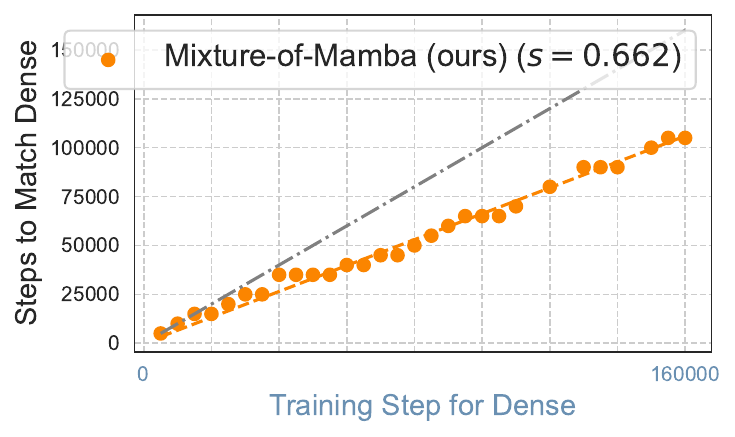}
       \caption{Text Loss Matching}
   \end{subfigure}

\caption{\textbf{Training and validation losses for image and text modalities across model scales in the Chameleon+Speech setting evaluated on the Shutterstock dataset.}  
Results are shown for Mixture-of-Mamba and Mamba Dense across five model scales: \textbf{37M}, \textbf{94M}, \textbf{443M}, \textbf{880M}, and \textbf{1.5B}.  
\textbf{(a, e, i, m, q)} Image evaluation loss demonstrates consistent gains for Mixture-of-Mamba (\textcolor{orange}{orange}) over Mamba Dense (\textcolor{cyan}{cyan}), even with the inclusion of the speech modality.  
\textbf{(b, f, j, n, r)} Image loss matching shows that Mixture-of-Mamba reaches the same loss values at earlier training steps compared to Mamba Dense, highlighting improved efficiency.  
\textbf{(c, g, k, o, s)} Text evaluation loss indicates consistent reductions for Mixture-of-Mamba relative to Mamba Dense across all scales.  
\textbf{(d, h, l, p, t)} Text loss matching illustrates that Mixture-of-Mamba reaches the same loss values at earlier training steps compared to Mamba Dense, maintaining its efficiency in the text modality.  
Overall, Mixture-of-Mamba achieves consistent improvements in both image and text modalities while maintaining its efficiency, even with the addition of the \textbf{speech modality}. These results confirm the robustness of Mixture-of-Mamba in multi-modal settings.}
\label{fig:appendix_chameleon_speech_Shutterstock}

\end{figure*}

\end{document}